%% file: neurips_2026.tex
\documentclass[11pt, a4paper, copyright, frontis-three]{frontis}
\uselogo{}
\setthirdlogo{assets/new-logo}
\fvlogodividerfalse        % (joint mode only) hide the | divider between logos
\fvtitlecentertrue         % center title + author (comment out for left-aligned)

\usepackage[sort&compress]{natbib}
\usepackage[utf8]{inputenc} % allow utf-8 input
\usepackage[T1]{fontenc}    % use 8-bit T1 fonts
\usepackage{hyperref}       % hyperlinks
\usepackage{url}            % simple URL typesetting
\usepackage{booktabs}       % professional-quality tables
\usepackage{array}          % ragged paragraph columns in tables
\usepackage{longtable}      % multi-page experiment table
\usepackage{amsmath}        % math environments
\usepackage{amsfonts}       % blackboard math symbols
\usepackage{amssymb}        % checkmarks and symbols
\usepackage{nicefrac}       % compact symbols for 1/2, etc.
\usepackage{microtype}      % microtypography
\usepackage{xcolor}         % colors
\usepackage{colortbl}       % row backgrounds in the grouped main-results table
\usepackage{graphicx}       % figures
\usepackage{float}          % fixed-position tables when draft ordering matters
\usepackage{tikz}           % architecture diagrams
\usepackage{fontawesome5}   % resource badges
\usepackage{enumitem}
\usepackage{wrapfig}
\usepackage{fvextra}
\usetikzlibrary{arrows.meta, positioning, fit}

\definecolor{todojunlin}{HTML}{1F5AA6}
\definecolor{todoche}{HTML}{6A4C93}
\definecolor{todocan}{HTML}{2D6A4F}
\definecolor{todoyu}{HTML}{B7791F}
\definecolor{todokaiyan}{HTML}{006D77}
\definecolor{todoweizhi}{HTML}{8A5A44}
\DeclareRobustCommand{\openmletodo}[3]{\textcolor{#1}{\textbf{[#2 TODO: #3]}}}

\makeatletter
\protected\def\todo_#1#{\openmletodo@dispatch{#1}}
\def\openmletodo@dispatch#1#2{%
  \@ifundefined{openmletodo@#1}%
    {\PackageError{OpenMLE}{Unknown todo owner '#1'}{Allowed todo owners: junlin, che, can, yu, kaiyan, weizhi.}%
     \openmletodo{black!65}{#1}{#2}}%
    {\csname openmletodo@#1\endcsname{#2}}}
\def\openmletodo@junlin#1{\openmletodo{todojunlin}{Junlin}{#1}}
\def\openmletodo@che#1{\openmletodo{todoche}{Che}{#1}}
\def\openmletodo@can#1{\openmletodo{todocan}{Can}{#1}}
\def\openmletodo@yu#1{\openmletodo{todoyu}{Yu}{#1}}
\def\openmletodo@kaiyan#1{\openmletodo{todokaiyan}{Kaiyan}{#1}}
\def\openmletodo@weizhi#1{\openmletodo{todoweizhi}{Weizhi}{#1}}
\makeatother
\newcommand{\cmark}{\(\checkmark\)}
\newcommand{\xmark}{\(\times\)}
\newcommand{\openmlegym}{\textsc{OpenMLE-Gym}}
\newcommand{\openmleerl}{\textsc{OpenMLE-ERL}}
\newcommand{\openmleevo}{\textsc{OpenMLE-Evo}}
\newcommand{\openmleevomax}{\textsc{OpenMLE-Evo-Max}}
\expandafter\def\csname 4b_model\endcsname{Frontis-MA1-4B}
\expandafter\def\csname 30b_model\endcsname{Frontis-MA1-30B}
\expandafter\def\csname 35b_model\endcsname{Frontis-MA1-35B}
\newcommand{\fourbmodel}{\csname 4b_model\endcsname}
\newcommand{\thirtybmodel}{\csname 30b_model\endcsname}
\newcommand{\thirtyfivebmodel}{\csname 35b_model\endcsname}

\newcommand{\baseThirtyFiveEvoResult}{39.39\%}
\newcommand{\frontisThirtyEvoResult}{53.03\%}
\newcommand{\frontisThirtyEvoMaxResult}{66.67\%}
\newcommand{\frontisThirtyFiveEvoResult}{60.61\%}
\newcommand{\frontisThirtyFiveEvoMaxResult}{71.21\%}
\newcommand{\frontisThirtyEvoHumanRank}{0.7055}
\newcommand{\frontisThirtyEvoMaxHumanRank}{0.8053}
\newcommand{\frontisThirtyFiveEvoHumanRank}{0.7647}
\newcommand{\frontisThirtyFiveEvoMaxHumanRank}{0.8126}
\definecolor{badgebg}{HTML}{E5F6F8}
\definecolor{badgeborder}{HTML}{8DCFD6}
\definecolor{nbboxa}{HTML}{CFE0F3}   % subfigure demo fills
\definecolor{nbboxb}{HTML}{FBDDB8}

\newcommand{\badgeicon}[1]{\makebox[1.0em][c]{#1}}

\newcommand{\buildbadge}[2]{% #1 = icon, #2 = label
\tikz[baseline=(t.base)]{
  \node[
    fill=badgebg,
    draw=badgeborder,
    rounded corners=4pt,
    minimum height=20pt,
    inner xsep=6pt,
    inner ysep=2pt
  ] (t) {%
    {\fontfamily{ppl}\selectfont\small\strut
      \badgeicon{#1}\kern0.35em #2}%
  };
}%
}

\newsavebox{\badgeProj}
\newsavebox{\badgeGH}
\newsavebox{\badgeHF}

\savebox{\badgeProj}{%
\buildbadge{\textcolor{black}{\faGlobe}}{Project}%
}
\savebox{\badgeGH}{%
\buildbadge{\textcolor{black}{\faGithub}}{GitHub}%
}
\savebox{\badgeHF}{%
\buildbadge{%
  \raisebox{-0.05em}{\includegraphics[height=0.85em]{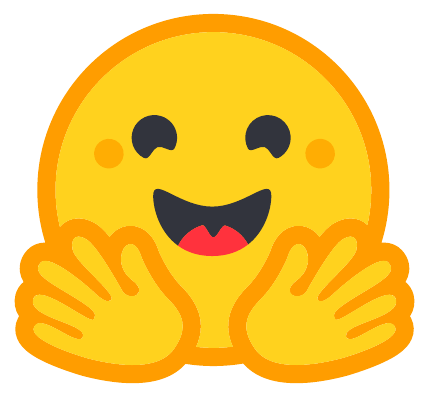}}%
}{HuggingFace}%
}

\newcommand{\badgerow}{%
\makebox[\linewidth][c]{%
  \href{https://frontisai.github.io/OpenRSI}{\usebox{\badgeProj}}\hspace{0.65em}%
  \href{https://github.com/FrontisAI/OpenRSI}{\usebox{\badgeGH}}\hspace{0.65em}%
  \href{https://huggingface.co/collections/FrontisAI/frontis-ma1}{\usebox{\badgeHF}}%
}%
}

\renewcommand{\titlefont}{\color{black}\normalfont\bfseries\fontsize{17}{21}\selectfont}
\title{Frontis-MA1: Training an AI4AI Model towards Recursive Self-Improvement in Machine Learning Engineering}
\author{Horizon Research, Frontis.AI \quad Tsinghua University \\
  \rule{0pt}{0.3em}\badgerow\vspace{-8pt}}

\begin{document}

\begin{abstract}
Recursive self-improvement (RSI) requires AI systems that improve the process of
building AI (i.e., AI4AI); machine learning engineering (MLE) offers a concrete,
executable testbed for studying this capability.
We introduce \textbf{OpenMLE}, an open full-stack system for RSI research in MLE,
spanning verifiable task environments with execution feedback (\openmlegym{}),
operator learning (\openmleerl{}), and long-horizon search (\openmleevo{}).
On this stack we post-train \textbf{\thirtyfivebmodel{}} as a
\textbf{\emph{meta-evolution agent}} for MLE, aligning post-training and
inference around four atomic program-evolution operators (Draft, Improve, Debug,
Crossover): the same operators are trained via execution-grounded SFT and RL on data
deduplicated against all evaluation benchmarks, then composed into long-horizon
search, coupling learning and evolution in a single loop.
On MLE-Bench Lite under 
% a fixed 12 GPU-hour per-task budget
a 12-hour per-task budget on one RTX 4090 capped at 12 GB VRAM, \thirtyfivebmodel{} improves Medal Average from
\baseThirtyFiveEvoResult{} to \frontisThirtyFiveEvoResult{} over its base model
with \openmleevo{}, and reaches \frontisThirtyFiveEvoMaxResult{} with
\openmleevomax{} (benchmark-independent experience priors and asynchronous search),
exceeding GPT-5.5 + Codex and approaching
GPT-5.6 Sol and the 2.8T Kimi K3.
On held-out NatureBench Lite, both components transfer: with the framework fixed,
swapping in the trained model raises Match-SOTA from 50\% to 70\%; with the model
fixed, swapping in \openmleevo{} raises it from 20\% to 50\%.
We release the model weights and the full OpenMLE stack to enable reproducible
research on executable AI4AI toward RSI.

\par\vspace{0.2em}
\textcolor{frontiscyan}{\rule{\linewidth}{0.3pt}}
\par\vspace{0.2em}

{\footnotesize\color{frontisgray}
\textbf{\color{black}Project Leaders}
\textcolor{frontiscyan}{$\vert$}
Junlin Yang, Che Jiang\par

\textbf{\color{black}Correspondence}
\textcolor{frontiscyan}{$\vert$}
Kaiyan Zhang,
\href{mailto:zhangkaiyan@frontis.cn}
{\texttt{zhangkaiyan@frontis.cn}}\par

% \textbf{\color{black}Contact}
% \textcolor{frontiscyan}{$\vert$}
% \texttt{\{yangjunlin,jiangche,zhangkaiyan\}@frontis.cn}\par
}
\end{abstract}

\maketitle

\begin{figure}[H]
  \centering
  \includegraphics[width=0.98\linewidth]{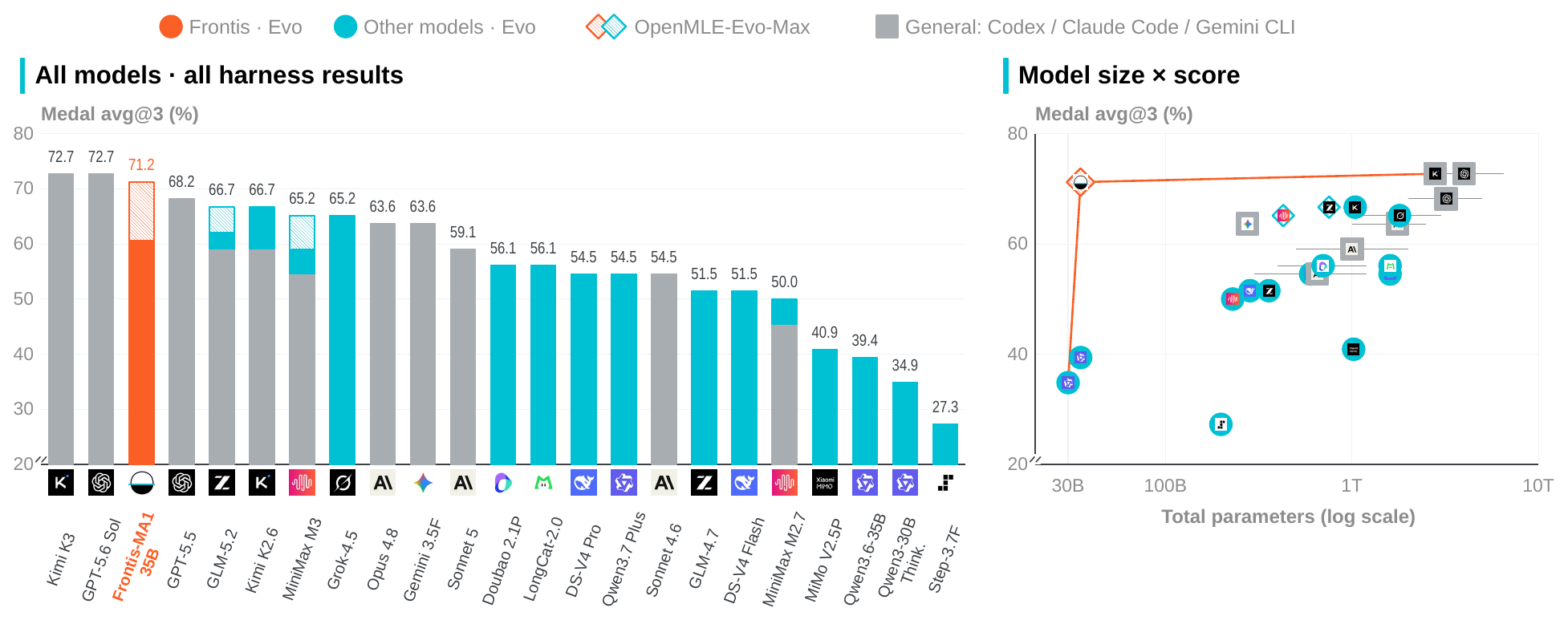}
  \caption{Results on MLE-Bench Lite. Bars show all completed harness results; the Pareto panel retains each model's best harness. Colors and hatching follow the shared model--harness legend.}
  \label{fig:main-exp-systems}
\end{figure}

% \begin{figure}[H]
%   \centering
%   \includegraphics[width=\linewidth]{figures/frontis-ma1_en_cc.pdf}
%   \caption{Main results of Frontis-MA1 35B on MLE-Bench-Lite}
%   \label{fig:frontis-results-overview}
% \end{figure}

\begingroup
\setlength{\baselineskip}{1.0\baselineskip}   % 注意前面有倍数
\tableofcontents
\endgroup

% \junlin{Reinsert a teaser only after regenerating it with the final 30B/35B release models and results.}

\input{sections/01_introduction}

\input{sections/02_problem_setting}

\input{sections/04_framework}

\input{sections/06_main_results}

\input{sections/03_related_work}

\input{sections/08_limitations_and_discussion}

\input{sections/09_conclusion}

\newpage
\input{sections/10_references}

\appendix

\input{sections/11_technical_appendices}

% \newpage
% \input{checklist.tex}

\end{document}

%% file: sections/01_introduction.tex
\newpage
\section{Introduction}
\label{sec:intro}

AI capability growth is no longer pushed only by human engineers.
Increasingly, AI systems write code, run experiments, search over designs, and help build the next generation of AI systems~\citep{romera2024mathematical,lu2024aiscientist,novikov2025alphaevolve,oh2025discovering}.
This broader direction, often called AI for AI (AI4AI), seeks to use AI systems to build and improve AI~\citep{liu2025ml,chan2026measuring}.
Its more ambitious endpoint is recursive self-improvement (RSI), where each improved system further improves the process that produces its successors~\citep{good1965speculations,schmidhuber2003godel,eth2025softwareintelligence,favaro2026whenai}.
Reaching that endpoint requires more than stronger one-shot generation or planning.
It requires agents that can perform \emph{AI training AI} and AutoResearch: inspect data, propose algorithms, execute experiments, diagnose failures, and decide how to spend the next unit of compute~\citep{lu2024aiscientist,nathani2025mlgym,karpathy2026autoresearch}.

Machine learning engineering (MLE) is a particularly direct instantiation of AI4AI: an agent must build a machine learning solution for a real-world task and iteratively improve it through execution feedback~\citep{huang2023mlagentbench,chan2024mlebench,qiang2025mledojo}.
A trajectory often begins with a valid pipeline and advances through repeated experiments toward a solution competitive with strong human or frontier-model pipelines~\citep{nam2025mlestar,hambardzumyan2026aira_2,du2026mlevolve}.
Each iteration consumes time and compute, and its outcome may arrive only after minutes or hours.
This makes MLE a concrete and demanding testbed for studying how agents improve AI systems under delayed, noisy, and heterogeneous feedback~\citep{jing2024dsbench,nathani2025mlgym,lupidi2026airs}.

\begin{figure}[H]
  \centering
  \includegraphics[width=\linewidth]{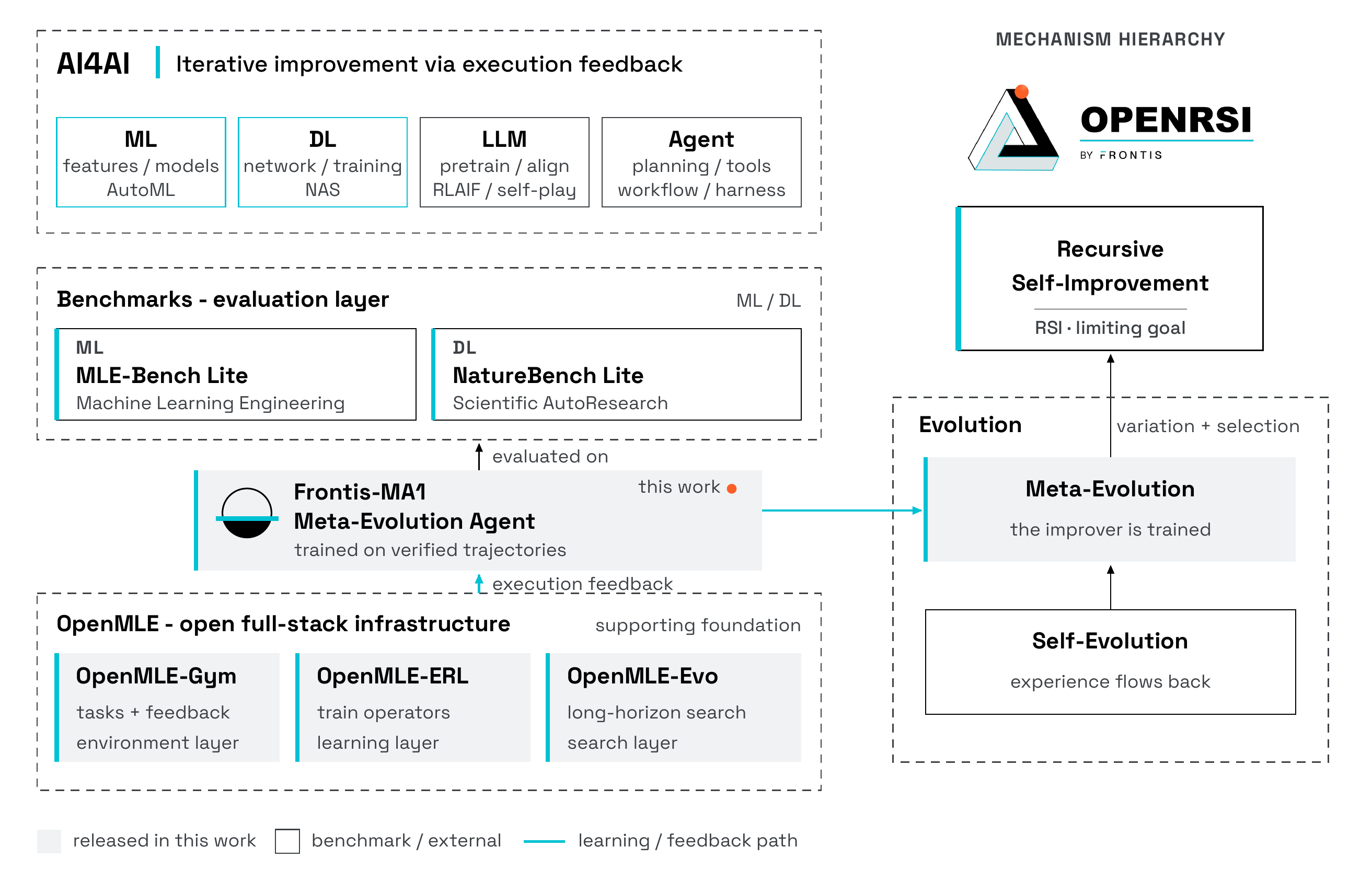}
    \caption{Positioning of this work. \textbf{Left:} within AI4AI, machine learning
  engineering (MLE) is our task domain; the OpenMLE stack trains and deploys
  Frontis-MA1, which is both the product of the stack and its engine, evaluated
  only on third-party benchmarks. \textbf{Right:} the mechanism ladder from
  evolution to recursive self-improvement; the meta-evolutionary loop (orange)
  places this work at the level where the improver itself is trained---the level after which Frontis-\textbf{MA}1 (\textbf{M}eta-evolution \textbf{A}gent) is named.}
  \label{fig:positioning}
\end{figure}

Prior work advances MLE agents along three complementary but overlapping strands. The first develops inference-time harnesses based on structured or evolutionary search~\citep{jiang2025aide,fang2025mlzero,nam2025mlestar,du2025automlgen,liu2025ml,zhu2026toward,toledo2025airesearchagents,internscience2026mlevolve}. The second builds executable tasks and environments~\citep{qiang2025mledojo,qiang2025mlesmith,nathani2025mlgym,lupidi2026airs}. The third uses execution feedback to post-train MLE agents~\citep{liu2025mlagent,li2025mlerl,yang2025rlmleagents,cai2026acegrpo}. Some systems bridge subsets of these strands---MLE-Dojo supports model tuning, while AceGRPO recycles iterative execution traces into training~\citep{qiang2025mledojo,cai2026acegrpo}---but, among the representative public systems audited in Appendix Table~\ref{tab:mle-open-release-matrix}, none jointly spans scalable task and environment construction, execution-grounded agent post-training, and an evolutionary harness that deploys the trained agent in long-horizon search, together with the artifacts needed to reproduce the full loop.

We introduce \textbf{OpenMLE}, an open full-stack technical solution for training language models to construct and iteratively improve machine learning solutions using executable feedback: \openmlegym{} constructs 5,758 quality-gated executable tasks and provides isolated execution, structured feedback, and task-specific evaluation; \openmleerl{} uses budget-adaptive supervision fine-tuning and reinforcement learning to turn verified solutions and revisions into stronger MLE behavior; and \openmleevo{} organizes long-horizon search around structured experience, non-greedy selection over quality, progress, and novelty, and operator-conditioned memory. OpenMLE exposes \textsc{Draft}, \textsc{Improve}, \textsc{Debug}, and \textsc{Crossover} as a shared interface between post-training and inference. This interface lets verified evolutionary transitions supervise the same transformations that search later composes, making the trained model the variation engine of the evolutionary harness and forming the meta-evolutionary loop illustrated in Figure~\ref{fig:positioning}, in which the improver itself is trained.
A model trained and deployed in this role is a \textbf{\emph{meta-evolution agent}} as introduced in \citep{jiang2026selfimprovingagents}.

Using OpenMLE, we train \textbf{\thirtyfivebmodel{}}
(\textbf{M}eta-evolution \textbf{A}gent, generation~1) as our primary
model and evaluate it at the model, harness, and system levels on MLE-Bench Lite under a fixed 12 GPU-hour per-task budget. At the model level, under the identical \openmleevo{} harness, \thirtyfivebmodel{} improves over Qwen3.6-35B-A3B from $39.39\%$ to \frontisThirtyFiveEvoResult{} Medal Average and from $0.5828$ to \frontisThirtyFiveEvoHumanRank{} Human Rank. At the harness level, matched comparisons show that \openmleevo{} outperforms general-purpose Claude Code or Codex scaffolds across four frontier models and original AIRA-Evo on \thirtyfivebmodel{}. At the system level, \thirtyfivebmodel{} with \openmleevomax{}\footnote{\openmleevomax{} is an enhanced OpenMLE configuration; see Section~\ref{sec:experimental-setup} for details.} reaches \frontisThirtyFiveEvoMaxResult{} Medal Average and \frontisThirtyFiveEvoMaxHumanRank{} Human Rank, exceeding GPT-5.5 + Codex. As a controlled cross-model replication, \thirtybmodel{} improves over Qwen3-30B-A3B-Thinking-2507 from $34.85\%$ to \frontisThirtyEvoResult{} under the same \openmleevo{} harness. Finally, controlled NatureBench~\citep{wang2026naturebench} Lite comparisons provide evidence that both execution-grounded post-training and adapted evolutionary search transfer beyond competition-style MLE.
We will release the datasets, training and evaluation code, sandbox infrastructure, harness code, and final post-trained checkpoints. This release will make the complete OpenMLE workflow reproducible, enabling broader study of meta-evolution in executable AI4AI settings and its role on the path toward RSI. Our contributions are:
\begin{enumerate}[leftmargin=*,itemsep=2pt,topsep=0pt,partopsep=0pt,parsep=0pt]
\item We present OpenMLE as a full-stack technical solution for studying recursive self-improvement in executable MLE, connecting verifiable environments, post-training, and test-time evolution in one validated workflow.
\item We introduce \openmlegym{}, which unifies 5,758 challenging tasks with isolated execution, structured feedback, and task-specific evaluators for scalable training, search, and evaluation.
\item We introduce \openmleerl{} to train reusable \textsc{Draft}, \textsc{Improve}, \textsc{Debug}, and \textsc{Crossover} operators through execution-grounded supervised fine-tuning and reinforcement learning.
\item We introduce \openmleevo{} to support experience-guided long-horizon search over the environments and feedback exposed by \openmlegym{}.
\item Using the complete stack, we train and release \thirtyfivebmodel{} as our primary model, together with \thirtybmodel{} as a companion model for controlled cross-model replication. Evaluations separate the gains from post-training and search on MLE-Bench Lite and provide transfer evidence on NatureBench Lite. We also release the datasets, training and evaluation code, execution infrastructure, and search framework needed to reproduce and extend OpenMLE.
\end{enumerate}

%% file: sections/02_problem_setting.tex
\section{Problem Formulation}
\label{sec:problem-setting}

\textbf{From AI for AI to meta-evolution.}
AI for AI describes the object of optimization: AI systems participate in creating or improving other AI systems, including training code, model architectures, algorithms, agent harnesses, and AI hardware.
Evolution describes the optimization process: an AI repeatedly modifies a candidate system according to feedback from its execution.
Meta-evolution closes an additional learning loop by reusing these evolutionary trajectories to train the model that proposes future modifications.
Recursive self-improvement (RSI) requires a stronger and sustained loop in which each upgraded system further improves the process that produces its successors.
OpenMLE studies a concrete step from evolution toward meta-evolution: executable search experience is reused through supervised fine-tuning and reinforcement learning to improve the program-transformation model.

\textbf{Executable evolution for AI-building tasks.}
We use machine learning engineering (MLE) as a measurable testbed for this process.
Following MLE-Dojo and MLE-Bench~\citep{qiang2025mledojo,chan2024mlebench}, each task $\tau$ contains a natural-language specification, visible data assets, a submission contract, a task-specific evaluator, and a sandboxed execution environment.
At step $t$, the inference-time search algorithm selects an operator
$a_t$ and constructs its operator context $c_t$ from zero or more parent
programs and their execution feedback.
The model then proposes
\[
p_t \sim g_\theta(\cdot\mid\tau,a_t,c_t),
\qquad
s_t=R_\tau(\mathcal{E}(p_t,\tau)).
\]
Here, $g_\theta$ denotes the operator-conditioned program-generation
policy parameterized by the language-model parameters $\theta$.
Given task $\tau$, operator $a_t$, and context $c_t$, it defines a
distribution over candidate programs.
The sandbox $\mathcal{E}$ executes $p_t$ and returns the task score $s_t$ mapped by a task-specific evaluator $R_\tau$ together
with status, logs, artifacts, and runtime metadata.
Because task metrics have different ranges and directions, Section~\ref{sec:execution-grounded-rl} converts $s_t$ into a signed score $\tilde{s}_t$, for which larger is always better, and then into a normalized reward.

Each program and its execution feedback are stored in a task-local
program database, and program-transformation operators are repeatedly
applied to generate new candidates.
Within a finite execution budget, evolutionary inference seeks the
candidate with the highest signed score:
\[
p^\star
=
p_{\arg\max_{t\in\mathcal{I}}\tilde{s}_t},
\]
where $\mathcal{I}$ denotes the index set of all candidate programs
recorded in the task-local program database during the budgeted search.
OpenMLE instantiates the operator space with \textsc{Draft},
\textsc{Improve}, \textsc{Debug}, and \textsc{Crossover}, while leaving
their global composition to the inference-time search algorithm.

\textbf{Learning to evolve from executable experience.}
Meta-evolution improves $\theta$ so that
$g_\theta(\cdot\mid\tau,a,c)$ assigns higher probability to programs
with stronger execution outcomes.
Both SFT and RL can be summarized as optimizing
\[
\mathcal{L}_{\mathrm{evo}}(\theta)
=
-
\mathbb{E}_{(\tau_i,a_i,c_i,p_i)}
\left[
w(s_i)
\log g_\theta
\left(
p_i\mid\tau_i,a_i,c_i
\right)
\right],
\]
where
\[
s_i=R_{\tau_i}\!\left(\mathcal{E}(p_i,\tau_i)\right)
\]
is the execution score and $w(s_i)$ converts this outcome into a
learning weight.
For SFT, quality filtering retains high-scoring programs $p_i^+$ and
assigns them positive supervision.
For RL, newly sampled programs are weighted by their processed
execution rewards and entropic advantages within the clipped policy
objective.
Thus, both stages update the same parameters $\theta$ to make
high-quality executable programs more likely under
$g_\theta(\cdot\mid\tau,a,c)$.

%% file: sections/04_framework.tex
\section{OpenMLE-Gym: Building Scalable Verifiable Environments}
\label{sec:framework}
\label{sec:task-curation-data-construction}

\begin{takeaway}
\begin{itemize}[leftmargin=*,itemsep=1pt,topsep=2pt,partopsep=0pt,parsep=0pt]
\item \textbf{MLE requires a compute-backed, verifiable gym.} LLM agents can be improved through interaction with executable, data-grounded environments. For MLE, the environment must host large, resource-intensive tasks, execute candidate programs under controlled budgets, and apply task-specific evaluators to return reproducible diagnostics and rewards. \openmlegym{} provides this substrate for post-training, search, and evaluation toward recursive self-improvement.
\item \textbf{The gym scales through automated task construction.} Our construction and quality-control pipeline turns curated anchors, Kaggle datasets, and Kaggle competitions into 5,758 quality-gated executable environments spanning eight modalities and six core task types, with executability validation and semantic quality gating.
\end{itemize}
\end{takeaway}

AI-for-AI requires agents to build and improve AI systems through executable experimentation, making machine learning engineering (MLE) a concrete testbed for this loop. Such agents need large-scale, diverse, high-quality task packages spanning data inspection, preparation, modeling, prediction, and evaluation. Yet these resources remain scarce: existing benchmarks are limited in openness, executable diversity, or coverage, while heterogeneous files, formats, and evaluation protocols make them difficult to scale.

Static task collections alone do not support post-training or search. Agent-generated programs must execute under controlled resources and produce reliable, informative feedback that guides subsequent actions. Gym-style environments make this interaction contract explicit by coupling tasks with actions, transitions, observations, rewards, and stopping conditions~\citep{brockman2016openai,xi2024agentgym,nathani2025mlgym,qiang2025mledojo}. OpenMLE-Gym therefore unifies scalable executable tasks with standardized packages, isolated execution, structured diagnostics, and task-specific evaluation for post-training, test-time search, and AI-for-AI.

\subsection{Environment Contract}
\label{sec:gym-environment-contract}
Each OpenMLE-Gym task is an environment instance defined by five elements. The task/state consists of the task specification, public data, hidden evaluator, resource budget, and current workspace state. The action is an agent-submitted MLE program together with its execution requirements. The transition is sandbox execution: the environment materializes the workspace, runs the program against public data, and invokes the evaluator when a valid submission is produced. The observation is a structured record containing execution status, task score, logs, error types, generated artifacts, and runtime metadata. The reward is the verifiable task-specific score returned by the evaluator.
Task completion, controller stopping criteria, and time or compute budgets determine whether an interaction terminates or is truncated.

\begin{figure}[t]
  \centering
  \includegraphics[page=1,width=\linewidth]{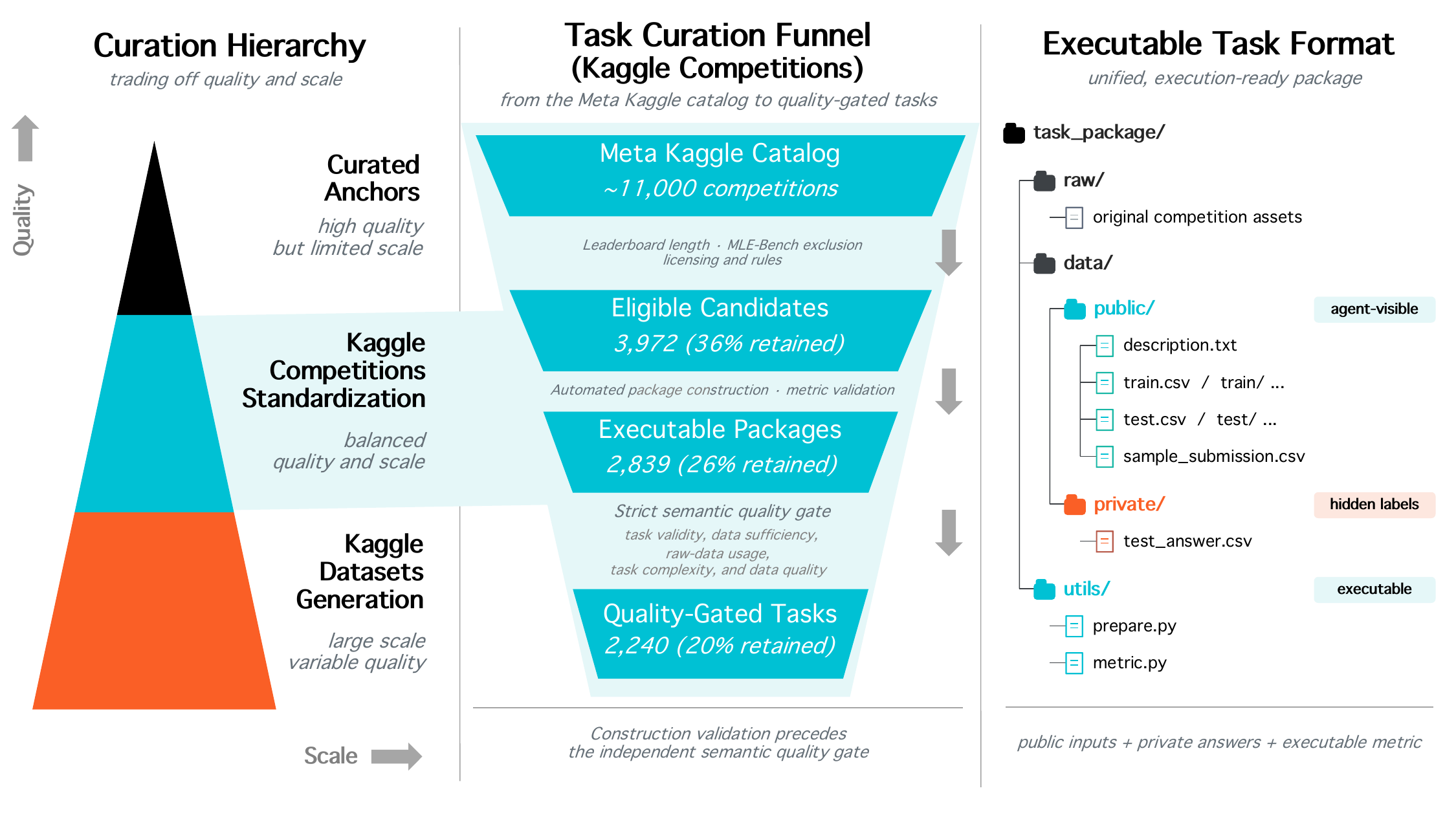}
  \caption{OpenMLE-Gym task curation and executable format. \textbf{Left:} source hierarchy. \textbf{Middle:} Kaggle Competition filtering from the Meta Kaggle catalog to quality-gated packages. \textbf{Right:} public task inputs, private answers, and executable utilities.}
  \label{fig:task-curation-summary}
\end{figure}

\subsection{Scalable Task Construction}
\textbf{OpenMLE-Gym} is our unified executable task suite, constructed from three source-specific paths that occupy complementary positions in the quality--scale trade-off (Figure~\ref{fig:task-curation-summary}, left). \textbf{Curated Anchors} provide the highest-confidence task designs because they are manually selected from existing papers and benchmarks, but their dependence on prior expert curation limits scale. We download their original assets and process them directly into executable packages. \textbf{Kaggle Datasets} substantially broaden task and data coverage, although automatically induced objectives can have more variable quality. We utilize and extend the existing MLE-Smith dataset-to-task pipeline~\citep{qiang2025mlesmith}, then apply package-level quality control. \textbf{Kaggle Competitions} offer a middle ground: human-authored problem specifications, evaluation metrics, and submission protocols provide stronger task grounding, while prior participant engagement and leaderboard records offer additional external evidence that tasks support meaningful, comparable evaluation. Meanwhile, the Meta Kaggle catalog supports collection at scale. We build our own crawling and competition-construction pipeline, with MLE-Bench-overlapping competitions excluded to preserve evaluation integrity.

Following the principles of MLE-Dojo~\citep{qiang2025mledojo}, we map all three sources into a shared executable task package. Original assets remain under \texttt{raw/}. Agent-visible descriptions, training data, test inputs, and sample submissions are placed under \texttt{data/public/}, while hidden answers are isolated under \texttt{data/private/}. A task-specific \texttt{metric.py} validates prediction files and returns scalar execution feedback. This contract gives heterogeneous tasks the same agent-facing interaction surface while preserving their original semantics.

For the competition-derived branch, our automated framework converts a Kaggle competition slug into a standardized executable package. It downloads and inventories the competition files, then combines local evidence---including schemas, data types, dimensions, and sampled rows---with Meta Kaggle records describing the problem, evaluation criterion, and submission protocol. From this grounded context, the framework constructs the task description and generates \texttt{prepare.py} to deterministically split labeled data, isolate public inputs from private answers, and produce a schema-compatible sample submission. It then generates \texttt{metric.py} and validates the complete package by executing the preparation code and scoring the sample submission. Failures and assertion errors are returned as feedback for bounded retries; packages that remain unbuildable or fail to produce a valid scalar metric are removed before semantic quality filtering. A detailed stage-by-stage view is provided in Appendix~\ref{app:data-construction} (Figure~\ref{fig:competition-task-construction}).

\subsection{Task Quality Filtering}
To ensure the quality of constructed task packages, we propose an LLM-based quality filter. For each task package, the filter jointly inspects the task description, raw files, processing script, processed outputs, representative data samples, etc. It returns structured judgments along five dimensions: task validity, data sufficiency, raw-data usage, task complexity, and data quality. Together, these criteria identify degenerate targets solvable by trivial rules, inadequate training or evaluation signal, superficial use of source assets, mismatched difficulty, data leakage, annotation errors, and malformed processing. For example, the competition branch applies this final semantic gate only after leaderboard-length screening, MLE-Bench overlap removal, licensing and competition-rule screening and executable package and metric validation (Figure~\ref{fig:task-curation-summary}, middle). We retain only metric-valid tasks receiving the strict \texttt{recommended} decision, thereby separating semantic quality assessment from the executability check performed during construction.

\begin{figure*}[!t]
  \centering
  \includegraphics[page=2,width=\textwidth]{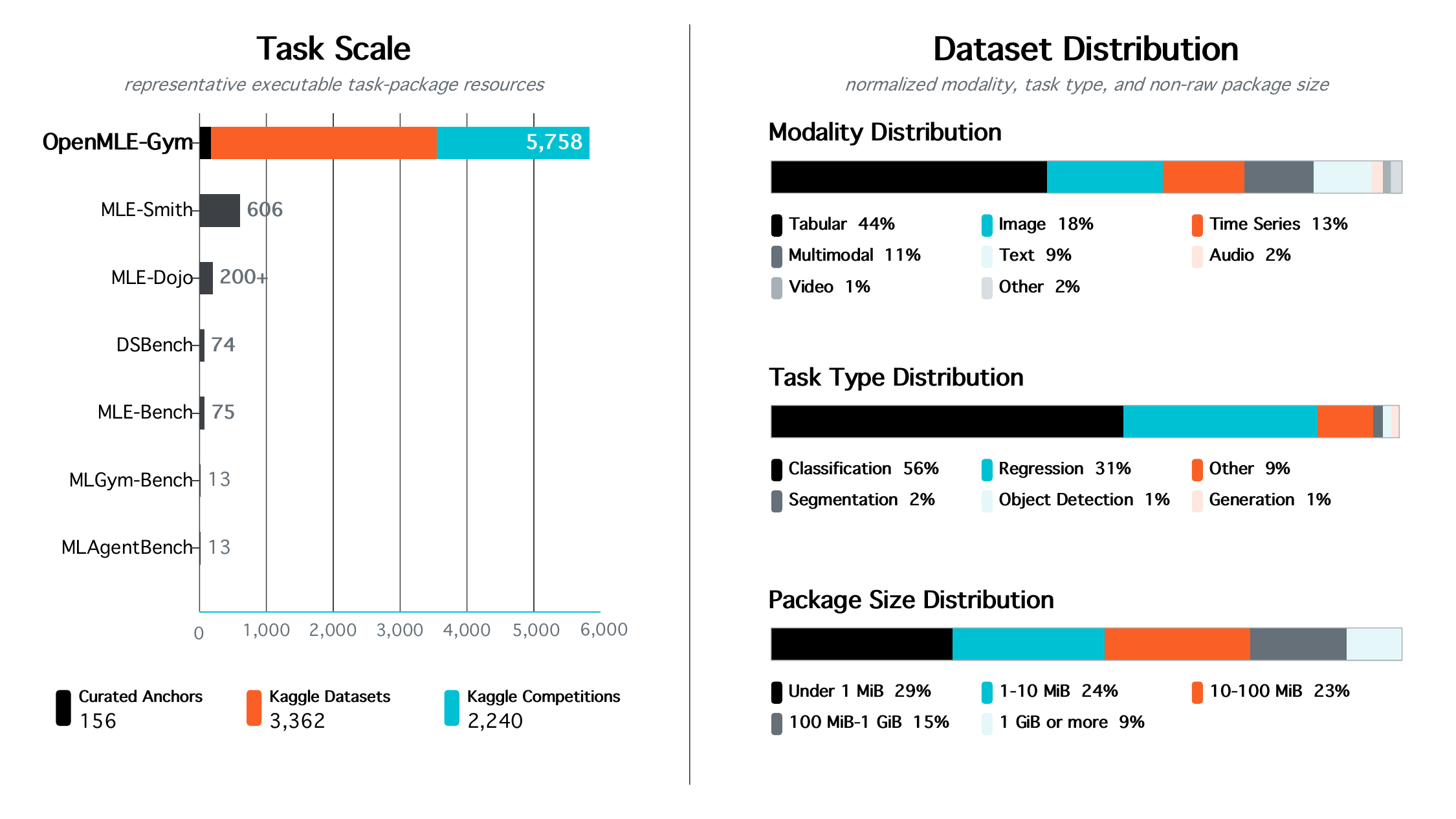}
  \caption{Scale and composition of OpenMLE-Gym. \textbf{Left:} comparison with executable task-package resources and source breakdown of 5,758 tasks. \textbf{Right:} normalized modality, task-type, and non-raw package-size distributions.}
  \label{fig:task-scale-distribution}
\end{figure*}

\subsection{Sandbox Execution at Scale}
\label{sec:gym-sandbox-execution}
The contract is realized at scale by a shared sandbox execution backend. A centralized scheduler receives API requests, records each job, tracks worker availability, and dispatches requests to CPU/GPU Docker workers according to resource requirements. Each worker materializes an isolated task workspace, mounts the task data and evaluator, executes the candidate program, and writes logs, submissions, outputs, and artifacts back to shared storage. This separation of control, execution, and storage supports reproducible scoring and scalable parallel execution across long-running MLE jobs.

The backend returns six feedback modes: successful completion, runtime error, missing code, missing submission, scoring failure, and timeout. Each record preserves the triggering condition together with status, score when available, logs, error type, runtime metadata, and workspace artifacts, allowing the agent to distinguish invalid execution from weak task performance. Detailed architecture and representative feedback cases are provided in Appendix~\ref{app:sandbox}.

\subsection{Composition and Statistics}
OpenMLE-Gym contains 5,758 executable tasks\footnote{Owing to source-data licensing and copyright constraints, we release full task-package data for 1,415 tasks. For the remaining 4,343 tasks, we release the corresponding \texttt{prepare.py} and \texttt{metric.py} scripts without redistributing the source data.}: 156 manually selected Curated Anchors, 3,362 Kaggle Dataset tasks generated from the official MLE-Smith table and quality-controlled at the package level, and 2,240 tasks retained from our Kaggle Competition pipeline (Figure~\ref{fig:task-scale-distribution}, left). Across the three sources, the pool covers tabular, text, time-series, image, and other data modalities, with classification and regression complemented by more engineering-intensive task types. After canonicalizing modality and task-type labels, 11\% of tasks are multimodal and classification and regression together account for 87\% (Figure~\ref{fig:task-scale-distribution}, right). Additional construction details are provided in Appendix~\ref{app:data-construction}.

\section{OpenMLE-ERL: Reinforcing Reusable Evolutionary Operators}

\begin{takeaway}
    \begin{itemize}[leftmargin=*,itemsep=1pt,topsep=2pt,partopsep=0pt,parsep=0pt]
    \item \textbf{RSI training should improve the best solution a model can find within a fixed budget.} 
    Execution-grounded SFT increases strong solutions and useful revisions under repeated sampling; RL uses adaptive score bounds and entropic advantages to reward candidate quality rather than validity alone.
    \item \textbf{Verification cost must shape collection and training.} Budget-adaptive SFT stops at an accepted-example quota or execution limit, preserving budget for sparse-success tasks; asynchronous RL admits completed generation-and-execution groups immediately instead of waiting for the slowest job.
    \item \textbf{Evolutionary search depends on learning revisions, not only fresh drafts.} 
    We train \textsc{Draft}, \textsc{Improve}, \textsc{Debug}, and \textsc{Crossover} from trajectory transitions; stateful RL selects parents using reward, child-reward variance, and visit-based cooling.
    \end{itemize}
\end{takeaway}

\textbf{Motivation.}
Evolutionary AutoResearch is judged by the best executable program found within a finite search budget. A controller may invoke \textsc{Draft}, \textsc{Debug}, \textsc{Improve}, and \textsc{Crossover} hundreds or thousands of times, so the model must learn more than one-shot solution generation: it must repeatedly repair, refine, and recombine programs as the search unfolds. Post-training should therefore expand the set of strong programs reachable through repeated sampling while improving the individual transformations that the inference harness composes.

SFT and RL play complementary roles in meeting this objective. Execution-grounded SFT distills complete solutions and successful local revisions from stronger teachers, giving the model a broader set of executable behaviors before online learning begins. RL then moves probability toward the better candidates within that broadened distribution. This division is motivated by prior analyses of RLVR: RL can raise Pass@1 by reinforcing already-rewarded solutions yet provide limited gains at large $K$, whereas teacher distillation can introduce behaviors absent from the base model's sampled support~\citep{yue2025doesrl}.

Executable evolutionary training also differs from short-horizon RLVR on mathematics and code generation~\citep{shao2024deepseekmath}. Many programs fail to produce a usable reward, while successful programs receive continuous task scores drawn from different metrics and ranges. Feedback arrives only after sandbox runs that may last minutes or hours, and every non-\textsc{Draft} action depends on the parent program selected for expansion. OpenMLE-ERL addresses these constraints across both stages: SFT collection uses execution results and per-task budgets to select useful supervision, while RL preserves score differences among strong candidates, removes synchronization stalls, and trains each operator on informative program states.

\begin{figure}[!htbp]
  \centering
  \includegraphics[width=\textwidth]{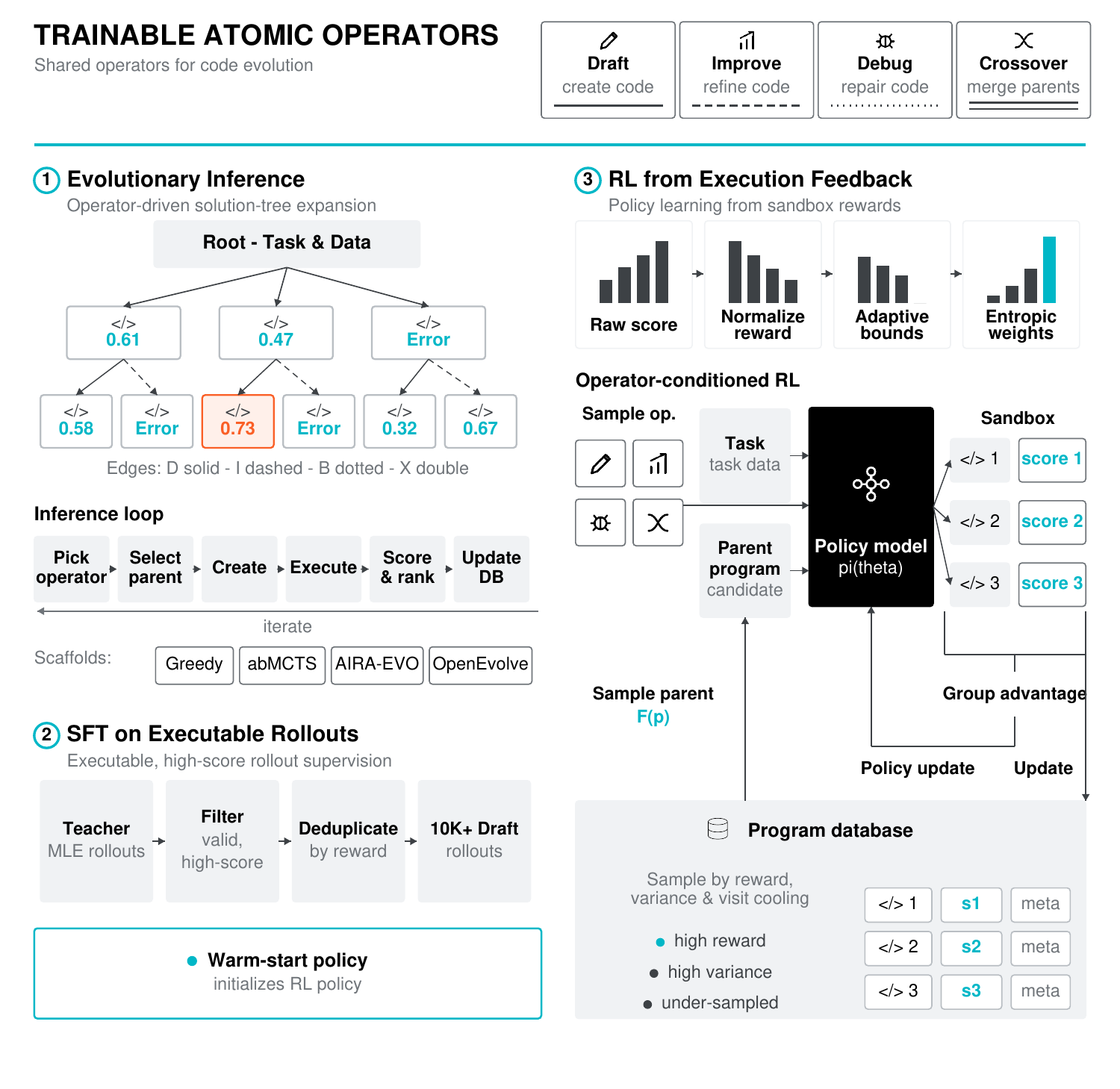}
  \caption{Overview of the OpenMLE training and inference workflow. Trainable atomic operators are used by evolutionary inference, warm-started through executable SFT rollouts, and further optimized with online RL from execution feedback.}
  \label{fig:openmle-framework}
\end{figure}

\subsection{Trainable Atomic Operators for Evolutionary Inference}
\label{sec:atomic-operators}

A central design principle of OpenMLE is to separate the local skills learned by the model from the search algorithm used at inference time. Rather than training full trajectories, OpenMLE trains a compact set of reusable program-transformation operators over executable candidates. This avoids sparse controller-specific supervision and lets the same learned operators be composed by different evolutionary search procedures under a shared sandbox protocol.

We instantiate the atomic operators as \textsc{Draft}, \textsc{Improve}, \textsc{Debug}, and \textsc{Crossover}. This operator vocabulary follows AIRA-style executable MLE search and AIDE-style code-space exploration~\citep{toledo2025airesearchagents,hambardzumyan2026aira_2,jiang2025aide}, but OpenMLE adapts it as explicit SFT and RL targets for open-model post-training.
Detailed operator definitions, prompt templates, and controller-specific search details are provided in the appendix.

\subsection{Execution-Grounded Supervised Warm Start}
\label{sec:supervised-warm-start}

\textbf{Execution-grounded, budget-adaptive collection.}
For each task, we execute sampled programs in the sandbox and retain examples according to their validity and task-specific scores. Collection stops when either an accepted-example quota is reached or the task exhausts its execution budget, allowing easy tasks to terminate early while allocating more attempts to tasks with sparse successes. This difficulty-aware rejection strategy directs verification compute toward tasks for which further exploration can still recover useful supervision~\citep{tong2024dartmath}.

\textbf{Parallel and evolutionary sampling paths.}
The \emph{parallel path} independently samples and executes complete \textsc{Draft} solutions, contributing 17,245 full-response examples to the released corpus. The \emph{evolutionary path} applies \textsc{Improve}, \textsc{Debug}, and \textsc{Crossover} over executed programs and retains useful steps from high-quality local trajectory segments, contributing 9,014 trajectory-step examples. Figure~\ref{fig:sft-rl-rollout} illustrates both paths. Overall, the two paths form the 26,259-example SFT corpus released with OpenMLE; detailed corpus statistics are provided in Appendix~\ref{app:sft-warm-start}.

\subsection{Execution-Grounded Reinforcement Learning}
\label{sec:execution-grounded-rl}

\begin{figure}[H]
  \centering
  \includegraphics[width=\linewidth]{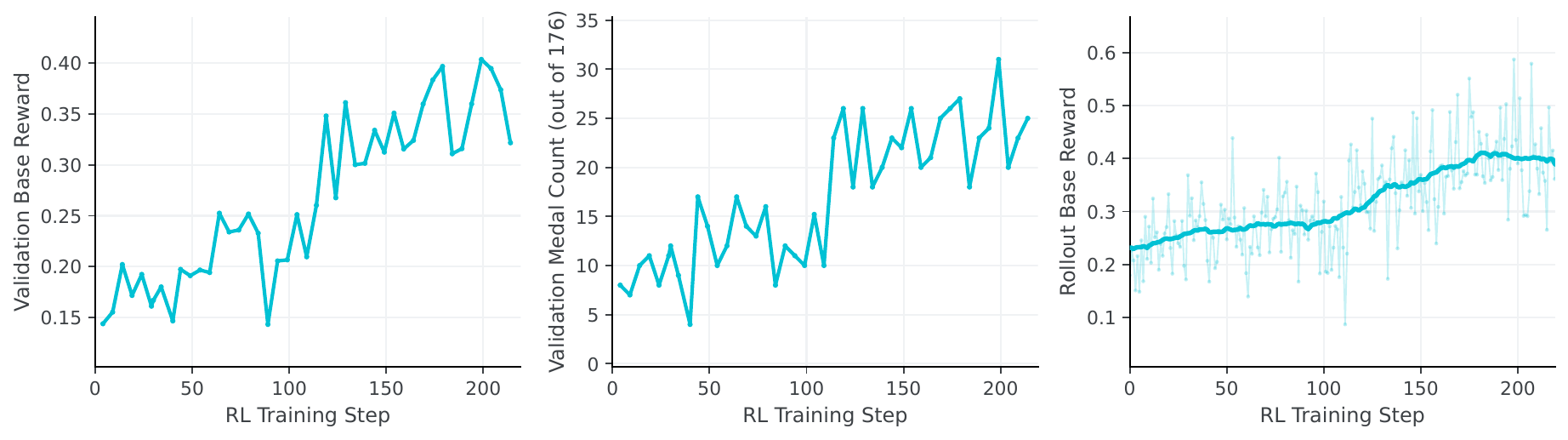}
  \caption{RL training curves for \thirtyfivebmodel{}.}
  \label{fig:qwen36-rl-curves}
\end{figure}

\begin{figure}[H]
  \centering
  \includegraphics[width=\linewidth]{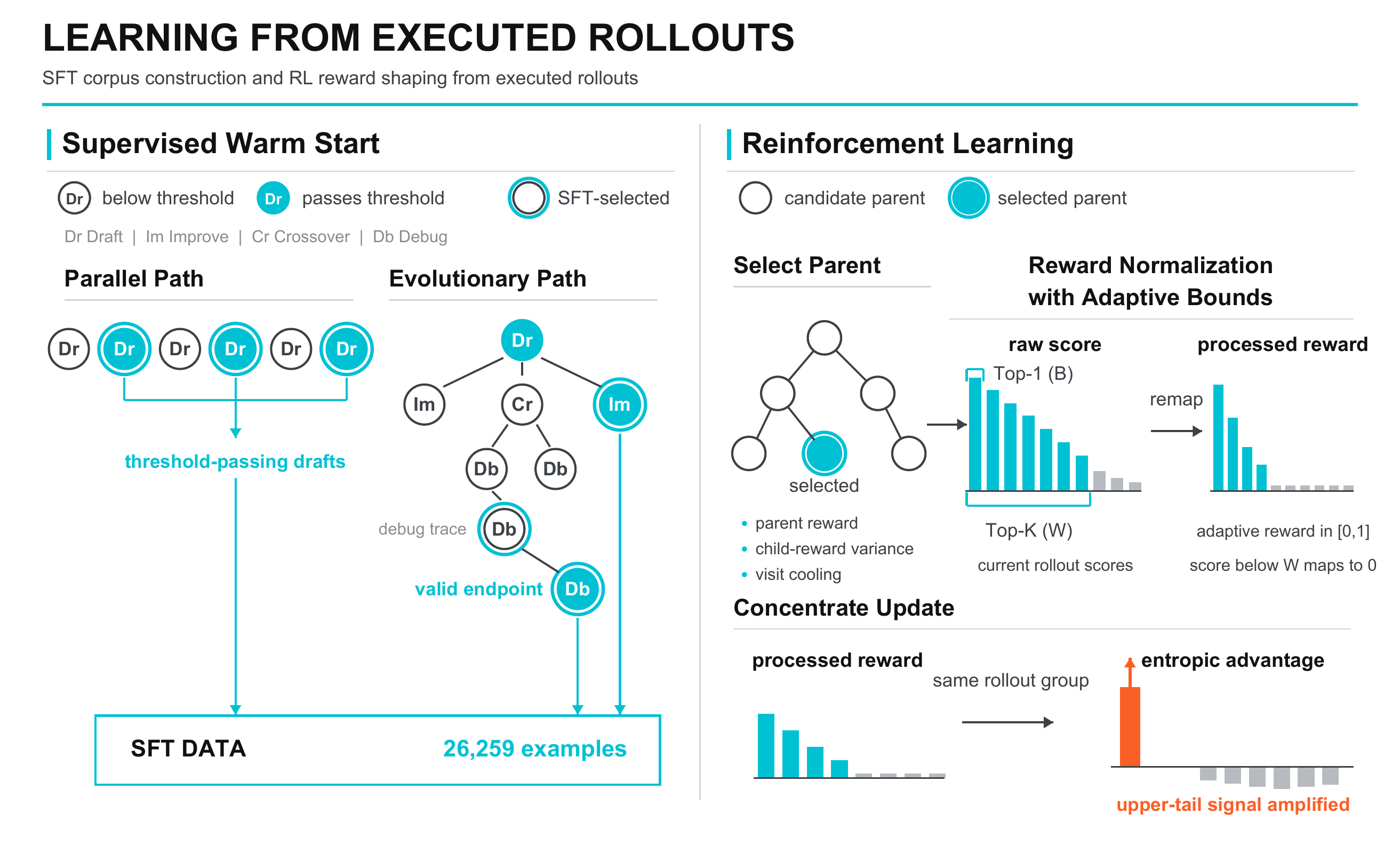}
  \caption{Learning from executed rollouts. The Parallel Path retains threshold-passing \textsc{Draft} solutions. In the Evolutionary Path, a valid endpoint can emerge only after repeated \textsc{Debug} steps; we trace back to the preceding non-debug operator and use an LLM to retain useful steps from that repair trace. The selected examples from both paths form the 26,259-example released SFT corpus under a budget-adaptive stopping rule. For a chosen operator, RL selects a parent using parent reward, child-reward variance, and visit cooling. Top-1/Top-$K$ adaptive bounds map seven nonzero task scores to four nonzero processed rewards in the illustrative rollout group, with scores below the resolved lower bound clipped to zero; entropic advantages then amplify the upper-tail learning signal.}
  \label{fig:sft-rl-rollout}
\end{figure}

Building on the supervised warm start, the reinforcement-learning portion of Figure~\ref{fig:sft-rl-rollout} previews how state selection, adaptive reward normalization, and upper-tail weighting act on an evolutionary rollout.

\textbf{Making heterogeneous outcomes comparable.}
One task may optimize accuracy and another log loss; even after aligning their directions, raw score ranges remain incomparable. Let $\tilde{s}$ denote a raw score converted to the convention that larger is better. We first define a bounded \emph{base reward} from a fixed pair of task bounds:
\begin{equation}
\label{eq:base-reward}
r_{\mathrm{base}}(\tilde{s}; b_{\mathrm{best}}, b_{\mathrm{worst}}) =
\mathrm{clip}\left(\frac{\tilde{s} - b_{\mathrm{worst}}}{b_{\mathrm{best}} - b_{\mathrm{worst}}}, 0, 1\right)^\alpha,
\qquad \alpha > 0.
\end{equation}
Fixed bounds establish a common interval, but leaderboard or theoretical extrema can be much wider than the score region reached by the current policy. In that case, meaningfully different programs collapse to nearly identical rewards. OpenMLE therefore derives tighter adaptive bounds from each task's historical on-policy score frontier and remaps $\tilde{s}$ to a processed reward $r_{\mathrm{proc}}$. The bounds evolve with the policy, preserving resolution where current candidates actually lie while retaining a stable reward direction. This follows the broader lesson from adaptive verifiable learning environments and evolving-rubric training that the evaluator scale must preserve discriminative on-policy feedback~\citep{zeng2025rlve,shao2025dr}. Appendix~\ref{app:reward-advantage} specifies the bound construction and edge-case handling.

\begin{figure}[!t]
  \centering
  \begin{minipage}[t]{0.425\linewidth}
    \centering
    \includegraphics[width=\linewidth]{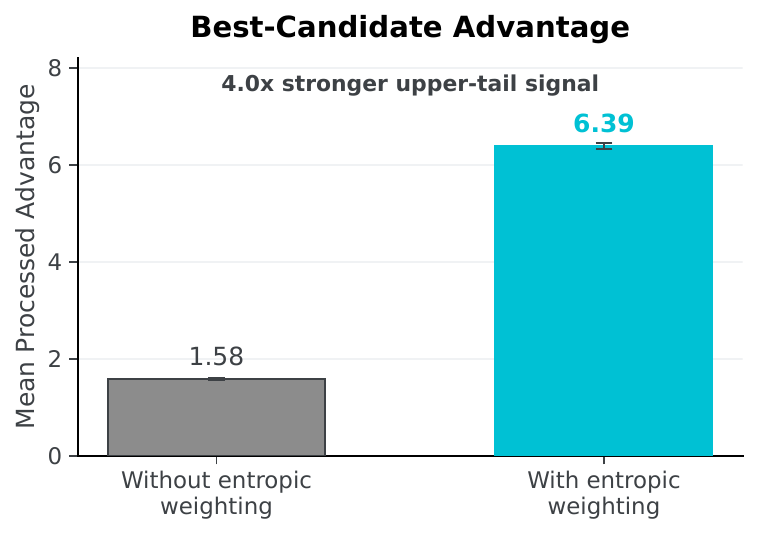}\\[-0.5ex]
    \small (a) Entropic advantage signal
  \end{minipage}
  \hfill
  \begin{minipage}[t]{0.55\linewidth}
    \centering
    \includegraphics[width=\linewidth]{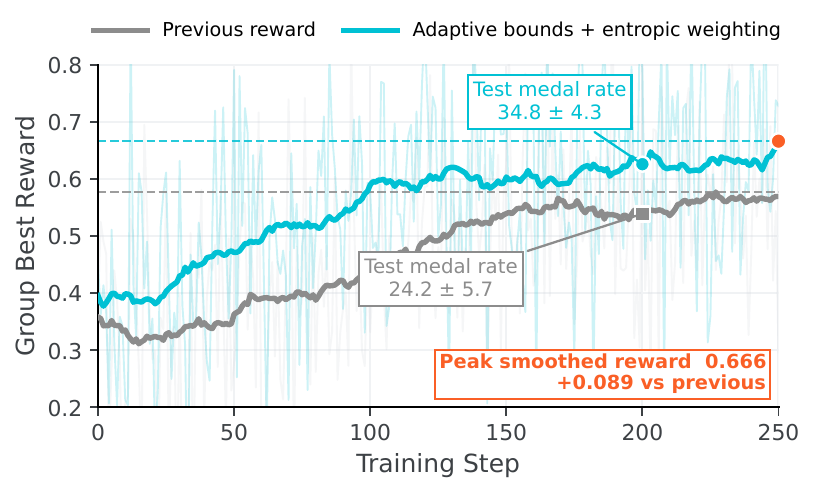}\\[-0.5ex]
    \small (b) Group Best Reward
  \end{minipage}
  \caption{Effect of upper-tail reward shaping. (a) Entropic weighting increases the processed advantage assigned to the best candidate in a rollout group. (b) Combining entropic weighting with adaptive bounds yields a stronger smoothed Group Best Reward trajectory than the previous reward construction. The two test medal rates in panel (b) use a simpler early-stage harness rather than \openmleevo{}.}
  \label{fig:entropic-advantage-effect}
\end{figure}

\textbf{Concentrating learning signal on the upper tail.}
After scores become comparable, the remaining question is how strongly each candidate should affect the update. MLE evaluation rewards the quality of the best program found; a barely viable submission should therefore not receive the same positive reward as a top-performing one. OpenMLE uses an entropic advantage that amplifies reward gaps near the top of each rollout group, following the upper-tail principle studied in TTT-Discover~\citep{yuksekgonul2026learning} and related Best-of-$N$/Pass@$k$ objectives~\citep{walder2025passkpolicy,chen2025passk,bagirov2025bestofn,peng2025simko}. Omitting the stabilizing max-centering used in implementation, the transform is
\begin{equation}
\label{eq:adv}
A_i^{\mathrm{ent}}\approx
\frac{\exp(\beta r_{\mathrm{proc},i})}
{\frac{1}{G-1}\sum_{j\ne i}\exp(\beta r_{\mathrm{proc},j})}
-1,
\end{equation}
where $\beta$ controls concentration and is selected under a fixed entropy/KL budget. These advantages replace the usual GRPO-style group-normalized signal in the clipped policy objective. Adaptive bounds first make within-group differences visible; entropic weighting then directs substantially more learning signal to the best candidates rather than uniformly reinforcing all non-failing programs. Figure~\ref{fig:entropic-advantage-effect} shows both the resulting best-candidate emphasis and its observed effect during training. Exact post-processing formulas appear in Appendix~\ref{app:reward-advantage}.

\textbf{Removing stragglers with asynchronous rollouts.}
Unlike token-level verification, the dominant latency in MLE RL comes from executing candidate programs, and runtimes vary substantially across tasks and solutions. In a synchronous batch, completed groups remain idle until its slowest sandbox job returns. OpenMLE instead launches generation-and-execution groups independently and lets the trainer consume each completed group from a queue. This decouples policy updates from the longest job in a nominal batch while preserving group-level advantages. Because the realized speedup depends on the task-runtime distribution and worker allocation, we keep the main-text claim qualitative and report the measured timing study in Appendix~\ref{app:async-rollouts}.

\textbf{Selecting informative states for operator learning.}
Evolutionary RL must choose not only a task and an operator, but also the program state on which that operator acts. Uniform parent sampling spends updates on exhausted or uninformative regions; greedy sampling repeatedly trains on the current incumbent and suppresses diversity. After selecting the operator to practice, OpenMLE samples parent programs with a fitness-proportional utility that combines three terms:
\begin{equation}
\label{eq:sample}
F(p) = \mathrm{norm}(R_p) + \mathrm{norm}(\mathrm{Var}_{c \in \mathrm{child}(p)} R_c) + \mathrm{norm}(C_p),
\end{equation}
Here $R_p$ favors strong parent programs, child-reward variance $\mathrm{Var}_{c \in \mathrm{child}(p)} R_c$  identifies regions where operator outcomes remain informative, and $C_p$ is a cooling coefficient that decreases with repeated visits. The first term exploits promising solutions, the second targets states with unresolved learning signal, and the third prevents a single incumbent from monopolizing the rollout budget. This selection rule exposes \textsc{Improve}, \textsc{Debug}, and \textsc{Crossover} to useful yet diverse local contexts; Appendix~\ref{app:evolutionary-fitness} gives the exact implementation.

\section{OpenMLE-Evo: Scaling Experience-driven Long-Horizon Search}
\label{sec:tts-harness}

\begin{takeaway}
\begin{itemize}[leftmargin=*,itemsep=1pt,topsep=2pt,partopsep=0pt,parsep=0pt]
\item \textbf{Test-time scaling becomes test-time learning when search learns from experience.} For AI-for-AI and recursive self-improvement, generating more candidates is not enough: the search process must convert execution outcomes into reusable evidence that changes what it explores next. Evolutionary test-time search provides this closed loop---propose, execute, learn from experience, and adapt future expansion over long horizons.
\item \textbf{Experience-driven evolution hinges on two coupled scientific questions.} Which node should be expanded next, beyond greedy score maximization? And how should memory be constructed so that the selected operator receives actionable evidence rather than an ever-growing trace? \openmleevo{} addresses the first with non-greedy, multi-factor selection over quality, progress, and novelty, and the second with operator-conditioned, on-demand synthesis from bounded relevant experience.
\end{itemize}
\end{takeaway}

\begin{figure}[t]
    \centering
    \includegraphics[width=\linewidth]{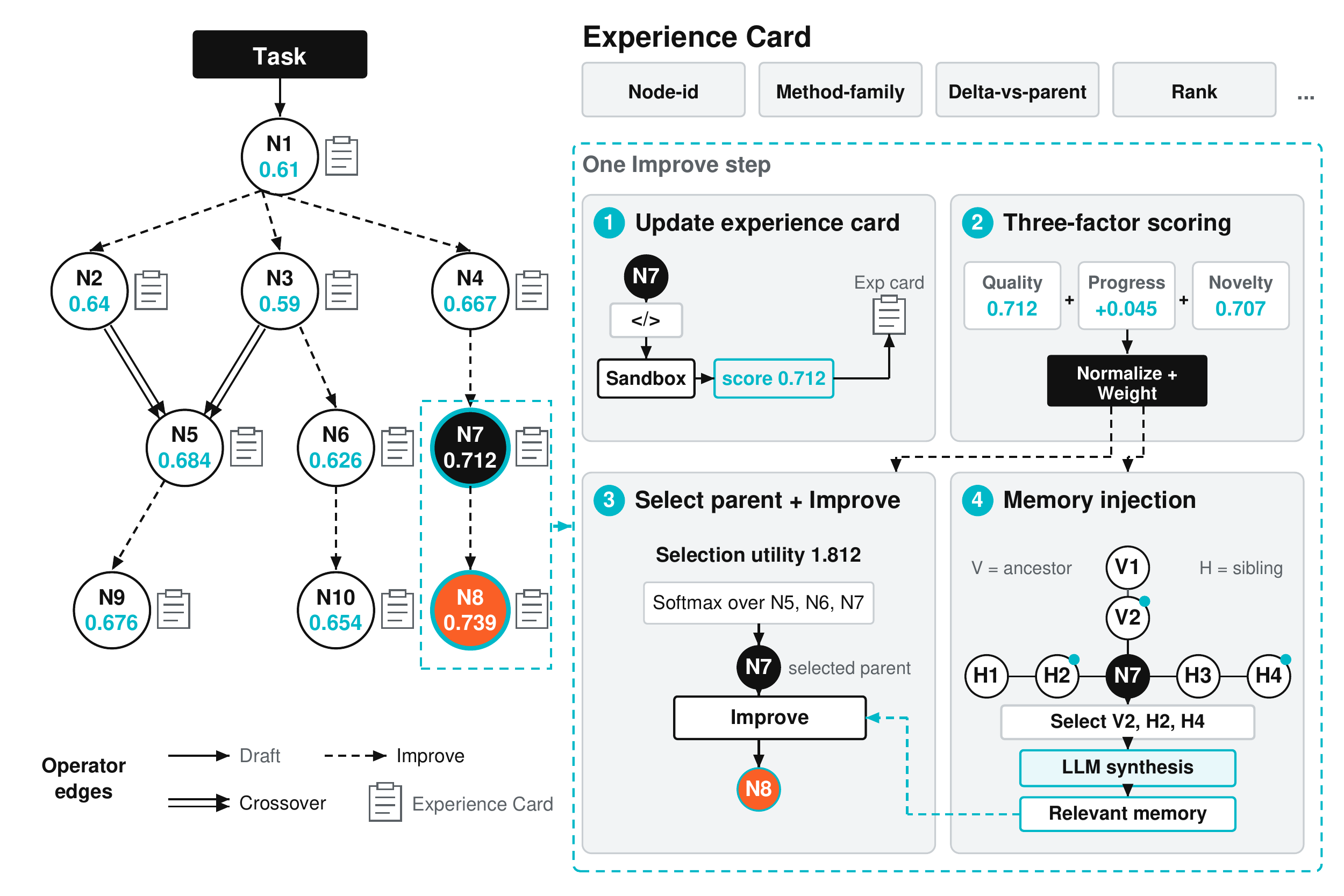}
    \caption{\textbf{OpenMLE-Evo Search Harness.} \textbf{Left:} The search tree expands candidate solutions through drafting, improvement, and crossover operations, with each evaluated node paired with a structured experience card. \textbf{Right:} Experience-card metadata is used to update the global search state, compute parent-selection weights from quality, progress, and novelty, select the next parent, and retrieve relevant memories from key ancestors and siblings for the subsequent improvement operation.}
    \label{fig:openmle-evo-experience-guided-search}
\end{figure}

\textbf{Motivation.}
AI systems that improve AI artifacts must use execution outcomes to determine what to try next, rather than merely sample more candidates. Evolutionary test-time search operationalizes this feedback loop by maintaining executable programs, selecting parents, applying transformations, and evaluating the resulting candidates. Over long horizons, effective search therefore requires three capabilities: a persistent and queryable representation of prior attempts, compute allocation across high-quality, improving, and novel branches, and bounded context tailored to the transformation being applied.

AIDE, AIRA, and AIRA$_2$ establish iterative search over executable programs through tree- or population-based exploration, repeated execution, and candidate refinement~\citep{jiang2025aide,toledo2025airesearchagents,hambardzumyan2026aira_2}. \openmleevo{} adopts an AIRA-Evo-style population loop to compose OpenMLE's trained \textsc{Draft}, \textsc{Improve}, \textsc{Debug}, and \textsc{Crossover} operators, but redesigns how the loop uses execution evidence. Standard AIRA-Evo stores largely free-form memory, synthesizes it eagerly, selects parents primarily by scalar fitness, and supplies different operators with broadly similar histories. \openmleevo{} instead stores structured experience records, selects parents by quality, progress, and novelty, and synthesizes bounded memory on demand for the operator being invoked. The details of each component are described in the following sections, while the overall experience-guided search framework is illustrated in Figure~\ref{fig:openmle-evo-experience-guided-search}.

\subsection{Structured Experience Accumulation}
\openmleevo{} accumulates search experience at two complementary levels. First, after a candidate is evaluated in the sandbox, the harness creates a node-level \textbf{experience card}. Its core metadata is extracted deterministically from the search state and execution result, capturing the candidate’s provenance, performance, execution outcome, and resource usage. The full schema details and a grounded record are provided in Appendix~\ref{app:openmleevo-experience-records}. This gives every node a compact and consistently structured record of both the attempted change and its observed outcome. Second, the harness aggregates the cards from all evaluated nodes into a task-global experience board. The board maintains population-level statistics such as explored method families, family-wise best candidates, underexplored directions, repeated failures, score trends, and the parent graph. It therefore exposes the state of the surrounding search neighborhood, allowing a newly selected node to understand not only its own history but also how its ancestors, siblings, and related method families have performed. Together, the card and board prevent node-level signals from being lost in the expanding search space. This queryable deterministic state supplies quality, progress, and novelty to parent selection, and lineage, neighborhood, failure, and resource evidence to operation-conditioned retrieval and on-demand memory synthesis.

\subsection{Experience-Guided Parent Selection}

Original AIRA-Evo derives parent-sampling probabilities primarily from
normalized fitness. Consequently, although parent selection remains
stochastic, the search is driven almost entirely by a node's current
validation score and tends to concentrate expansion on already strong
nodes. Other informative signals, such as how much a node improves upon
its ancestry or whether it introduces a previously underexplored solution
family, are not explicitly considered.

In \openmleevo{}, we instead transform the deterministic metadata stored
in each experience card into three complementary factors: normalized
validation score $\tilde{s}_i$, normalized positive improvement over the
strongest parent $\widetilde{\Delta}_i$, and method-family novelty $\nu_i$.
For candidates $i \in \mathcal{I}$ in a sampled island, we define an
experience-guided utility and sample the next parent according to
\begin{equation}
U_i
=
\lambda_s \tilde{s}_i
+
\lambda_{\Delta} \widetilde{\Delta}_i
+
\lambda_n \nu_i,
\qquad
P(i \mid \mathcal{I})
=
\frac{\exp(U_i/\tau)}
{\sum_{j \in \mathcal{I}} \exp(U_j/\tau)}.
\label{eq:openmleevo-parent-selection}
\end{equation}

Thus, each parent-selection decision jointly considers three aspects of a
candidate: its current solution quality, the progress it achieved relative
to its lineage, and the algorithmic novelty of the direction it represents.
This produces a more comprehensive expansion policy: it preserves selection
pressure toward high-quality solutions while still allocating search budget
to candidates that demonstrate meaningful progress or introduce promising,
underexplored approaches. Detailed factor definitions and sampling procedures are provided in
Appendix~\ref{app:openmleevo-parent-selection}.

\subsection{Operation-Triggered Memory Synthesis}
The original AIRA-Evo memory path eagerly invokes a language model to summarize the history of every evaluated node by default. This spends inference budget on nodes that are never selected by a later operator, while producing a summary before the decision context that should shape it is known. \openmleevo{} separates deterministic storage from language-model synthesis: after sandbox evaluation, it preserves the experience card and experience board, but defers rich natural-language memory until an \textsc{Improve}, \textsc{Crossover}, or \textsc{Debug} call has selected its relevant nodes. It then invokes the memory model only for the selected parent(s) and their retrieved ancestors, siblings, or error-related attempts, and caches the resulting method and parent-comparison summaries. This on-demand policy avoids unnecessary calls over long search trajectories, improves the efficiency of LLM-based experience extraction, and lets an optimized operation-aware extraction template produce more relevant, higher-quality memory. Prompt templates and representative memory records are deferred to Appendix~\ref{app:openmleevo-memory-synthesis}.

\subsection{Operator-Conditioned Context Construction}
Once a parent has been sampled, \openmleevo{} constructs a small, operator-conditioned context instead of appending the full free-form history. For \textsc{Improve}, it concatenates the selected node's deterministic experience record including its validation score, improvement over its parent, method family, runtime, rank, incumbent status, and direction novelty with a vertical trace of its recent ancestors and a horizontal set of direct siblings, namely prior candidates sharing at least one parent. The sibling set is ranked by the same score--improvement--novelty utility used for parent selection, and only the most informative siblings are retained; the operator can therefore contrast the chosen trajectory with nearby alternatives rather than unrelated programs elsewhere in the search. A global experience board, recomputed from all accumulated cards, further supplies the current best method family, family-level success and failure statistics, underexplored directions, recent improvement trends, and recurring error signatures. \textsc{Crossover} applies this construction separately to both parents and adds a method-family complementarity cue, whereas \textsc{Debug} retrieves prior attempts with the same error signature, falling back to recent attempts when exact matches are unavailable. Concise method and parent-comparison summaries are generated lazily for the retrieved nodes and cached, while the core retrieval signals remain deterministic. This branch-local, bounded context avoids the redundancy of an ever-growing history and gives each operator evidence that is directly actionable for refinement, recombination, or repair. The resulting prompt also specifies the remaining search budget, remaining steps, and per-run execution limit, so that these decisions remain feasible under the task's actual computational constraints; Appendix~\ref{app:openmleevo-memory-synthesis} specifies the retrieval sets and generated fields.

%% file: sections/06_main_results.tex
\section{Experiments}
\label{sec:experiments}

\subsection{Experimental Setup}
\label{sec:experimental-setup}

\paragraph{Experimental setup.}
We evaluate on the official 22-task MLE-Bench Lite split released with MLE-Bench unless otherwise stated~\citep{chan2024mlebench}.
Unless specified otherwise, each OpenMLE-Evo configuration is evaluated with three independent runs under a fixed per-task budget of 12 hours on a single RTX 4090 (12 GB VRAM)---a smaller per-task sandbox-compute budget than that used by the vast majority of reported evaluations on MLE-Bench.\footnote{This comparison uses the per-run evaluation compute configurations---accelerator allocation and wall-clock budget---reported in the \href{https://github.com/openai/mle-bench/tree/main/runs\#readme}{official MLE-Bench runs registry}. It does not compare model-inference cost or normalize different accelerators to FLOPs.}
We report three aggregate metrics. \emph{Valid Rate} is the mean number of the 22 tasks for which a run produces a valid submission, written as $x/22$; \emph{Medal Average} is the mean fraction of tasks receiving any Kaggle medal; and \emph{Human Rank} is the fraction of human leaderboard participants whose score is surpassed by the submitted solution, averaged across tasks and runs, so higher is better.
Our primary model is \thirtyfivebmodel{}, which is used for the headline model, system, trajectory, and transfer analyses. We additionally evaluate \thirtybmodel{} as a companion model to test whether the post-training gain reproduces on a second backbone and model scale.

\paragraph{OpenMLE-Evo-Max.}
\openmleevomax{} extends the \openmleevo{} configuration described in Section~\ref{sec:tts-harness} in two ways.
First, it uses a general pipeline to distill reusable cross-task priors from public competition artifacts; all MLE-Bench-related sources are excluded before distillation.
Second, it enables asynchronous multi-GPU parallel search while keeping the total sandbox compute budget unchanged, following insights from AIRA$_2$~\citep{hambardzumyan2026aira_2}.

\subsection{Training and Search Gains Compose}
\label{sec:main-results-lite}

\begin{takeaway}
Under the identical \openmleevo{} harness, execution-grounded post-training
improves the Medal Average of our primary \thirtyfivebmodel{} over its
corresponding Qwen3.6 backbone by $21.22$ percentage points; the companion
\thirtybmodel{} reproduces this gain at $18.18$ percentage points over its
Qwen3 backbone. Combining \thirtyfivebmodel{} with \openmleevomax{} further
reaches $71.21\%$, exceeding GPT-5.5 with Codex by $3.03$ percentage points
and showing that training and search provide complementary improvements.
\end{takeaway}

\input{figure_source/main_exp_results_table.tex}

\paragraph{Primary 35B model and system results.}
Under the identical standard \openmleevo{} harness, execution-grounded post-training improves \thirtyfivebmodel{} over its Qwen3.6-35B-A3B base from $39.39\%$ to \frontisThirtyFiveEvoResult{} Medal Average.
Under this common harness, \thirtyfivebmodel{} also outperforms MiniMax M3, Doubao Seed 2.1 Pro, and DeepSeek-V4-Pro, showing that the post-trained model is competitive beyond its base-model comparison.

After injecting MLE-Bench-disjoint cross-task priors and widening parallel tree search through \openmleevomax{}, \thirtyfivebmodel{} reaches \frontisThirtyFiveEvoMaxResult{}, exceeding GLM-5.2 and MiniMax M3 under the same enhanced harness.
The resulting model--harness system also surpasses GPT-5.5 paired with Codex, demonstrating that the training and search gains compose at the system level. To ensure the robustness of these results, we additionally report the mean and standard deviation across three evaluation runs in Appendix~\ref{app:openmle-repeat-evaluation}.

\paragraph{Cross-model replication on 30B.}
We separately apply the same post-training approach to \thirtybmodel{}. Under the identical standard \openmleevo{} harness, it improves over Qwen3-30B-A3B-Thinking-2507 from $34.85\%$ to \frontisThirtyEvoResult{} Medal Average; with \openmleevomax{}, it further reaches \frontisThirtyEvoMaxResult{} Medal Average and \frontisThirtyEvoMaxHumanRank{} Human Rank. This controlled comparison provides supporting evidence that the post-training and search gains are not confined to the primary 35B checkpoint.
\begin{figure}[!t]
\centering
\includegraphics[width=\linewidth]{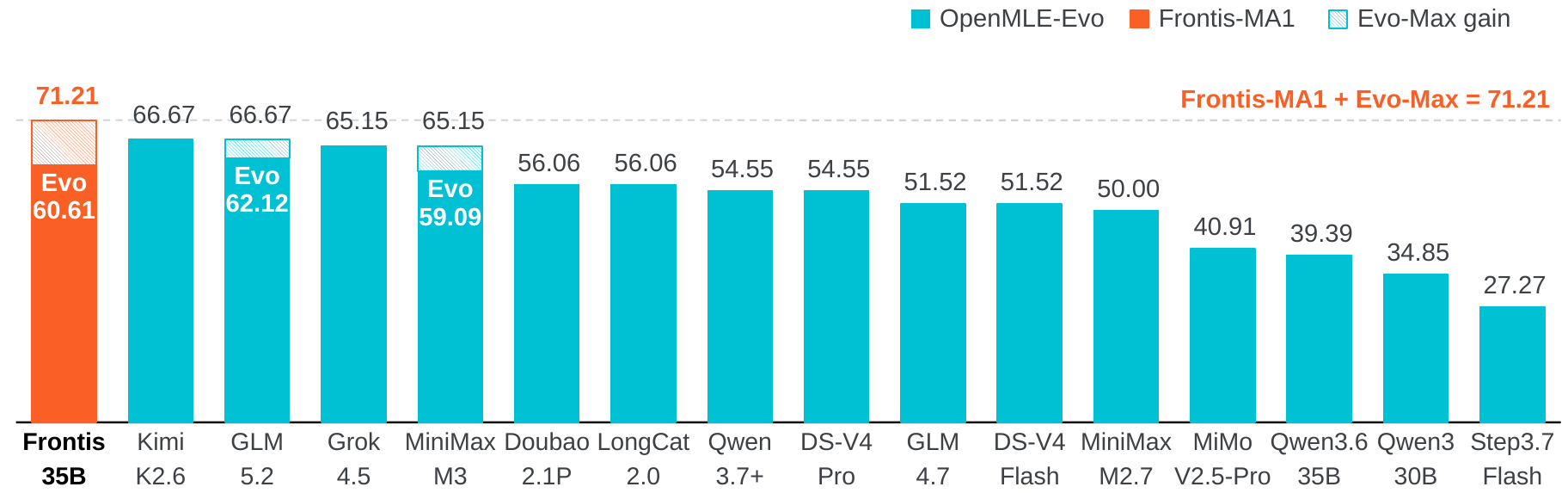}
\caption{Model performance on MLE-Bench Lite under the common \openmleevo{} harness. Solid bars report standard \openmleevo{} results, while hatched caps show the additional gains from \openmleevomax{} for \thirtyfivebmodel{}, GLM-5.2, and MiniMax M3.}
\label{fig:model-exp}
\end{figure}

\paragraph{Harness gains from domain-specific search.}
\label{sec:harness-comparison}

\begin{figure}[!t]
\centering
\includegraphics[width=\linewidth]{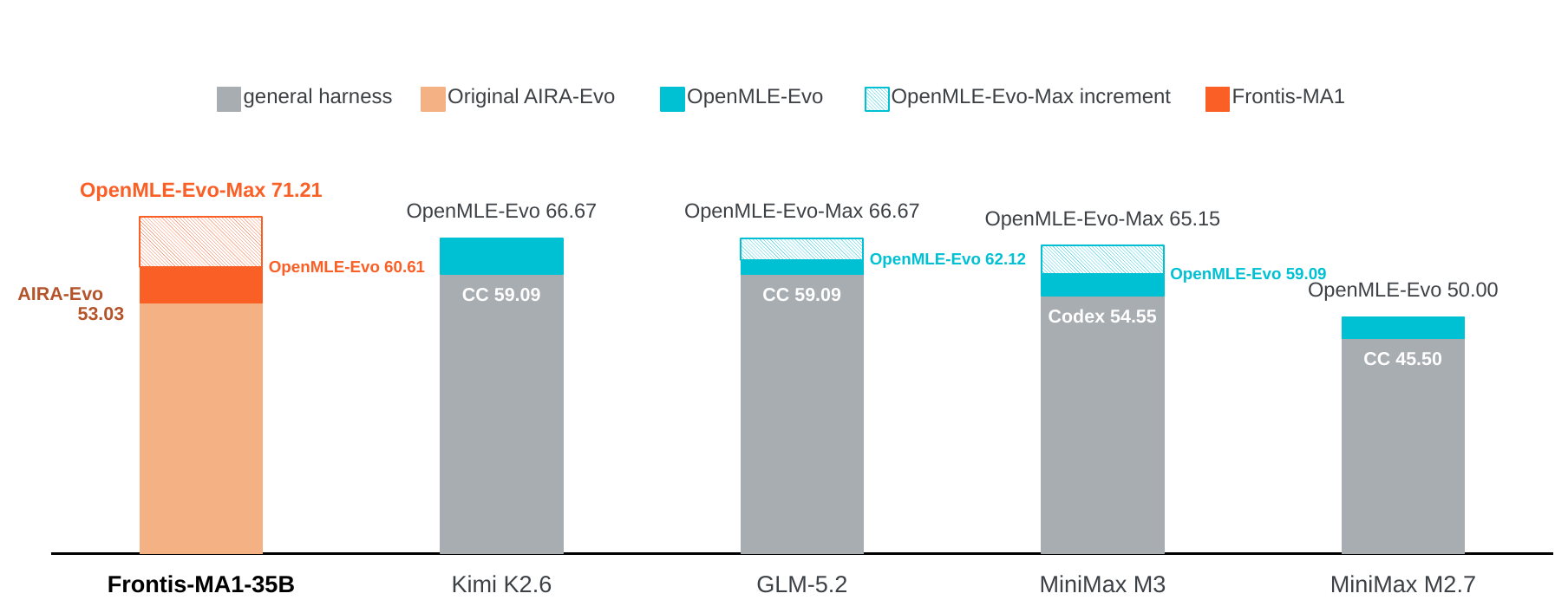}
    \caption{Harness comparison on MLE-Bench Lite. Gray reports the general-purpose Claude Code or Codex result, light orange reports original AIRA-Evo for the matched \thirtyfivebmodel{} comparison, cyan reports \openmleevo{}, and hatched caps show the additional \openmleevomax{} gain when available. The OpenMLE model is highlighted in saturated orange.}
\label{fig:harness-exp}
\end{figure}

Figure~\ref{fig:harness-exp} compares harnesses while holding the underlying model fixed.
Across multiple model families, \openmleevo{} consistently converts the same model into a stronger MLE system than general-purpose coding-agent harnesses such as Claude Code and Codex.
Against original AIRA-Evo, the matched \thirtyfivebmodel{} comparison increases Medal Average from $53.03\%$ to \frontisThirtyFiveEvoResult{} under \openmleevo{}.
The consistency of this gain indicates that the improvement comes from search specialized for iterative machine learning engineering, instead of a single favorable model–harness pairing.

\subsection{Long-Horizon Self-Improvement}
\label{sec:search-dynamics-analysis}

\begin{takeaway}
\openmleevo{} continues to improve well beyond the first executable solution. Later \textsc{Improve} and \textsc{Crossover} operations turn accumulated experience into decisive performance gains. This shows that experience-guided long-horizon search converts additional test-time compute into sustained progress rather than redundant sampling.
\end{takeaway}

\begin{figure*}[t]
  \centering
  \includegraphics[width=1.0\textwidth]{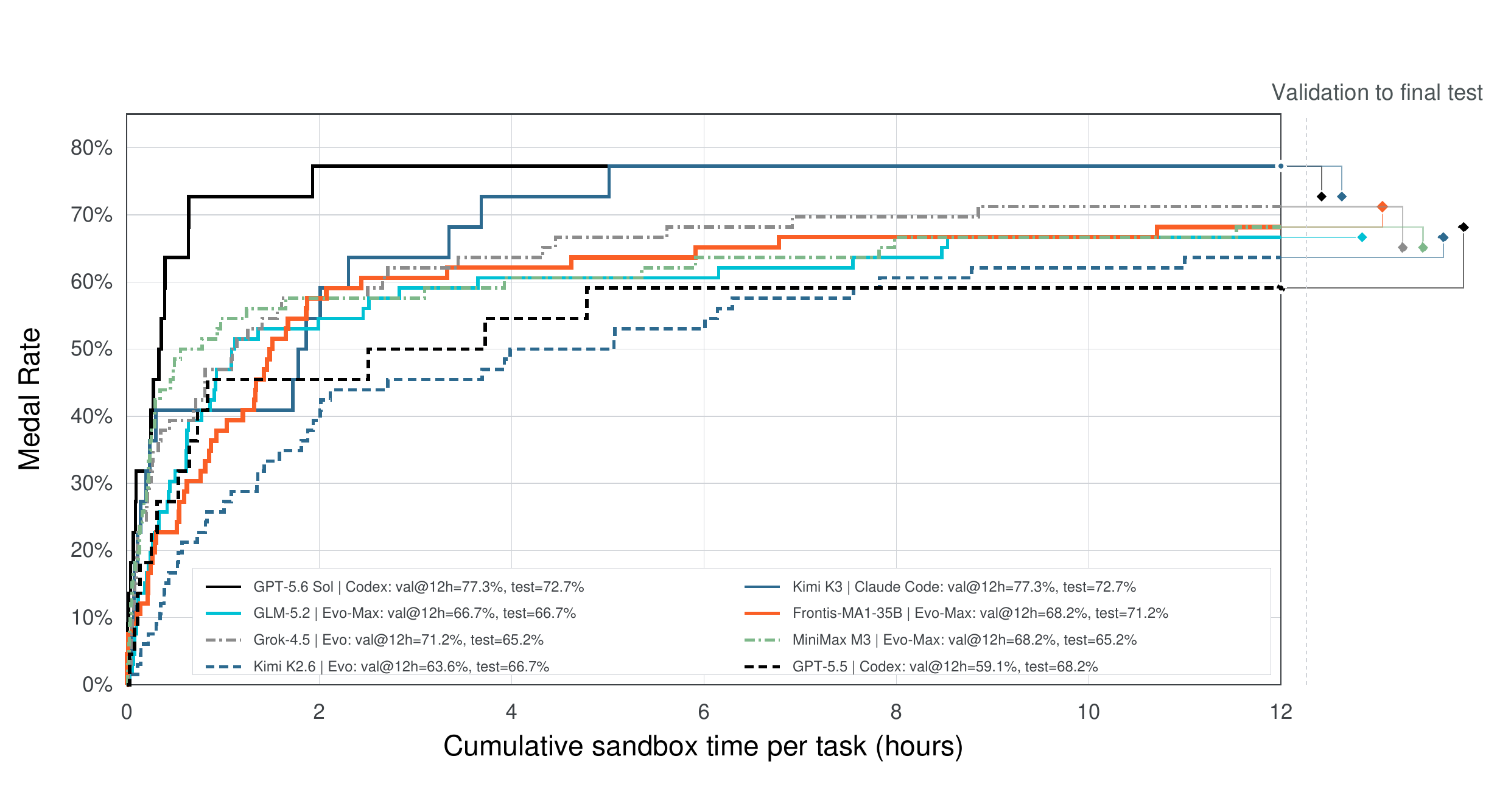}
  \caption{Cross-harness Medal Rate evolution over the 22-task MLE-Bench Lite subset for the top 8 systems. Each step curve reports the fraction of task outcomes whose best-so-far validation score has reached any Kaggle medal by the indicated cumulative sandbox time. The separate final-test panel reports Medal Rate after validation-based node selection, and colored bridges expose the validation-to-test change for each system.}
  \label{fig:case-study-medal-rate-evolution}
\end{figure*}

\textbf{Aggregate long-horizon improvement.}
Figure~\ref{fig:case-study-medal-rate-evolution} shows that combining \thirtyfivebmodel{} with \openmleevomax{} yields sustained improvement throughout long-horizon search. The resulting solutions generalize strongly to the final test, achieving a $71.21\%$ Medal Rate, compared with $68.18\%$ on validation. This performance is comparable to GPT-5.6 Sol with xhigh reasoning and Kimi K3 with their respective harnesses, both of which achieve $72.73\%$ on the final test.

\begin{figure}[htbp]
  \centering
  \includegraphics[width=1.0\textwidth]{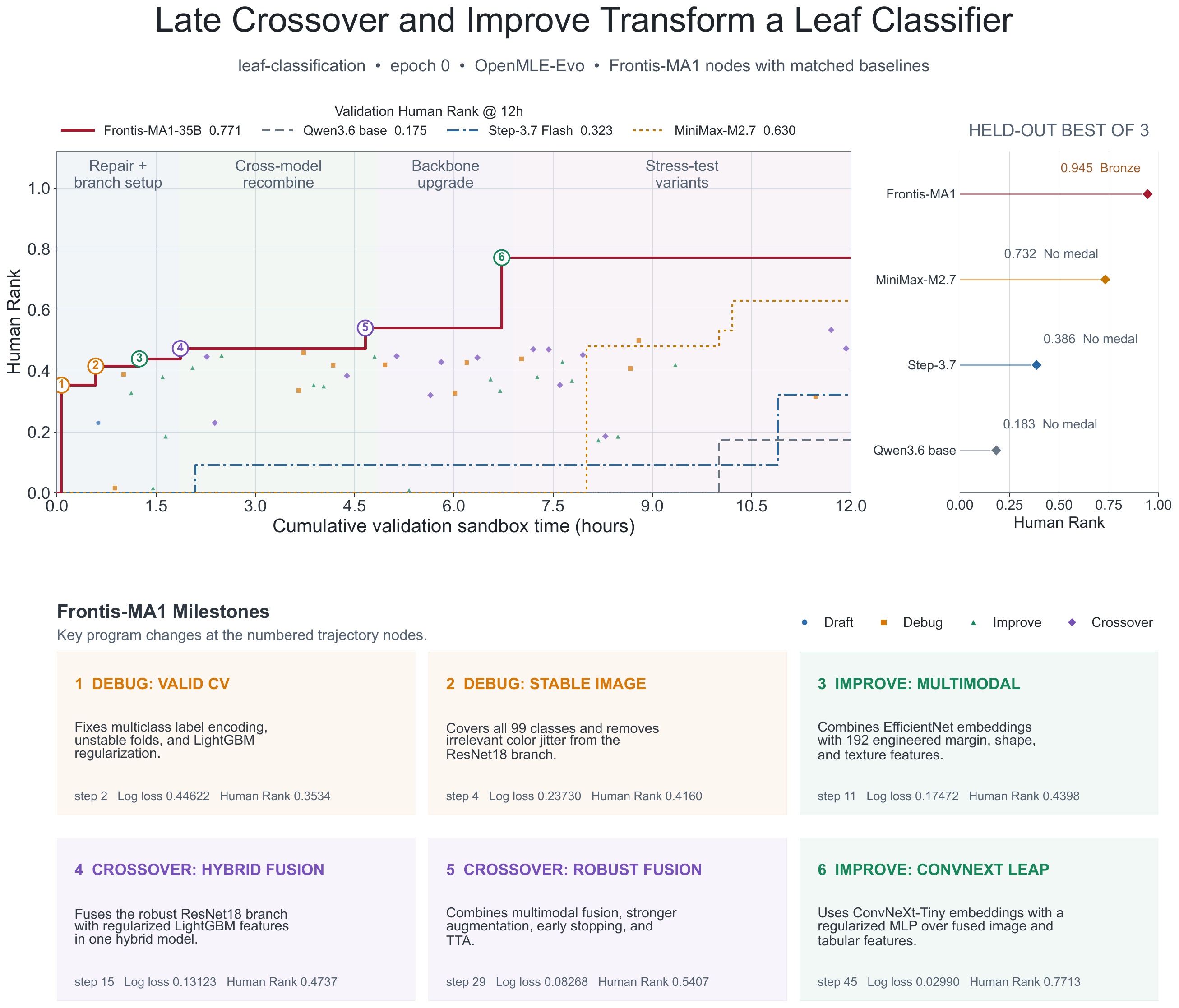}
  \caption{Cross-model search trajectories on \textit{leaf-classification}. The left panel compares best-so-far validation-derived Human Rank over 12 cumulative sandbox hours for the same epoch under the common \openmleevo{} protocol; markers and numbered operation cards expose only the detailed \thirtyfivebmodel{} trajectory. Its two Crossovers establish and strengthen multimodal image--tabular fusion, and a late Improve operation produces the largest jump by upgrading the fused representation to ConvNeXt-Tiny. The separate right panel reports held-out best-of-three final-test Human Rank, where the selected \thirtyfivebmodel{} node obtains Bronze.}
  \label{fig:case-study-leaf-late-ops}
\end{figure}

\textbf{Structured recombination turns additional search into better ML models.}
Figure~\ref{fig:case-study-leaf-late-ops} shows that the main difference is not merely finding an executable solution, but continuing to improve its structure. Under the same \openmleevo{} protocol, the comparison models either plateau at low Human Rank or improve without reaching a medal-quality held-out solution. In contrast, \thirtyfivebmodel{} first uses \textsc{Debug} to establish viable image and tabular branches, then uses \textsc{Crossover} to preserve their complementary evidence and \textsc{Improve} to upgrade the fused model only after that design is stable. These latter operations produce 85.0\% of the total validation gain, indicating that the long horizon is spent accumulating and recombining useful branch evidence, as opposed to repeatedly repairing a single program.
 The resulting trajectory reaches validation Human Rank $0.7713$ and held-out Human Rank $0.9455$ with Bronze; the strongest comparison reaches only $0.6303$ on validation and earns no medal. The lead also holds in the three-epoch averages, making the mechanism less consistent with a single lucky trajectory.

\begin{figure}[!t]
  \centering
  \includegraphics[width=1.0\textwidth]{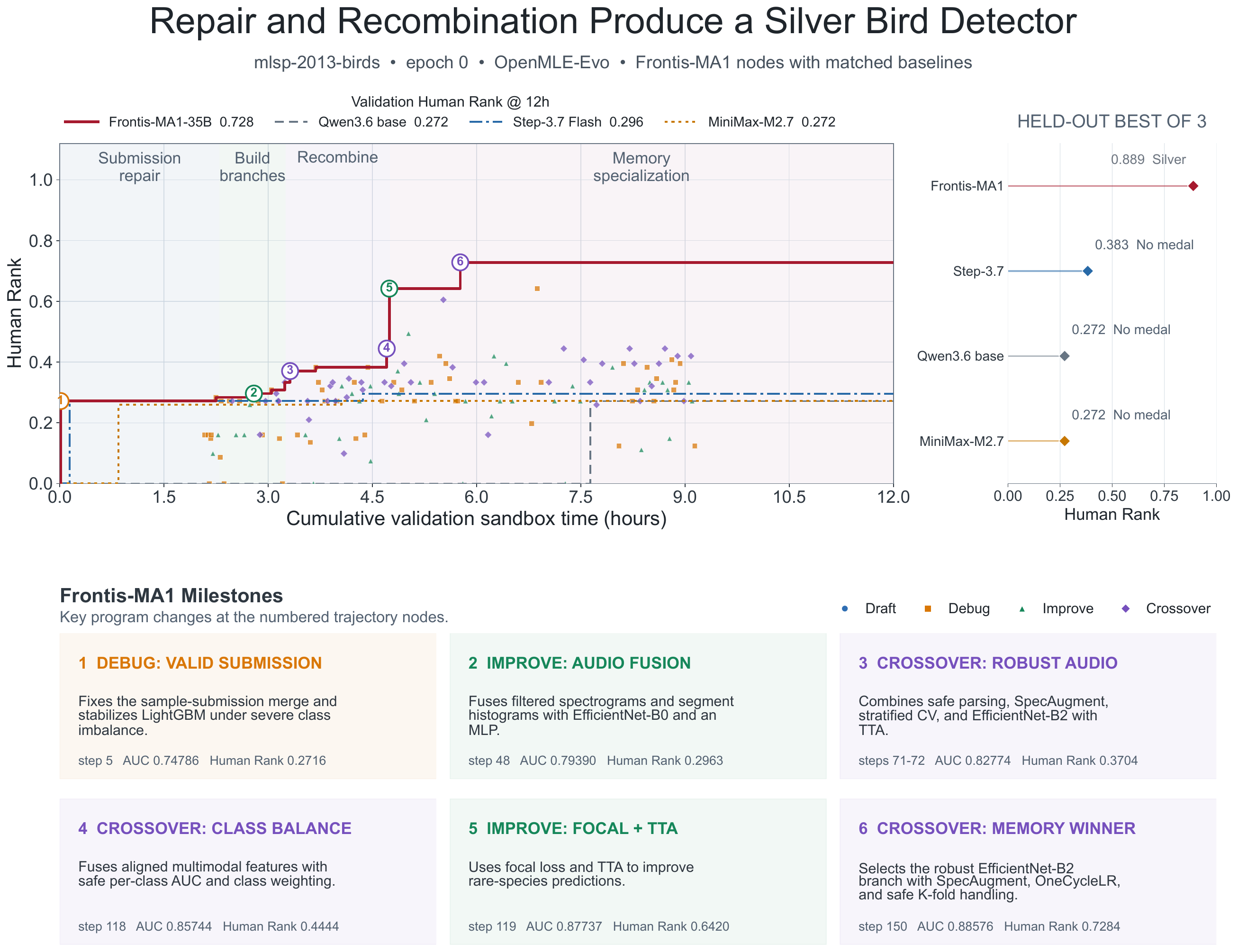}
  \caption{Cross-model search trajectories on \textit{mlsp-2013-birds} using the same corrected task prompt. The left panel compares best-so-far validation-derived Human Rank over 12 cumulative sandbox hours under the common \openmleevo{} protocol; markers and numbered operation cards expose only the detailed \thirtyfivebmodel{} trajectory. Early Debug makes the submission executable, while later Improve and Crossover operations build progressively stronger multimodal audio branches. The separate right panel reports held-out best-of-three final-test Human Rank, where the final Memory-guided Crossover obtains Silver.}
  \label{fig:case-study-mlsp-operation-chain}
\end{figure}

\textbf{Memory-guided recombination breaks the search plateau.}
In figure~\ref{fig:case-study-mlsp-operation-chain}, the comparison trajectories remain near their first viable solutions, whereas \thirtyfivebmodel{} treats submission repair as a starting point and subsequently builds specialized audio branches. Here memory matters through selection, not volume: it preserves which branches contributed robust parsing, imbalance handling, augmentation, and representation quality, while marking an inferior ResNet50 direction as evidence to avoid. \textsc{Improve} and \textsc{Crossover} can therefore combine compatible gains instead of inheriting an undifferentiated history; together they account for $91.9\%$ of the total validation improvement. This produces validation Human Rank $0.7284$ and held-out Human Rank $0.8889$ with Silver, while the strongest comparison reaches only $0.2963$ in validation and none earns a medal. The advantage remains after averaging all three epochs, and the shared corrected prompt rules out the earlier submission-contract ambiguity as its explanation.

\subsection{Solution Ceiling}
\label{sec:solution-ceiling}

\begin{takeaway}
Within matched model and search comparisons, the gains are not limited to moving more solutions across the Bronze threshold: stronger systems also produce a larger share of Gold solutions.
\end{takeaway}

\begin{figure}[htbp]
  \centering
  \includegraphics[width=1.0\textwidth]{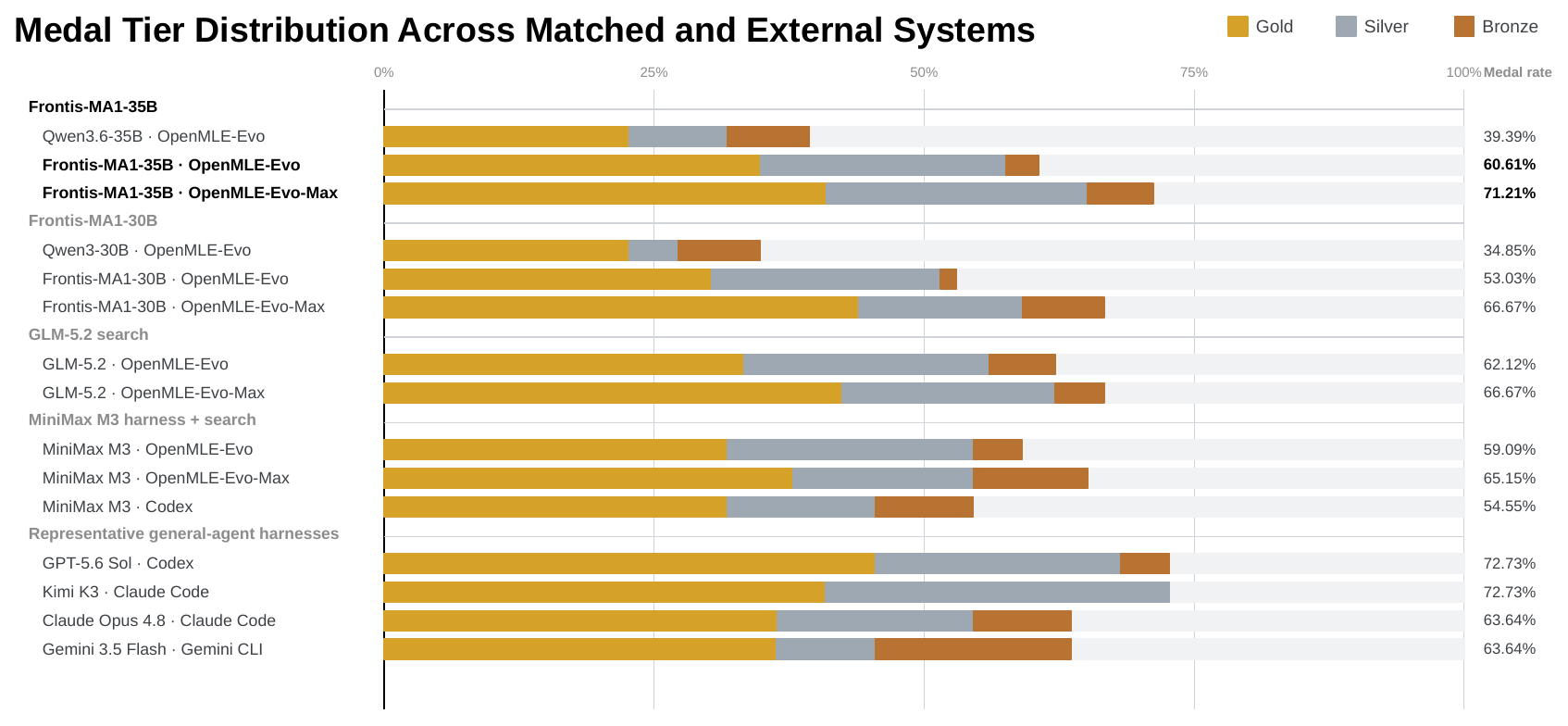}
  \caption{Gold, Silver, and Bronze decomposition for matched OpenMLE comparisons and representative general-agent harnesses. Bar lengths report the fraction of evaluated outcomes at each medal tier; labels report the final Medal Rate.}
  \label{fig:solution-ceiling-medals}
\end{figure}

Figure~\ref{fig:solution-ceiling-medals} reveals a consistent qualitative shift for the primary 35B model: post-training and \openmleevomax{} do not merely increase medal coverage, but move successful solutions toward Gold.
The companion 30B comparison reproduces the same direction of change, while the fixed-model GLM-5.2 and MiniMax M3 comparisons show that the pattern also extends to search improvements. Together, these results indicate improved solution quality rather than only more Bronze-threshold crossings. Compared with external systems, \thirtyfivebmodel{} with \openmleevomax{} outperforms Claude Opus 4.8 with Claude Code and Gemini 3.5 Flash with Gemini CLI, and matches Kimi K3's Gold rate.

\subsection{Search Efficiency and Mechanism}
\label{sec:search-efficiency-mechanism}

\begin{figure*}[htbp]
  \centering
  \includegraphics[width=0.99\textwidth]{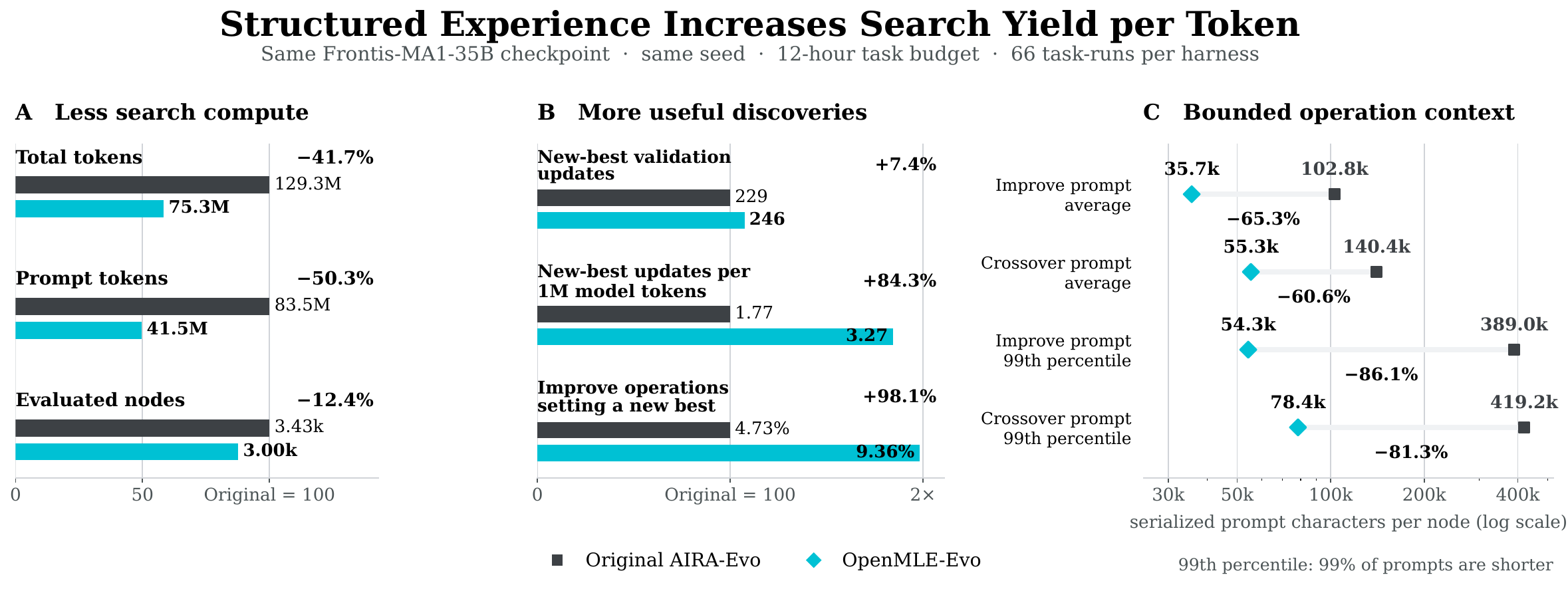}
  \caption{Search efficiency and context-length comparison between original AIRA-Evo (gray) and \openmleevo{} (cyan) over 66 matched task--runs. Panel A normalizes each resource metric to original AIRA-Evo${}=100$ and annotates the absolute totals. Panel B reports validation-trajectory productivity. A ``new-best validation update'' is a strict increase in selection reward after a task--run's first finite score; the per-million-token denominator includes both prompt and completion tokens, and the \textsc{Improve} rate is the fraction of \textsc{Improve} operations that set a new best. Panel C reports serialized user-prompt length in Unicode characters on a logarithmic axis; the 99th percentile is the length below which 99\% of prompts fall.}
  \label{fig:search-efficiency-mechanism}
\end{figure*}

\begin{takeaway}
The available trajectory evidence indicates that bounded, operation-conditioned context can improve search productivity while targeted recombination and multi-factor parent selection preserve complementary hypotheses that score-only lineage search may discard.
\end{takeaway}
\paragraph{OpenMLE-Evo versus original AIRA-Evo.}
Figure~\ref{fig:search-efficiency-mechanism} compares complete single-worker trajectories from the same \thirtyfivebmodel{} checkpoint under the same seed and 12-hour task budget, covering 66 task--run evaluations per harness.
\textbf{Lower search cost.}
Panel A shows that \openmleevo{} reduces total model-token consumption from $129.3$M to $75.3$M ($-41.7\%$) and prompt tokens from $83.5$M to $41.5$M ($-50.3\%$), while the number of evaluated nodes falls only from $3430$ to $3004$ ($-12.4\%$).
The much larger reduction in tokens than in nodes indicates that the saving comes primarily from making each expansion cheaper, not merely from terminating the search earlier or evaluating far fewer candidates.
\textbf{Higher search yield.}
Panel B counts a new-best validation update whenever a node strictly improves the task--run's best selection reward after its first finite result.
Although it evaluates fewer nodes, \openmleevo{} records 246 such updates rather than 229, raising new-best updates per million total model tokens from $1.77$ to $3.27$ ($+84.3\%$).
The fraction of \textsc{Improve} operations that establish a new best likewise rises from $44/931$ ($4.73\%$) to $72/769$ ($9.36\%$), showing that refinement calls are more likely to produce useful progress.
\textbf{Bounded operation context.}
Panel C connects this improved yield to the intended context mechanism.
For \textsc{Improve}, the mean serialized user-prompt length falls from $102.8$K to $35.7$K Unicode characters ($-65.3\%$), and its 99th percentile falls from $389.0$K to $54.3$K ($-86.1\%$).
For \textsc{Crossover}, the corresponding mean falls from $140.4$K to $55.3$K ($-60.6\%$), while the 99th percentile falls from $419.2$K to $78.4$K ($-81.3\%$).
The compression is especially strong in the tail, consistent with structured operation-conditioned memory preventing long histories from being repeatedly serialized into every request.
Together, the panels show that \openmleevo{} improves the productivity of each refinement.
We characterize search efficiency using token usage, context length, and validation-trajectory productivity.

\begin{figure*}[htbp]
  \centering
  \includegraphics[width=0.98\textwidth]{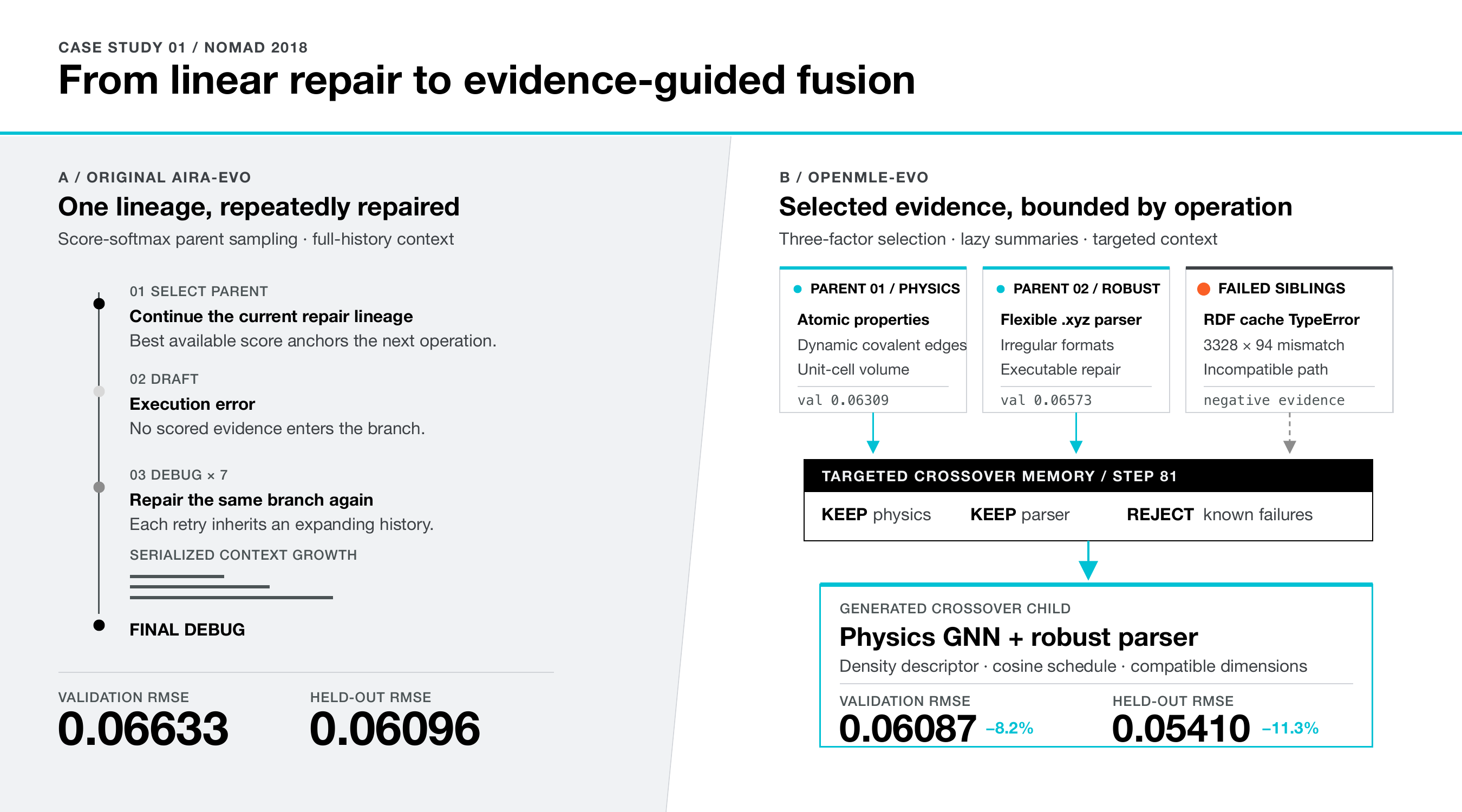}
  \caption{Single-branch repair versus targeted Crossover on \textit{nomad2018-predict-transparent-conductors}. Original AIRA-Evo repeatedly debugs one lineage using full-history context. \openmleevo{} selects complementary physics and parsing parents, excludes failures observed in nearby siblings, and forms an operation-bounded Crossover child. Lower RMSE is better.}
  \label{fig:case-study-nomad-targeted-crossover}
\end{figure*}

\textbf{Targeted Crossover escapes a single-branch Debug loop.}
Figure~\ref{fig:case-study-nomad-targeted-crossover} contrasts two grounded \textit{nomad2018-predict-transparent-conductors} traces. Original AIRA-Evo follows a single lineage after its Draft fails: seven successive \textsc{Debug} attempts inherit an expanding full history, and the final repair reaches validation RMSE $0.06633$ and held-out RMSE $0.06096$. The \openmleevo{} trace instead constructs a targeted \textsc{Crossover} at step 81 from complementary evidence. One parent contributes atomic properties, dynamic covalent edges, and unit-cell volume (validation RMSE $0.06309$); the other contributes a robust parser for irregular \texttt{.xyz} geometries (validation RMSE $0.06573$). Horizontal Memory also marks an RDF-cache \texttt{TypeError} and a $3328 \times 94$ feature mismatch as negative evidence, thereby avoiding their silent serialization into the child context. The resulting program combines the physics-informed GNN with the robust parser, density descriptors, and cosine scheduling, reaching validation RMSE $0.06087$ and held-out RMSE $0.05410$, respectively $8.2\%$ and $11.3\%$ below the original trace. The comparison illustrates the intended mechanism: operation-conditioned Memory converts distinct branch strengths and known failures into a bounded recombination request instead of repeatedly repairing one lineage.

\begin{figure*}[htbp]
  \centering
  \includegraphics[width=0.98\textwidth]{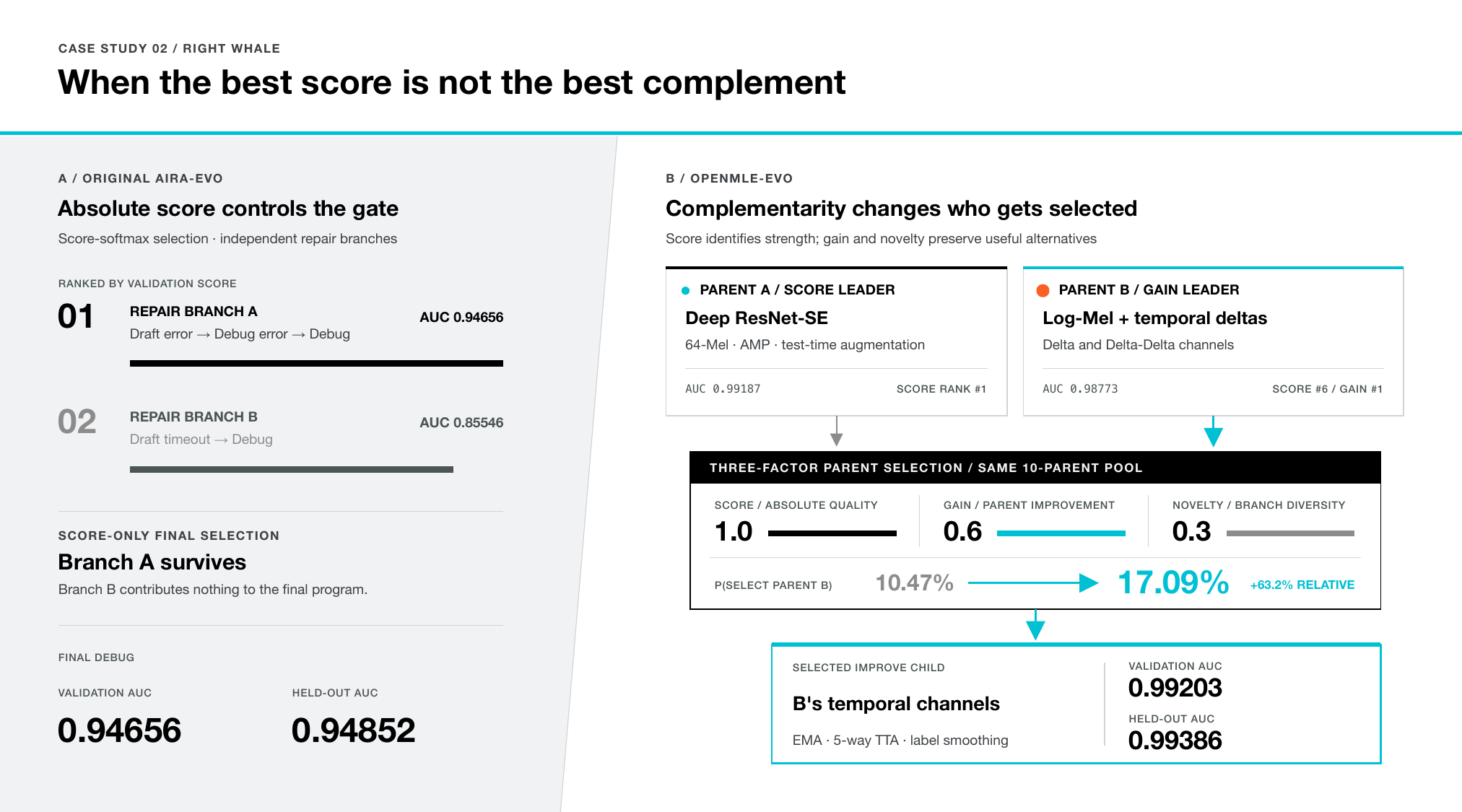}
  \caption{Score-only versus three-factor parent selection on right-whale detection. Parent A leads in current validation AUC, whereas Parent B ranks sixth by score but first by gain and retains promising temporal channels. Score/Gain/Novelty weighting raises Parent B's selection probability from $10.47\%$ to $17.09\%$ over the same ten-parent pool, enabling an Improve child derived from Parent B. Higher AUC is better.}
  \label{fig:case-study-right-whale-three-factor}
\end{figure*}

\textbf{Three-factor selection preserves a complementary parent.}
Figure~\ref{fig:case-study-right-whale-three-factor} shows why selecting parents by current score alone can discard useful structure on \textit{the-icml-2013-whale-challenge-right-whale-redux}. In the original AIRA-Evo trace, two independently repaired branches reach validation AUCs $0.94656$ and $0.85546$; score-only selection keeps the stronger branch, producing held-out AUC $0.94852$. The \openmleevo{} step-10 candidate pool exposes a subtler choice. Parent A is the score leader at AUC $0.99187$ and provides a deeper ResNet-SE pipeline with 64-Mel features, AMP, and test-time augmentation. Parent B scores slightly lower at $0.98773$ and ranks only sixth by score, but ranks first by gain after improving $0.00568$ over its own parent; it retains a promising Log-Mel representation with Delta and Delta-Delta temporal channels. With Score/Gain/Novelty weights $1.0/0.6/0.3$, Parent B moves to the top utility rank, and its selection probability within the same ten-parent pool increases from $10.47\%$ under score-only softmax to $17.09\%$, a $63.2\%$ relative increase. Parent B is then selected for an \textsc{Improve} operation, whose child reaches validation AUC $0.99203$ and held-out AUC $0.99386$. This within-pool probability recomputation directly exposes the selector's effect: the additional factors do not force a lower-scoring branch to win, but keep a high-gain, structurally distinct branch actionable long enough to be selected and refined. Because the full framework also changes targeted Memory, the end-to-end difference from original AIRA-Evo should not be attributed to the three weights alone.

\subsection{Meta-Ability and Transfer}
\label{sec:meta-ability-transfer}

\begin{figure}[htbp]
  \centering
  \includegraphics[width=\textwidth]{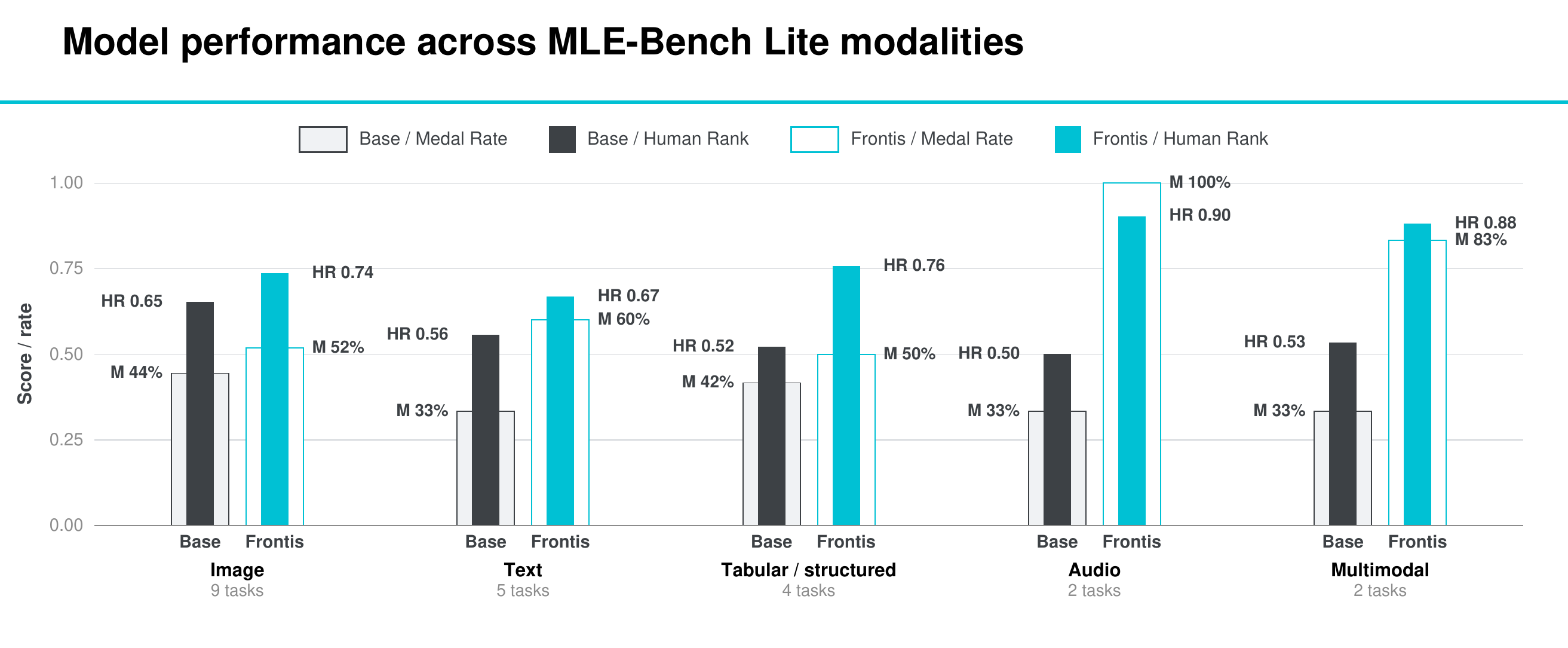}
  \caption{Modality-stratified results on the MLE-Bench Lite subset. Base denotes Qwen3.6-35B-A3B and Frontis denotes \thirtyfivebmodel{}, both evaluated with \openmleevo{}. Wide outlined bars show Medal Rate, and narrow filled bars show mean Human Rank; both use all three task--epoch outcomes per task, with invalid final submissions assigned zero Human Rank. }
  \label{fig:mle-modality-stratified}
\end{figure}

\begin{takeaway}
Modality-stratified MLE-Bench results show broad gains across data types, and controlled NatureBench Lite results provide initial evidence of transfer beyond competition-style MLE; small modality groups and the ten-task transfer set limit the breadth of both conclusions.
\end{takeaway}
\paragraph{Cross-modality meta-ability on MLE-Bench.}
Before testing out-of-benchmark transfer, we ask whether the learned improvement capability is tied to a narrow input modality.
We partition the 22 MLE-Bench Lite tasks into five mutually exclusive groups: image, text, tabular/structured, audio, and multimodal.
Relative to Qwen3.6-35B-A3B under the same \openmleevo{} harness, \thirtyfivebmodel{} raises mean Human Rank in all five groups and never lowers group-level Medal Rate (Figure~\ref{fig:mle-modality-stratified}).
The 14 additional medals are distributed across every group (image/text/tabular/audio/multimodal: $+2/+4/+1/+4/+3$), so the aggregate gain is not explained by one modality alone.

\paragraph{Generalization to NatureBench.}
\label{sec:naturebench}

NatureBench evaluates whether coding agents can recover or improve upon published scientific results~\citep{wang2026naturebench}.
Its full benchmark contains 90 containerized tasks distilled from peer-reviewed Nature-family papers across six scientific domains.
Each task hides the test ground truth and paper method behind a host-side evaluator, and compares heterogeneous scientific metrics through the direction-normalized relative gap
\begin{equation}
g = \mathrm{dir}\,\frac{m-m_{\mathrm{SOTA}}}{|m_{\mathrm{SOTA}}|},
\label{eq:naturebench-gap}
\end{equation}
where $m$ is the submitted result and $\mathrm{dir}\in\{-1,+1\}$ accounts for whether the task metric is minimized or maximized.
We report \emph{Match-SOTA} (All M), the fraction of tasks with $g\geq 0$, and \emph{Surpass-SOTA} (All S), the stricter fraction with $g>0.1$.

For a tractable but heterogeneous transfer study, we use NatureBench Lite, a fixed 10-task subset spanning all six domains, six represented input-modality families, and four ML task types; Appendix~\ref{app:naturebench-lite} lists the tasks.
We retain the NatureBench task containers, hidden evaluator, validity rules, web-search-disabled setting, and four-hour search budget per task.
The \openmleevo{} NatureBench adapter changes the task interface, resource scheduling, and feedback plumbing needed to run the same evolutionary program operators against NatureBench's evaluator; it does not expose hidden labels or the paper solution.
Table~\ref{tab:naturebench-lite-results} reports the reference agent results and our controlled model--harness comparisons.

The transfer results expose contributions from both the post-trained model and the adapted search framework.
Holding the NatureBench adapter fixed, \thirtyfivebmodel{} improves over its Qwen3.6-35B-A3B base by 10 percentage points in All S ($3/10$ versus $2/10$) and 20 points in All M ($7/10$ versus $5/10$).
Holding the base model fixed, the \openmleevo{} NatureBench adapter improves over original AIRA-Evo by 10 points in All S ($2/10$ versus $1/10$) and 30 points in All M ($5/10$ versus $2/10$).
Consequently, the combined \thirtyfivebmodel{} system matches the $3/10$ All S and $7/10$ All M attained by GPT-5.4, GLM-5.1, and MiniMax-M3 on this subset, and exceeds the reported DeepSeek-V4-Pro, Claude Opus 4.6, and MiniMax-M2.7 configurations.
This provides evidence that execution-grounded post-training transfers beyond competition-style MLE and that the adapted long-horizon search can convert more of that capability into paper-relative scientific progress.

\begin{table}[H]
\centering
\caption{Results on NatureBench (NB) Lite. S and M denote Surpass-SOTA ($g>0.1$) and Match-SOTA ($g\geq 0$), respectively. Public reference agents use Codex for GPT models, Gemini CLI for Gemini 3.5 Flash, and Claude Code for all other models. Bold indicates the best result.}
\label{tab:naturebench-lite-results}
\vspace{0.4em}
\scriptsize
\setlength{\tabcolsep}{3.5pt}
\renewcommand{\arraystretch}{1.06}
\resizebox{\linewidth}{!}{%
\begin{tabular}{@{}cllcc@{}}
\toprule
Rank & Model & Agent harness & All S $\uparrow$ & All M $\uparrow$ \\
\midrule
\multicolumn{5}{@{}l}{\textit{NatureBench Lite reference agents}} \\
1  & Claude Opus 4.7    & Claude Code & \textbf{70.0\% (7/10)} & \textbf{100.0\% (10/10)} \\
2  & GLM-5.2             & Claude Code & \textbf{70.0\% (7/10)} & \textbf{100.0\% (10/10)} \\
3  & Gemini 3.5 Flash    & Gemini CLI  & 60.0\% (6/10) & 80.0\% (8/10) \\
4  & GPT-5.5             & Codex       & 40.0\% (4/10) & \textbf{100.0\% (10/10)} \\
5  & Qwen 3.7 Max        & Claude Code & 40.0\% (4/10) & 60.0\% (6/10) \\
6  & Kimi K2.6           & Claude Code & 30.0\% (3/10) & 90.0\% (9/10) \\
7  & GPT-5.4             & Codex       & 30.0\% (3/10) & 70.0\% (7/10) \\
8  & GLM-5.1             & Claude Code & 30.0\% (3/10) & 70.0\% (7/10) \\
9  & MiniMax-M3          & Claude Code & 30.0\% (3/10) & 70.0\% (7/10) \\
10 & DeepSeek-V4-Pro     & Claude Code & 20.0\% (2/10) & 60.0\% (6/10) \\
11 & Claude Opus 4.6     & Claude Code & 20.0\% (2/10) & 50.0\% (5/10) \\
12 & MiniMax-M2.7        & Claude Code & 0.0\% (0/10)  & 30.0\% (3/10) \\
\midrule
\multicolumn{5}{@{}l}{\textit{OpenMLE controlled comparisons}} \\
-- & \thirtyfivebmodel{} & \openmleevo{} NB adapter & \textbf{30.0\% (3/10)} & \textbf{70.0\% (7/10)} \\
-- & Qwen3.6-35B-A3B & \openmleevo{} NB adapter & 20.0\% (2/10) & 50.0\% (5/10) \\
-- & Qwen3.6-35B-A3B & Original AIRA-Evo & 10.0\% (1/10) & 20.0\% (2/10) \\
\bottomrule
\end{tabular}%
}
\end{table}

\textbf{NatureBench trajectory: protein variant effect prediction.}
Under the same NatureBench adapter, \thirtyfivebmodel{} reaches task-level aggregate improvement $g=0.1161$ across 11 protein-assay instances, versus $g=0.0243$ for its Qwen3.6-35B-A3B base. The search advances from a valid \textsc{Draft} at $0.0679$ through \textsc{Debug} and \textsc{Improve} nodes to a \textsc{Crossover} incumbent at $0.1016$. Rather than greedily refining only that incumbent, the three-factor selector revisits a distinct $0.0955$ branch whose score, recent gain, and novelty remain promising. Vertical and horizontal Memory preserve successful physicochemical features while exposing nearby timeout, \texttt{KeyError}, and nested-mapping failures; the resulting \textsc{Improve} node retains the robust flat mapping and adds training-label-derived positional priors with five-fold LightGBM ensembling, reaching $0.1161$ without hidden test labels or the paper solution. The trace illustrates how post-training supplies an effective scientific refinement while structured memory keeps a non-incumbent hypothesis actionable long enough to overtake the current best.

%% file: figure_source/main_exp_results_table.tex
\begingroup
\definecolor{mainTableGray}{HTML}{F0F2F4}
\definecolor{mainTableCyan}{HTML}{00C1D4}
\definecolor{mainTableMidGray}{HTML}{8C8C8C}
\newcolumntype{M}[1]{>{\raggedright\arraybackslash}p{#1}}
\newcolumntype{N}[1]{>{\centering\arraybackslash}p{#1}}
\newcommand{\mainTablePanel}[1]{%
  \rowcolor{mainTableGray}
  \multicolumn{5}{@{}l@{}}{\rule{0pt}{2.35ex}\textbf{#1}}\\*[-0.15ex]
}
\newcommand{\mainTableGroup}[1]{%
  \multicolumn{5}{@{}l@{}}{%
    \rule{0pt}{2.35ex}%
    \textcolor{mainTableCyan}{\rule{1.6pt}{1.25ex}}\hspace{5pt}%
    \textbf{#1}}\\*[-0.15ex]
}
\newcommand{\mainTableFrontis}[1]{%
  \textcolor{mainTableCyan}{\rule{1.35pt}{1.15ex}}\hspace{4pt}\textbf{#1}}
\newcommand{\mainTableEvoMax}{\textbf{\openmleevomax{}}}
\newcommand{\mainTableBest}[1]{\textbf{#1}}

\small
\setlength{\tabcolsep}{3.2pt}
\renewcommand{\arraystretch}{1.18}
\setlength{\LTpre}{0pt}
\setlength{\LTpost}{0pt}
\setlength{\LTcapwidth}{\linewidth}
\begin{longtable}{@{}M{0.30\linewidth}M{0.22\linewidth}N{0.12\linewidth}N{0.145\linewidth}N{0.125\linewidth}@{}}
\caption{MLE-Bench Lite results. Panel A presents the primary \thirtyfivebmodel{} comparison first, followed by the companion \thirtybmodel{} replication and matched search comparisons; Panels B--C provide broader system context.}
\label{tab:arxiv-main-results}\\[-0.15em]
\toprule
\textbf{Model / system} &
\textbf{Framework} &
\textbf{Valid Rate $\uparrow$} &
\textbf{Medal Average $\uparrow$} &
\textbf{Human Rank $\uparrow$} \\
\midrule
\endfirsthead

\multicolumn{5}{@{}l@{}}{%
  \small\textit{\tablename\ \thetable\ --- continued from previous page}}\\[0.35em]
\toprule
\textbf{Model / system} &
\textbf{Framework} &
\textbf{Valid Rate $\uparrow$} &
\textbf{Medal Average $\uparrow$} &
\textbf{Human Rank $\uparrow$} \\
\midrule
\endhead

\midrule
\multicolumn{5}{r@{}}{%
  \scriptsize\textcolor{mainTableMidGray}{\textit{Continued on next page}}}\\
\endfoot

\bottomrule
\endlastfoot

\mainTablePanel{A. Controlled comparisons}
\mainTableGroup{\thirtyfivebmodel{}}
Qwen3.6-35B-A3B & \openmleevo{} & 19.67/22 & 39.39\% & 0.5828 \\
\rowcolor{mainTableGray}
\mainTableFrontis{\thirtyfivebmodel{}} & \openmleevo{} & 21.67/22 & \frontisThirtyFiveEvoResult{} & \frontisThirtyFiveEvoHumanRank{} \\
\rowcolor{mainTableGray}
\mainTableFrontis{\thirtyfivebmodel{}} & \mainTableEvoMax & 22.00/22 & \mainTableBest{\frontisThirtyFiveEvoMaxResult{}} & \mainTableBest{\frontisThirtyFiveEvoMaxHumanRank{}} \\
\addlinespace[0.45em]

\mainTableGroup{\thirtybmodel{}}
Qwen3-30B-A3B\footnote{The detailed version is Qwen3-30B-A3B-Thinking-2507.} & \openmleevo{} & 17.33/22 & 34.85\% & 0.5573 \\
\textbf{\thirtybmodel{}} & \openmleevo{} & 21.67/22 & \frontisThirtyEvoResult{} & \frontisThirtyEvoHumanRank{} \\
\textbf{\thirtybmodel{}} & \mainTableEvoMax & 22.00/22 & \frontisThirtyEvoMaxResult{} & \frontisThirtyEvoMaxHumanRank{} \\
\addlinespace[0.45em]

\mainTableGroup{Matched harness comparison · GLM-5.2}
GLM-5.2 & Claude Code & 21.00/22 & 59.09\% & 0.7948 \\*
GLM-5.2 & \openmleevo{} & 19.67/22 & 62.12\% & 0.7069 \\*
GLM-5.2 & \mainTableEvoMax & 22.00/22 & \mainTableBest{66.67\%} & \mainTableBest{0.8164} \\
\addlinespace[0.45em]

\mainTableGroup{Matched harness comparison · MiniMax M3}
MiniMax M3 & Codex & 22.00/22 & 54.55\% & 0.7099 \\
MiniMax M3 & \openmleevo{} & 22.00/22 & 59.09\% & 0.7994 \\
MiniMax M3 & \mainTableEvoMax & 22.00/22 & \mainTableBest{65.15\%} & \mainTableBest{0.8007} \\
\addlinespace[0.45em]

\mainTableGroup{Matched harness comparison · Kimi K2.6}
Kimi K2.6 & Claude Code & 18.00/22 & 59.09\% & 0.7062 \\
Kimi K2.6 & \openmleevo{} & 21.67/22 & \mainTableBest{66.67\%} & \mainTableBest{0.7859} \\
\addlinespace[0.45em]

\mainTableGroup{Matched harness comparison · MiniMax M2.7}
MiniMax M2.7 & Claude Code & 18.00/22 & 45.50\% & 0.5547 \\
MiniMax M2.7 & \openmleevo{} & 22.00/22 & \mainTableBest{50.00\%} & \mainTableBest{0.7039} \\
\addlinespace[0.55em]

\mainTablePanel{B. Broader OpenMLE-Evo system context}
Grok-4.5 & \openmleevo{} & 22.00/22 & 65.15\% & 0.8052 \\
LongCat-2.0 & \openmleevo{} & 21.00/22 & 56.06\% & 0.7343 \\
Doubao Seed 2.1 Pro & \openmleevo{} & 20.33/22 & 56.06\% & 0.7170 \\
Qwen3.7 Plus & \openmleevo{} & 21.67/22 & 54.55\% & 0.7234 \\
DeepSeek-V4-Pro & \openmleevo{} & 21.67/22 & 54.55\% & 0.6849 \\
DeepSeek-V4-Flash & \openmleevo{} & 21.33/22 & 51.52\% & 0.6957 \\
GLM-4.7 & \openmleevo{} & 21.33/22 & 51.52\% & 0.6521 \\
MiMo-V2.5-Pro & \openmleevo{} & 17.00/22 & 40.91\% & 0.5213 \\
Step-3.7 Flash & \openmleevo{} & 19.00/22 & 27.27\% & 0.4953 \\
\addlinespace[0.55em]

\mainTablePanel{C. General-purpose coding-agent references}
GPT-5.6 Sol & Codex & 22.00/22 & 72.73\% & 0.8891 \\
Kimi K3 & Claude Code & 22.00/22 & 72.73\% & 0.8574 \\
GPT-5.5 & Codex & 21.00/22 & 68.18\% & 0.7833 \\
Claude Opus 4.8 & Claude Code & 22.00/22 & 63.64\% & 0.8219 \\
Gemini 3.5 Flash & Gemini CLI & 20.00/22 & 63.64\% & 0.7499 \\
Claude Sonnet 5 & Claude Code & 22.00/22 & 59.09\% & 0.7730 \\
Claude Sonnet 4.6 & Claude Code & 22.00/22 & 54.55\% & 0.7670 \\

\end{longtable}
\endgroup

%% file: sections/03_related_work.tex
\section{Related Work}

\textbf{AutoResearch systems and evaluation targets.}
AutoResearch systems increasingly automate hypothesis formation, experimentation, and artifact production, from end-to-end scientific workflows and agentic tree search to focused, compute-bounded model-training loops~\citep{lu2024aiscientist,yamada2025aiscientistv2,karpathy2026autoresearch}.
Recent systems further explore persistent project state, hypothesis-tree refinement, cross-task skill accumulation, and continuously evolving multi-agent workflows~\citep{chen2026toward,jin2026hypothesistree,kim2026haste,zhu2026evomaster}.
Benchmarks isolate complementary slices of this research loop: RE-Bench and PostTrainBench target open-ended AI R\&D and autonomous model post-training~\citep{wijk2025rebench,rank2026posttrainbench}; MLGym, AIRS-Bench, MLRC-Bench, and ResearchGym study research-oriented problem solving and end-to-end ML projects~\citep{nathani2025mlgym,lupidi2026airs,zhang2025mlrcbench,garikaparthi2026researchgym}; and PaperBench and NatureBench emphasize reproducing papers or published scientific results~\citep{starace2025paperbench,wang2026naturebench}.
Within executable MLE, MLE-Bench evaluates agents on end-to-end Kaggle-style competitions~\citep{chan2024mlebench}, whereas MLS-Bench targets improving ML components in ways that generalize across controlled settings and scales~\citep{lyu2026mlsbench}.
We evaluate OpenMLE on MLE-Bench Lite as its primary MLE benchmark and on NatureBench Lite as a focused test of transfer to broader scientific AutoResearch.

\textbf{Executable MLE environments and scalable task resources.}
Classical AutoML optimizes over predefined model and pipeline spaces~\citep{thornton2013autoweka,feurer2015autosklearn,olson2016tpot,erickson2020autogluon}, whereas LLM-based MLE agents operate through open-ended, code-mediated experimentation.
Building on this benchmark lineage, MLAgentBench and DSBench likewise require agents to write, execute, debug, and submit solutions on realistic ML and data-science tasks~\citep{huang2023mlagentbench,jing2024dsbench}.
Complementary efforts expand the underlying execution and training infrastructure: MLE-Dojo standardizes interactive MLE environments~\citep{qiang2025mledojo}, while MLE-Smith and SandMLE scale the construction of executable training tasks~\citep{qiang2025mlesmith,zhou2026sandmle}.
These resources make execution feedback available at increasing scale, but executable tasks alone do not specify how experience should be transformed into reusable model capabilities or composed at inference time.
\openmlegym{} builds on this line by unifying scalable task environments with isolated execution, structured feedback, and task-specific evaluators as a shared substrate for post-training, search, and evaluation.

\textbf{Inference-time scaffolds and evolutionary search.}
Given executable evaluators, inference-time scaffolds amplify frozen models by allocating trials, maintaining search state, and selecting or transforming candidate programs.
MLE systems instantiate this idea through multi-agent decomposition~\citep{trirat2024automlagent,gandhi2024budgetmlagent,li2025autokaggle,fang2025mlzero}, structured search~\citep{jiang2025aide,chi2024sela}, and targeted refinement or domain knowledge~\citep{liu2025ml,ou2025automind,nam2025mlestar,du2025automlgen,zhang2025deepanalyze}.
AIRA-style analyses further identify search policy, operator quality, throughput, and ideation diversity as key performance factors~\citep{toledo2025airesearchagents,audranreiss2025ideationdiversity,hambardzumyan2026aira_2}.
More generally, program-evolution systems use language models as mutation operators and executable evaluators as selection signals~\citep{novikov2025alphaevolve,lange2025shinkaevolve,cemri2026adaevolve,ye2026evaluation}, while ThetaEvolve and TTT-Discover update problem-solving behavior from test-time feedback~\citep{wang2025thetaevolve,yuksekgonul2026learning}.
These methods establish the value of structured search, but typically rely on transformation behavior supplied by the frozen backbone or its prompts; \openmleevo{} instead composes operators explicitly trained for the same roles used during search.

\textbf{Learning MLE agents from executable experience.}
A complementary line internalizes verifiable outcomes through supervised or reinforcement-learning updates rather than retaining all improvement logic in an external scaffold.
RLVR has produced strong learned reasoning capabilities in mathematics and coding~\citep{guo2025deepseek,team2025kimi,zeng2025rlve} and has increasingly been extended to long-horizon agent tasks~\citep{team2026kimi,zeng2026glm,minimax2026m27}.
MLE-specific efforts similarly learn from executable task rewards~\citep{liu2025mlagent,li2025mlerl,yang2025rlmleagents,cai2026acegrpo}, establishing that MLE experience can be internalized by the model.
However, these efforts expose different subsets of the task, environment, training, evaluation, and model artifacts needed for a reproducible end-to-end workflow, and their learned policies are not uniformly organized around an interface shared with long-horizon search (Appendix Table~\ref{tab:mle-open-release-matrix}).
\openmleerl{} follows this learning direction but trains reusable \textsc{Draft}, \textsc{Improve}, \textsc{Debug}, and \textsc{Crossover} transformations whose interfaces match the operations invoked by \openmleevo{}.

\textbf{AI-for-AI and trainable improvers.}
AI-for-AI broadens executable improvement from optimizing task solutions to improving the models, operators, and harnesses that generate future solutions~\citep{jiang2026selfimprovingagents}.
Search--learning systems begin to return execution experience to the generator, as in test-time policy updates and schemes that alternate evolutionary search with hindsight fine-tuning~\citep{wang2025thetaevolve,yuksekgonul2026learning,pourcel2025soar}.
Related work extends improvement from candidate programs to agent and harness design~\citep{hu2024adas,zhang2026hyperagents,lee2026metaharness,zhang2026aevo}.
OpenMLE contributes at the interface of these directions: verified evolutionary experience post-trains the same program-transformation operators that subsequently govern evolutionary search, creating a meta-evolutionary coupling between learning and inference.
Relative to a public landscape that often exposes individual or partial components of the workflow, OpenMLE is organized as an open stack spanning task packages, sandbox execution, operator training, search, evaluation, and released model weights.
This coupling provides a concrete testbed for studying progress toward RSI in executable MLE, rather than a claim that OpenMLE realizes general, autonomous recursive self-improvement.

%% file: sections/08_limitations_and_discussion.tex
\section{Limitations and Future Work}
\label{sec:limitations-and-future-work}

OpenMLE provides an open path from executable environments to post-trained models and long-horizon search, but it does not yet realize the full vision of recursive self-improvement (RSI). We highlight five capability boundaries that we consider most consequential for closing this gap.

\textbf{Richer objectives for improving the improver.}
OpenMLE currently learns primarily from the measured outcome of an executed
solution. This signal reveals whether a program works and how well it scores,
but it does not fully capture whether a research direction is promising,
generalizable, robust, or worth additional computation. As a result, the
system is better equipped to optimize solutions than to judge which ideas
deserve to be pursued. A more capable improver will require objectives that
represent not only final performance, but also the quality of hypotheses,
reasoning processes, critiques, and transferable research strategies.

\textbf{Integrating evolutionary search with general coding agents.}
Our present system composes trained operators through an external evolutionary
harness. This separation makes training and search tractable, but it also
bounds the range of actions and interactions the model can initiate on its
own. Moving beyond this boundary requires combining evolutionary search with
general coding agents, bringing population-based exploration and flexible
agentic problem solving into a unified framework.

\textbf{Broader participation in AI development.}
The current environments ask an agent to improve external machine learning
artifacts. They provide a practical and verifiable testbed for meta-evolution,
but the agent participates in only a limited part of the broader AI development
process. Moving closer to recursive self-improvement requires agents to take
part in a larger share of this process, especially the improvement of language
models themselves.

\textbf{Evolving the evolutionary system.}
In OpenMLE, evolution operates primarily over candidate solutions, while the
evolutionary system itself remains largely fixed. A further step toward
recursive self-improvement is therefore to make the evolutionary system itself
an object of evolution.

\textbf{Richer use of experience in node expansion.}
Our experience-guided node-expansion framework remains a preliminary
prototype. Although each experience card preserves a broad set of
deterministic metadata about a node, the current parent-selection policy uses only three
factors that we consider especially important: solution quality,
parent-relative improvement, and method-family novelty. This design
demonstrates that structured experience can guide the allocation of search
budget, but it leaves much of the recorded evidence underutilized. Future
work could incorporate additional signals, while learning task-dependent rather than fixed
factor weights. More broadly, enabling the search policy itself to discover
which experience signals are predictive would provide a promising path from
hand-designed experience guidance toward an evolutionary system that
improves its own search behavior.

%% file: sections/09_conclusion.tex
\section{Conclusion}
We presented OpenMLE, an open full-stack technical solution for training and deploying language-model agents that construct and iteratively improve machine learning solutions through executable feedback. \openmlegym{} provides quality-gated tasks, isolated execution, and task-specific evaluation; \openmleerl{} learns \textsc{Draft}, \textsc{Improve}, \textsc{Debug}, and \textsc{Crossover} transformations through execution-grounded supervised fine-tuning and reinforcement learning; and \openmleevo{} composes the same operators into long-horizon search using structured experience, multi-factor parent selection, and operator-conditioned memory. This shared operator and execution interface makes \thirtyfivebmodel{} both the product of the training stack and the variation engine of its evolutionary harness.

The results show that model learning and search provide complementary gains. Under the identical \openmleevo{} harness, \thirtyfivebmodel{} raises Medal Average over its Qwen3.6-35B-A3B base from $39.39\%$ to \frontisThirtyFiveEvoResult{}, while the companion \thirtybmodel{} reproduces the gain on a second backbone and scale. Matched comparisons further show that \openmleevo{} outperforms general-purpose coding-agent scaffolds across four frontier models and original AIRA-Evo on \thirtyfivebmodel{}; combining \thirtyfivebmodel{} with \openmleevomax{} reaches \frontisThirtyFiveEvoMaxResult{}. Mechanism analyses show sustained late-horizon gains from refinement and recombination, while the matched AIRA-Evo comparison records shorter bounded contexts, lower token use, and more validation progress per token. Controlled comparisons on the ten-task NatureBench Lite subset provide initial evidence that both the post-trained model and the adapted search framework transfer beyond competition-style MLE. Together with the released stack, these results establish OpenMLE as a reproducible testbed for meta-evolution in executable AI4AI.

\section{Authors}

\begingroup
\newcommand{\affmark}[1]{\textsuperscript{#1}}
\setlength{\tabcolsep}{10pt}
\renewcommand{\arraystretch}{1.4}

\begin{tabularx}{\linewidth}{
@{}
>{\raggedright\arraybackslash}X
>{\raggedright\arraybackslash}X
>{\raggedright\arraybackslash}X
@{}
}
Junlin Yang\affmark{1,2,*,$\dagger$}
& Che Jiang\affmark{1,2,*,$\dagger$}
& Yu Fu\affmark{1,3,*} \\

Tianwei Luo\affmark{2,*}
& Can Ren\affmark{1,*}
& Weizhi Wang\affmark{1,2,*} \\

Kaikai Zhao\affmark{2,*}
& Hongyi Liu\affmark{2}
& Yuxin Zuo\affmark{2} \\

Yuru Wang\affmark{1,2}
& Yuchen Fan\affmark{4}
& Kai Tian\affmark{1,2} \\

Zhenzhao Yuan\affmark{1,2}
& Xiaojian Lin\affmark{2}
& Li Sheng\affmark{2} \\

Rushi Qiang\affmark{5}
& Guoli Jia\affmark{2}
& Xingtai Lv\affmark{1,2} \\

Ermo Hua\affmark{2}
& Dianqiao Lei\affmark{1,2}
& Youbang Sun\affmark{2} \\

Ning Ding\affmark{2}
& Bowen Zhou\affmark{2}
& Kaiyan Zhang\affmark{1,$\ddagger$}
\end{tabularx}

\vspace{0.8em}

{\footnotesize\color{frontisgray}
\setlength{\parskip}{2pt}

\textsuperscript{1} Horizon Research, Frontis.AI \quad
\textsuperscript{2} Tsinghua University\quad
\textsuperscript{3} Zhejiang University\par
\textsuperscript{4} Shanghai Jiao Tong University\quad
\textsuperscript{5} Georgia Institute of Technology\par

\vspace{0.35em}

\textsuperscript{*} Core Contributor
\quad
\textsuperscript{$\dagger$} Project Leader\quad
\textsuperscript{$\ddagger$}
Corresponding Author
}
\endgroup

%% file: sections/10_references.tex
{
\small
\bibliography{references}
}

%% file: sections/11_technical_appendices.tex
\newpage
\section{OpenMLE-Gym Details}

\subsection{Task Construction and Selection}
\label{app:data-construction}

This appendix details the construction, validation, and selection mechanisms summarized in Section~\ref{sec:task-curation-data-construction}. All three task sources follow the shared executable contract defined there; we focus below on the automated competition branch and its post-construction gates.

\textbf{Competition-construction state machine.}
The builder consumes the candidate slugs remaining after upstream filtering. Independent competitions can be dispatched concurrently, while each competition follows an ordered state machine so that downstream stages run only after their prerequisites succeed.

For each slug, the pipeline downloads and recursively unpacks the competition archive, preserves the source assets, and retrieves the pre-computed competition overview and dataset description. A tool-assisted file-perception stage then enumerates the local tree, probes tabular files for structural summaries and sampled rows, and reads textual documentation. The resulting inventory and competition metadata jointly ground the generated task description, avoiding reliance on either web metadata or filenames alone.

\textbf{Executable construction and validation.}
Conditioned on the generated description, the pipeline produces and executes \texttt{prepare.py} to create deterministic train/test splits, copy associated modality files, expose participant-visible inputs, and isolate held-out answers. Execution success alone is insufficient: the builder checks for the task description, training data, test inputs, sample submission, and private test answers before accepting the package. A failed attempt clears its partial processed outputs and returns the execution error to the next generation attempt, preventing malformed intermediate state from being reused.

After preparation succeeds, the pipeline generates the task-specific \texttt{metric.py} from the task description together with samples of the submission and private-answer schemas. Metric generation and metric validation are separate stages. The latter dynamically loads the generated evaluator and scores \path{data/public/sample_submission.csv} against \path{data/private/test_answer.csv}; import errors, execution errors, missing inputs, or a missing score fail this gate before semantic assessment.

Raw assets may remain in \texttt{raw/}. For storage-reduced packages, the generated file inventory is first preserved as \texttt{raw.txt} before the raw directory is removed. The quality record consumes this inventory when available; otherwise, raw-data use is assessed from the presence of source assets, \texttt{prepare.py}, and the resulting public/private data.

\begin{figure*}[t]
  \centering
  \includegraphics[page=3,width=\textwidth]{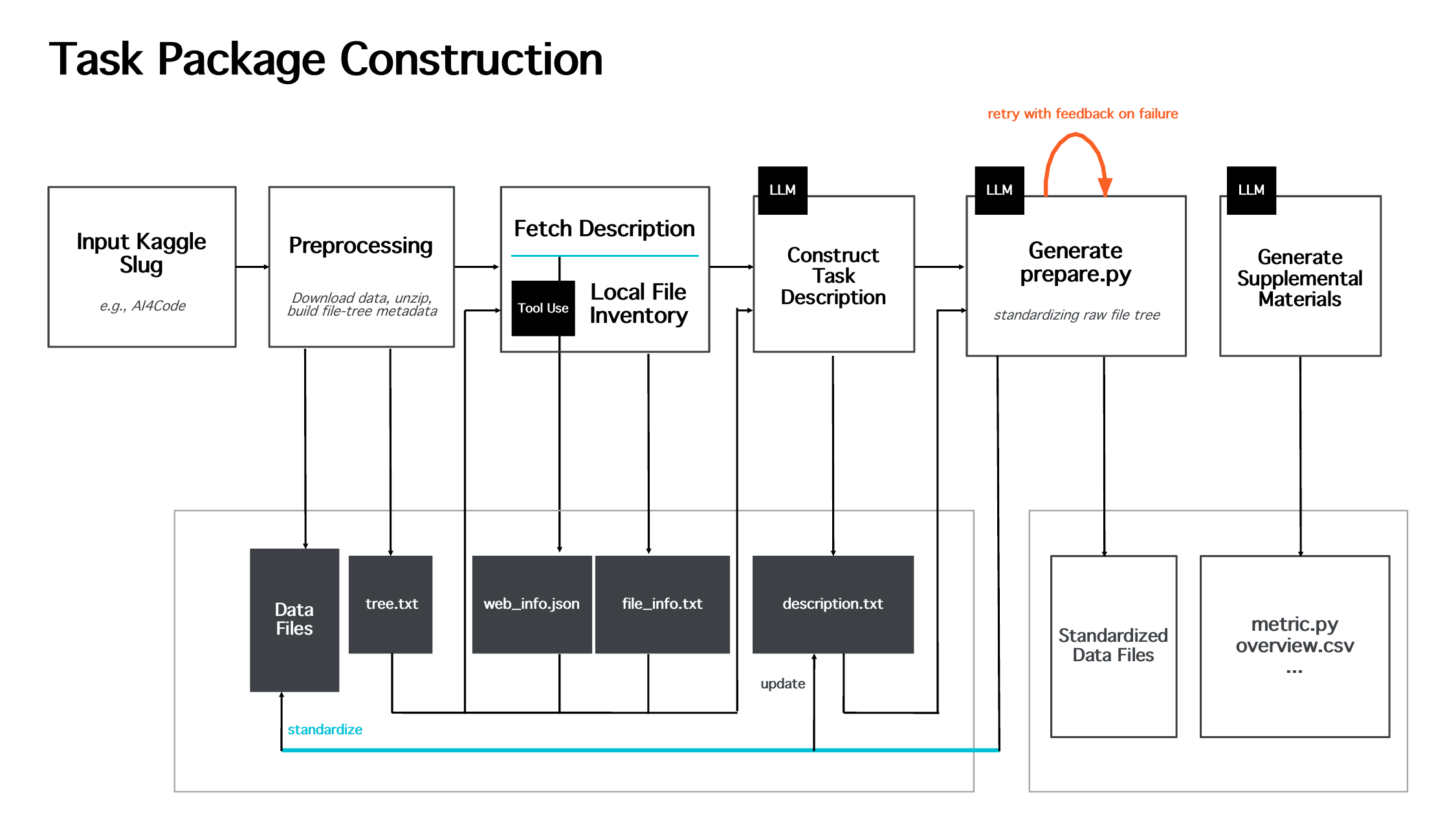}
  \caption{Automated competition task construction. Competition metadata and local-file evidence ground package generation; execution feedback revises failed preparation before metric validation and semantic gating.}
  \label{fig:competition-task-construction}
\end{figure*}

\textbf{Metadata profiling and semantic quality assessment.}
Post-construction profiling annotates modality and task type from the task description, measures raw and processed data size, classifies the expected CPU/GPU requirement, and records metric-validation output. These stages can process task chunks concurrently but restore the original task order when writing the aggregate metadata.

The quality evaluator first inspects package structure and executes the metric smoke test. Missing critical files or failed metric execution deterministically yield \texttt{not\_recommended}. For packages that pass these hard gates, the semantic judge receives the description, \texttt{prepare.py}, train/test sizes, processed-data size, auxiliary file types, sampled public and held-out rows, available raw inventory, and metric result. It returns scores for task validity, data sufficiency, raw-data usage, task complexity, and data quality, together with one of \texttt{recommended}, \texttt{conditional}, or \texttt{not\_recommended}. The evaluator records these judgments without mutating the package collection; final aggregation admits only metric-valid tasks with the strict \texttt{recommended} decision.

\textbf{Metadata accounting.}
The distribution plots in Figure~\ref{fig:task-scale-distribution} use mutually exclusive normalized groups. Tasks containing multiple modalities are assigned to Multimodal, compound task labels to Multitask, and rare labels to Other. Package size is measured from constructed task contents.

\subsection{OpenMLE-Gym Execution Infrastructure}
\label{app:sandbox}

\begin{figure}[h!]
  \centering
  \includegraphics[width=\linewidth]{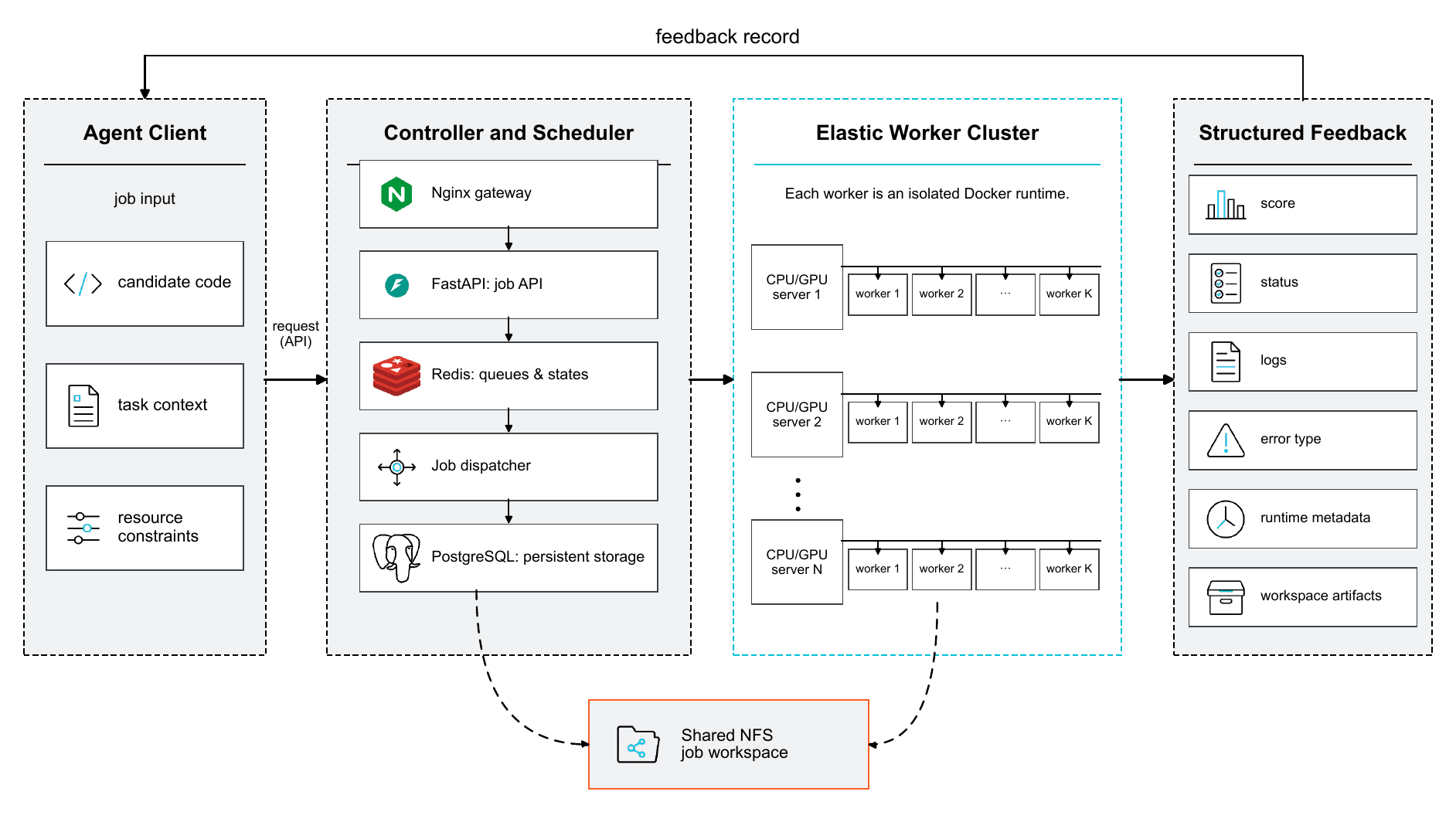}
  \caption{OpenMLE-Gym execution-backend architecture. Agent requests are dispatched to CPU/GPU Docker workers, which execute candidate programs against task data and evaluators and return structured feedback.}
  \label{fig:sandbox-architecture}
\end{figure}

\textbf{Architecture.}
Figure~\ref{fig:sandbox-architecture} expands the execution backend summarized in Section~\ref{sec:gym-sandbox-execution}, exposing the API boundary and the separation among scheduling, isolated workers, shared job state, and returned feedback.

\textbf{Representative feedback cases.}

Table~\ref{tab:sandbox-real-cases} grounds the six feedback modes summarized in Section~\ref{sec:gym-sandbox-execution} with representative jobs. Each row connects a concrete trigger to returned fields and their usable signal; the following transcript traces one failure end to end.

\begin{table}[h!]
  \centering
  \caption{Representative coverage of sandbox feedback modes from real job logs. Each row abstracts one real job into its trigger, returned feedback fields, and usable signal for training or evaluation.}
  \label{tab:sandbox-real-cases}
  \vspace{0.4em}
  \footnotesize
  \setlength{\tabcolsep}{5pt}
  \renewcommand{\arraystretch}{1.14}
  \begin{tabular}{@{}>{\raggedright\arraybackslash}p{0.19\linewidth}>{\raggedright\arraybackslash}p{0.73\linewidth}@{}}
    \toprule
    \textbf{Case} & \textbf{Trigger, feedback, and usable signal} \\
    \midrule
    \textbf{Success}
    & \textbf{Trigger:} candidate finishes training and writes \texttt{submission.csv}.\newline
      \textbf{Feedback:} \texttt{completed}; score $0.9991$; logs; runtime metadata; validated artifact.\newline
      \textbf{Signal:} valid scored trajectory. \\
    \addlinespace[2pt]
    \textbf{Runtime error}
    & \textbf{Trigger:} code calls an unsupported PyTorch API argument.\newline
      \textbf{Feedback:} \texttt{code\_execution\_error}; null score; exit code; traceback.\newline
      \textbf{Signal:} executable code defect before scoring. \\
    \addlinespace[2pt]
    \textbf{Missing code}
    & \textbf{Trigger:} submitted \texttt{main.py} is empty.\newline
      \textbf{Feedback:} \texttt{code\_missing}; null score; immediate pre-execution diagnostic.\newline
      \textbf{Signal:} invalid request filtered before worker execution. \\
    \addlinespace[2pt]
    \textbf{Missing submission}
    & \textbf{Trigger:} process exits successfully but does not write \texttt{submission.csv}.\newline
      \textbf{Feedback:} \texttt{submission\_missing}; null score; missing-artifact log.\newline
      \textbf{Signal:} distinguishes process success from valid submission. \\
    \addlinespace[2pt]
    \textbf{Scoring failed}
    & \textbf{Trigger:} submitted file violates the evaluator schema.\newline
      \textbf{Feedback:} \texttt{scoring\_failed}; null score; evaluator-side validation error.\newline
      \textbf{Signal:} invalid artifact separated from low task performance. \\
    \addlinespace[2pt]
    \textbf{Timeout}
    & \textbf{Trigger:} long-running training exceeds the execution budget.\newline
      \textbf{Feedback:} \texttt{timeout}; null score; timeout metadata; partial stdout.\newline
      \textbf{Signal:} resource or algorithmic inefficiency with retained debugging evidence. \\
    \bottomrule
  \end{tabular}
\end{table}

\textbf{End-to-end transcript.}

The following transcript traces a failed GPU job from the submitted program through execution, the structured result, and the diagnostic traceback needed to repair it.

\noindent\textbf{Step 1: Agent request.}
The agent submits a generated Python program together with the task data directory and resource constraints.
In this job, the task is brain-tumor MRI classification and the public data directory is exposed to the program through \texttt{DATA\_DIR}.

\begin{SandboxTranscript}
name: brain-tumor-detection-mri@1
code: <candidate Python program>
data_dir: <task-public-data-dir>
resource_type: gpu
gpu_count: 1
timeout: 3600
\end{SandboxTranscript}

The server materializes the submitted code as \texttt{main.py} inside the job workspace.
The excerpt below shows the part that later triggers the runtime error; unrelated dataset loading, model definition, and training code is omitted.

\begin{SandboxTranscript}
# Initialize model
device = torch.device("cuda" if torch.cuda.is_available() else "cpu")
model = BrainTumorClassifier(
    model_name="efficientnet_b0",
    pretrained=True,
    dropout=0.3,
).to(device)

# Loss function with label smoothing
criterion = nn.BCEWithLogitsLoss(label_smoothing=0.1)

optimizer = optim.AdamW(model.parameters(), lr=1e-4, weight_decay=1e-4)
...
\end{SandboxTranscript}

\noindent\textbf{Step 2: Sandbox execution.}
The control server materializes the job workspace, assigns an idle GPU worker, and runs the candidate inside an isolated Docker worker.
The actual command binds the task data path into the process environment, captures stdout/stderr into the job workspace, and preserves the process exit code.

\begin{SandboxTranscript}
export DATA_DIR=<task-public-data-dir>
python <sandbox-job-workspace>/code/main.py
2>&1 | tee -a <sandbox-job-workspace>/sandbox_stdout.log
\end{SandboxTranscript}

\noindent\textbf{Step 3: Sandbox feedback.}
The returned record preserves machine-readable status, runtime, and artifact fields alongside the human-readable log. Because the program fails during execution, the evaluator is not invoked and the score remains null.

\begin{SandboxTranscript}
job_id: job_<id>
score: null
status: failed
logs:
  stdout_stderr: <sandbox-job-workspace>/sandbox_stdout.log
  diagnostic_excerpt: TypeError in BCEWithLogitsLoss(...)
error_type: code_execution_error
runtime_metadata:
  resource_type: gpu
  exit_code: 1
  shell_runtime: 11.88s
  evaluation: skipped because code execution failed
workspace_artifacts:
  workspace: <sandbox-job-workspace>
  submission: missing
  preserved_files: code/main.py, sandbox_stdout.log
\end{SandboxTranscript}

\noindent\textbf{Step 4: Diagnostic log.}
The key diagnostic is the traceback, which points directly to the generated line that is incompatible with the installed PyTorch API.

\begin{SandboxTranscript}
Traceback (most recent call last):
  File "<sandbox-job-workspace>/code/main.py", line 151, in <module>
    criterion = nn.BCEWithLogitsLoss(label_smoothing=0.1)
TypeError: BCEWithLogitsLoss.__init__() got an unexpected keyword argument
  'label_smoothing'
\end{SandboxTranscript}

Together, the status, error, runtime, artifact, and traceback fields localize the failure before evaluation and provide concrete evidence for the next repair action.

\section{OpenMLE-ERL Details}
\label{app:training-configs}

\subsection{SFT Data Generation}
\label{app:sft-warm-start}

\textbf{Parallel Path Sampling.}
The teachers first generate multiple independent \textsc{Draft} solutions for each standardized task, and every candidate is executed in the corresponding task sandbox. The first collection batch uses GLM-4.7. Among candidates with valid execution scores for the same task, we remove solutions with duplicate scores, rank the remaining candidates by score, and retain at most the Top-4. The released corpus contains 11,519 examples from this batch. The second batch uses both GLM-4.7 and Qwen3-30B-A3B-Thinking-2507. After execution and output-length screening, candidates from both teachers are jointly ranked within each task: a GLM-4.7 candidate is retained when it falls in the joint Top-4, whereas a Qwen candidate is retained only when it ranks first overall. Consequently, each task still contributes at most four examples rather than four GLM candidates plus an additional Qwen candidate. After corpus-level assembly, the released corpus contains 5,726 examples from this batch, comprising 5,075 GLM-4.7 and 651 Qwen examples. The two batches contribute 17,245 full-response examples in total.

\textbf{Evolutionary Path Sampling.}
Complete solutions expose final programs but do not directly supervise how execution feedback should be used to revise an existing solution. We therefore use GLM-4.7 to drive AIRA-Evo searches with \textsc{Draft}, \textsc{Improve}, \textsc{Crossover}, and \textsc{Debug}, producing search trees with parent relations, program versions, execution feedback, and task scores. A local segment begins at a \textsc{Draft}, \textsc{Improve}, or \textsc{Crossover} node and follows its consecutive \textsc{Debug} descendants along true parent--child edges; the segment ends when a branch reaches another \textsc{Draft}, \textsc{Improve}, or \textsc{Crossover} node. A \textsc{Draft} segment must end with a positive score, an \textsc{Improve} segment must outperform its parent program, and a \textsc{Crossover} segment must outperform the better of its two parents. Retained endpoints must additionally attain a bronze, silver, or gold level.

Single-step segments whose roots are themselves medal-level endpoints are retained directly. For multi-step segments, DeepSeek-V4-Pro reads each complete segment---including the task constraints, parent code, endpoint code, adjacent code changes, and stepwise execution feedback---and determines whether the strategy, program structure, or error repair introduced by each step is inherited by later steps and contributes to the final effective solution. After corpus-level assembly, the evolutionary path contributes 9,014 trajectory-step examples to the released corpus.

\textbf{Assembly and final corpus distribution.}
We normalize complete solutions and trajectory steps into the same system--user--assistant message structure and perform exact deduplication over the normalized full \texttt{messages} content. We then apply the target model's chat template to the full message sequence and exclude examples longer than 32,768 tokens. The released SFT corpus contains 26,259 training examples.

Of the released corpus, 17,245 full responses account for 65.7\%, and 9,014 trajectory steps account for 34.3\%. At the atomic-operator level, \textsc{Draft}, \textsc{Improve}, \textsc{Crossover}, and \textsc{Debug} contribute 19,436, 1,741, 742, and 4,340 examples, respectively, corresponding to 74.0\%, 6.6\%, 2.8\%, and 16.5\%. The median full-message lengths are 8,407 tokens for full responses and 14,051 tokens for trajectory steps; the latter are longer because they include parent programs, execution feedback, and local search context.

\begin{figure}[t]
  \centering
  \includegraphics[width=\linewidth]{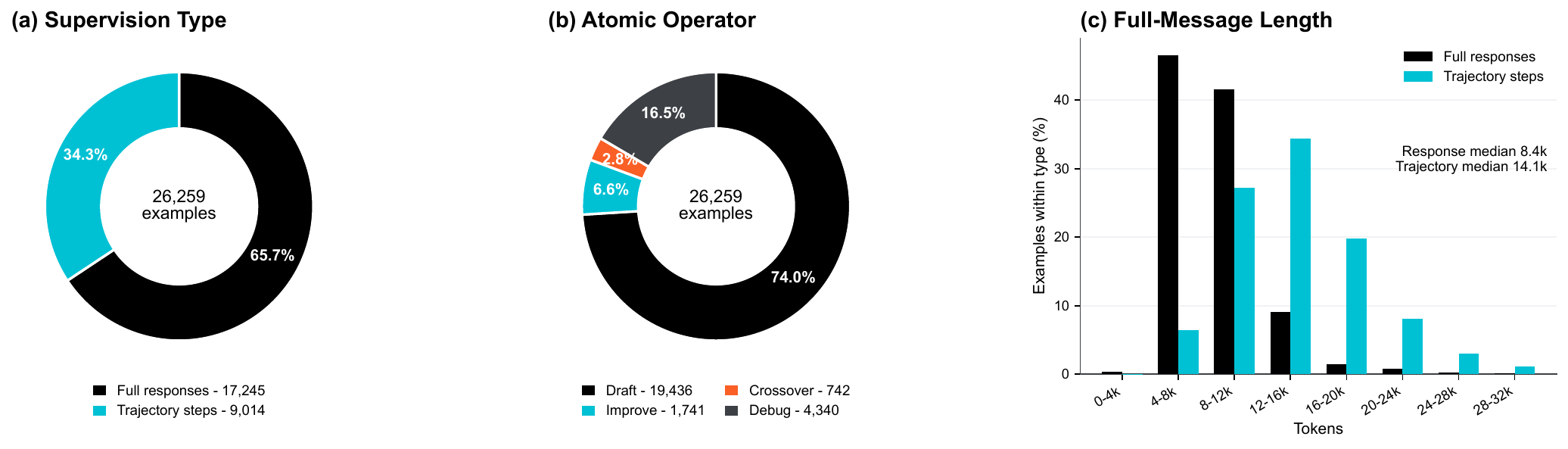}
  \caption{Distribution of the final SFT corpus. (a) Relative proportions of full-response and trajectory-step supervision among the 26,259 training examples. (b) Distribution of the four atomic operators. (c) Full-message length distributions for the two supervision types; each bar reports the percentage within its corresponding supervision type.}
  \label{fig:sft-final-data-distribution}
\end{figure}

\textbf{Prompt for trajectory-step selection.}
The annotator follows a causal-inheritance criterion: a step is retained only when its core strategy, necessary intermediate state, or critical error repair is inherited by later steps and makes a concrete contribution to the segment endpoint. Cosmetic edits, blind retries, changes that merely reduce training scale to avoid resource limits, and failed environment modifications or external-network accesses are discarded. We use DeepSeek-V4-Pro with temperature 0 and a maximum output length of 4,096 tokens, requiring a fixed JSON schema that accounts for every input step. The production system prompt is reproduced below.

\begin{Verbatim}
# Role
You are a machine learning SFT data annotation expert, responsible for selecting high-quality steps from Agent search-tree trajectory segments for supervised fine-tuning training.

# Core Objective
Inspect a given candidate segment and determine which steps should be kept as SFT training examples and which steps should be discarded.

# Input Structure and Usage Instructions
You will receive the following five parts of data. You must strictly reason according to a “causal chain” perspective:

1. TASK_SYSTEM_PROMPT: The original environment and constraint instructions of the task.
   Usage: Used to determine whether any step in the segment violates hard constraints (such as forbidden internet downloads, forbidden environment modifications, etc.).

2. TASK_USER_PROMPT: The original problem description of the task.
   Usage: Used to understand the task objective and assist in determining whether the Draft step builds a reasonable framework aligned with the task.

3. PARENT_CODES: Baseline code before the root step.
   Usage: Provides a reference baseline for Improve and crossover (which has two parent codes) steps, used to evaluate the substantive nature of modifications.
   Note: Draft starts from scratch and does not have PARENT_CODES.

4. FINAL_SOLUTION_CODE: The endpoint code of this segment (the most important semantic anchor).
   Core usage: It represents the final effective direction adopted by this segment. When judging whether a step is inherited, do not check whether the endpoint code literally contains the step’s code. Instead, evaluate whether the “core problem solved” or the “core strategy proposed” by the step remains a necessary component of the final solution.

5. TRAJECTORY_STEPS: The list of steps to be evaluated (JSON array).
   The first element of this array is the root step (operator is Draft, Improve, or Crossover), and all subsequent elements are continuous Debug steps.
   Each step contains the following fields:
   - step_index: The unique identifier of the step.
   - operator: The operator type, taking values draft, improve, crossover, or debug.
   - code content: For root steps (Draft/Improve/Crossover), the code field contains the full code generated at that step. For Debug steps, a full_code_diff field is provided in unified diff format, but it fully contains the complete updated source code after modification without omission or truncation.
   - feedback: Execution feedback after running the step, including status (success/failure), error messages, or validation scores.
   During evaluation, you must reconstruct the actual code changes from code or full_code_diff and judge the value of each step.

# Special Error Handling Principle
If the feedback of a step shows that its failure belongs to the following category, then the next step (which must be a Debug step) must be marked as keep_for_sft: false.

This error refers to a mismatch in C-extension binary interfaces or precompiled library versions. Typical patterns include:
- ValueError: numpy.dtype size changed, may indicate binary incompatibility. Expected X from C header, got Y from PyObject
- ImportError: this version of pandas is incompatible with numpy < X.X

# Global Hard Filtering Rules
If any of the following behaviors appear in the code, the step must be unconditionally marked as keep_for_sft: false, regardless of how strong the algorithm is:

- Environment modification: calling pip install, conda install, subprocess-based installation, modifying sys.path, overriding installed package versions, or using os.system for system-level commands unrelated to data processing.
- External resource downloading caused failure: feedback explicitly shows that the step attempted to download pretrained models, datasets, or external resources using wget, curl, requests.get, urllib.request, torch.hub.load (non-cached), transformers.AutoModel.from_pretrained (first-time download), and the failure is caused by download failure, timeout, permission denial, or offline environment restriction.
- Network access attempts caused failure: feedback explicitly shows that the step attempted to access external servers, non-local network addresses, or external services, and failed due to connection errors, timeouts, DNS/SSL/HTTP failures, or authentication issues.

# Golden Principle: Causal Inheritance Chain

Before applying operator-specific rules, perform the following reasoning steps:

1. Distinguish “Goal” vs “Method”:
   If a step proposes a specific method (e.g., using LightGBM instead of NN to address overfitting), while the endpoint retains the goal “solve overfitting” but uses a completely different method (e.g., HistGradientBoosting), then the specific method is NOT considered inherited.

   If the endpoint preserves the core mechanism introduced by the step (e.g., a Debug step fixes sequence alignment via truncation, and the endpoint still relies on sequence alignment even if truncation strategy is adjusted), then it is considered inherited.

2. Identify “Necessary Intermediate States”:
   If a step fixes a blocking execution error, and all subsequent steps depend on the corrected state to function, then this step MUST be marked as kept.

   If a step only performs performance tuning (e.g., reducing folds, epochs, or iterations), and is fully replaced by another tuning strategy later, then it should be discarded.

3. Distinguish “Strategy Contribution” vs “Implementation Fix”:
   Root steps (Draft / Improve / Crossover) are evaluated based on whether they introduce a core strategy. If such a step initially fails due to implementation bugs but is later fixed by Debug steps and preserved in the endpoint, it should still be kept. Debug steps are evaluated separately based on their fix contribution.

# Operator-Specific Evaluation Rules (must follow causal inheritance chain)

## Draft Rules
Keep condition: The step builds an initial solution framework aligned with the task description (including key components such as data loading, feature processing, model training, validation, inference, and submission generation), and this framework provides the main starting direction for the endpoint solution.

Drop conditions:
- The framework is hollow (only template code without task-specific logic).
- The core algorithmic direction is completely abandoned in the endpoint code (e.g., switching from CV object detection to NLP classification).
- Note: Draft must NOT be dropped due to later performance optimizations such as acceleration or library replacement.

## Improve Rules
Keep condition: Introduces a substantive strategy targeting specific issues (such as overfitting, feature insufficiency, or model bias), and the core idea is still preserved in the endpoint code (e.g., switching from MLP to tree-based models to solve overfitting, where tree-based models are still used in the endpoint, even if implementation differs).

Drop conditions:
- Non-substantive modifications (variable renaming, comments, logging, formatting).
- Pure hyperparameter tuning without strategy-level change.
- The improvement direction is explicitly removed or reverted in the endpoint.
- The endpoint only preserves a high-level theme (e.g., still ML or still tree-based models) but does NOT preserve the specific method introduced by this step.

Important:
- Evaluate Improve based on whether the concrete method is preserved, not whether the abstract goal is consistent.
- If Improve initially fails due to implementation bugs but its method is later fixed by Debug steps and preserved in the endpoint, it must be kept.

## Crossover Rules
Keep condition: Effectively combines ideas from two parent branches (e.g., features from A and models from B), and this fusion direction is adopted in the endpoint code.

Drop conditions:
- Only copies one branch without real fusion.
- The endpoint abandons the fusion direction and follows an unrelated third path.
- Note: If the step fails due to environment issues but is later corrected while preserving the fusion strategy, it must still be kept.

## Debug Rules
Keep condition: The step accurately identifies and fixes a blocking error, or replaces an infeasible previous approach with a feasible alternative; and this fix or alternative is inherited by later steps and becomes a key prerequisite for the endpoint’s effective path.

Drop conditions:
- Blind retry without identifying the concrete failure cause.
- Only adds logging, printing, comments, or formatting without fixing the underlying issue.
- Low-value mechanical edits, such as a one-line typo fix or tiny index change, unless they form a key reusable fix.
- Only avoids timeout or memory issues by reducing epochs, folds, batch size, sample size, feature count, training rounds, or model complexity; even if this step is the endpoint and runs successfully, it should be dropped unless it also introduces a reusable algorithmic or engineering fix.
- Temporary environment or old-API compatibility workaround that is later replaced by a more complete refactor in the endpoint.
- The fix or alternative is later rewritten, bypassed, or no longer depended on, so it is not on the endpoint’s successful path.
- Only leaves local code remnants or architectural similarity in the endpoint, but is not a key prerequisite for successful execution or core performance.

Notes:
- A Debug step does not have to be the final successful version itself. If its key fix is inherited by later steps and enters the endpoint, it should still be kept.
- A Debug step does not have to preserve the previous method. If the previous approach is infeasible under the task constraints, switching to a feasible alternative inherited by the endpoint should also be kept.

# Output Format
Must output valid JSON only, in the following format:
{
  "steps": [
    {
      "step_index": 27,
      "operator": "Improve",
      "keep_for_sft": true,
      "usefulness_reason": "Concise explanation in Chinese describing causal contribution or irrelevance to the endpoint."
    }
  ]
}

# Output Constraints
- The steps array must include every input step exactly once, in input order, with no omissions.
- The operator field must strictly match Draft, Improve, Crossover, or Debug.
- The usefulness_reason must be specific to the actual step content and avoid generic phrases.
- Do not output Markdown code blocks, only raw JSON.
\end{Verbatim}

Each request instantiates the following input template. \texttt{PARENT\_CODES} is omitted for \textsc{Draft}, contains one parent for \textsc{Improve}, and contains two parents for \textsc{Crossover}.

\begin{SandboxTranscript}
Annotate the following candidate trajectory using system instructions.

<TASK_SYSTEM_PROMPT>
{original task environment and constraints}
</TASK_SYSTEM_PROMPT>

<TASK_USER_PROMPT>
{original task description}
</TASK_USER_PROMPT>

<PARENT_CODES>
{one parent for Improve; two parents for Crossover; omitted for Draft}
</PARENT_CODES>

<FINAL_SOLUTION_CODE>
{complete endpoint code}
</FINAL_SOLUTION_CODE>

<TRAJECTORY_STEPS>
{root full code, subsequent full-file diffs, and execution feedback}
</TRAJECTORY_STEPS>
\end{SandboxTranscript}

\subsection{SFT Training Configuration}

Table~\ref{tab:sft-warm-start-config} summarizes the core training settings used for the executable SFT warm starts of \thirtybmodel{} and \thirtyfivebmodel{}.

\begin{table}[h!]
\centering
\caption{Core SFT hyperparameters for \thirtybmodel{} and \thirtyfivebmodel{}. Settings shared by both models are centered across the two model columns.}
\label{tab:sft-warm-start-config}
\vspace{0.4em}
\footnotesize
\setlength{\tabcolsep}{3.5pt}
\renewcommand{\arraystretch}{1.08}
\begin{tabular}{@{}>{\raggedright\arraybackslash}p{0.18\linewidth}>{\raggedright\arraybackslash}p{0.37\linewidth}>{\raggedright\arraybackslash}p{0.37\linewidth}@{}}
\toprule
Item & \thirtybmodel{} & \thirtyfivebmodel{} \\
\midrule
Base model & Qwen3-30B-A3B-Thinking-2507 & Qwen3.6-35B-A3B \\
\addlinespace[3pt]
Training stage & \multicolumn{2}{>{\centering\arraybackslash}p{0.76\linewidth}}{Full-parameter SFT} \\
\addlinespace[3pt]
Training framework & \multicolumn{2}{>{\centering\arraybackslash}p{0.76\linewidth}}{SLIME with Ray and Megatron-LM} \\
\addlinespace[3pt]
\addlinespace[3pt]
Template / thinking mode & \texttt{qwen3} loss mask; \texttt{<think>} supervision retained & \texttt{qwen3\_5}-compatible loss mask; \texttt{<think>} supervision retained \\
\addlinespace[3pt]
Context cutoff & \multicolumn{2}{>{\centering\arraybackslash}p{0.76\linewidth}}{32,768 tokens} \\
\addlinespace[3pt]
Precision & \multicolumn{2}{>{\centering\arraybackslash}p{0.76\linewidth}}{bfloat16} \\
\addlinespace[3pt]
Global batch size & \multicolumn{2}{>{\centering\arraybackslash}p{0.76\linewidth}}{128} \\
\addlinespace[3pt]
Per-device batch size & \multicolumn{2}{>{\centering\arraybackslash}p{0.76\linewidth}}{1 with dynamic batching} \\
\addlinespace[3pt]
Gradient accumulation & 64 microbatches per update & 32 microbatches per update \\
\addlinespace[3pt]
Learning rate & \multicolumn{2}{>{\centering\arraybackslash}p{0.76\linewidth}}{$3.0\times10^{-5}$} \\
\addlinespace[3pt]
Scheduler / warmup & \multicolumn{2}{>{\centering\arraybackslash}p{0.76\linewidth}}{cosine decay to 0, 0.1 warmup fraction} \\
\addlinespace[3pt]
Epochs & \multicolumn{2}{>{\centering\arraybackslash}p{0.76\linewidth}}{3} \\
\bottomrule
\end{tabular}
\end{table}

\Needspace{0.58\textheight}
\subsection{RL Training Configuration}

Table~\ref{tab:rl-training-config} summarizes the verified RL settings for \thirtybmodel{} and \thirtyfivebmodel{} while retaining the original configuration fields.

\begin{table}[h!]
\centering
\caption{RL hyperparameters for \thirtybmodel{} and \thirtyfivebmodel{}. Settings shared by both models are centered across the two model columns.}
\label{tab:rl-training-config}
\vspace{0.4em}
\footnotesize
\setlength{\tabcolsep}{3.2pt}
\renewcommand{\arraystretch}{1.06}
\begin{tabular}{@{}>{\raggedright\arraybackslash}p{0.18\linewidth}>{\raggedright\arraybackslash}p{0.37\linewidth}>{\raggedright\arraybackslash}p{0.37\linewidth}@{}}
\toprule
Item & \thirtybmodel{} & \thirtyfivebmodel{} \\
\midrule
RL initialization & SFT warm-start checkpoint trained from Qwen3-30B-A3B-Thinking-2507 & SFT warm-start checkpoint trained from Qwen3.6-35B-A3B \\
\addlinespace[3pt]
Training framework & \multicolumn{2}{>{\centering\arraybackslash}p{0.76\linewidth}}{SLIME with Ray and SGLang} \\
\addlinespace[3pt]
\addlinespace[3pt]
Operator sampling probability & \multicolumn{2}{>{\centering\arraybackslash}p{0.76\linewidth}}{\textsc{Draft} 0.50, \textsc{Improve} 0.17, \textsc{Debug} 0.17, \textsc{Crossover} 0.16} \\
\addlinespace[3pt]
Rollout group & \multicolumn{2}{>{\centering\arraybackslash}p{0.76\linewidth}}{16 prompts per rollout, 16 samples per prompt, global batch size 128, 2 optimizer steps per rollout} \\
\addlinespace[3pt]
Generation & \multicolumn{2}{>{\centering\arraybackslash}p{0.76\linewidth}}{temperature 1.0, maximum response length 24,576 tokens} \\
\addlinespace[3pt]
\addlinespace[3pt]
Advantage / objective & \multicolumn{2}{>{\centering\arraybackslash}p{0.76\linewidth}}{GSPO with TTT-Discover-style reward post-processing, clip $\epsilon=3.5\times10^{-4}$, TIS enabled} \\
\addlinespace[3pt]
Optimizer & \multicolumn{2}{>{\centering\arraybackslash}p{0.76\linewidth}}{Adam, learning rate $1.0\times10^{-6}$, constant schedule, weight decay 0.1, $\beta_1=0.9$, $\beta_2=0.98$} \\
\bottomrule
\end{tabular}
\end{table}

\subsection{Asynchronous Rollout}
\label{app:async-rollouts}

\textbf{Implementation.}
A synchronous rollout step must wait for all executions in the batch, so slow sandbox jobs can leave training resources idle.
OpenMLE instead uses a fully asynchronous rollout worker that draws task groups from the data source, launches generation-and-reward jobs independently, and pushes completed groups into a queue consumed by the trainer.

\begin{figure}[h!]
  \centering
  \includegraphics[width=0.55\linewidth]{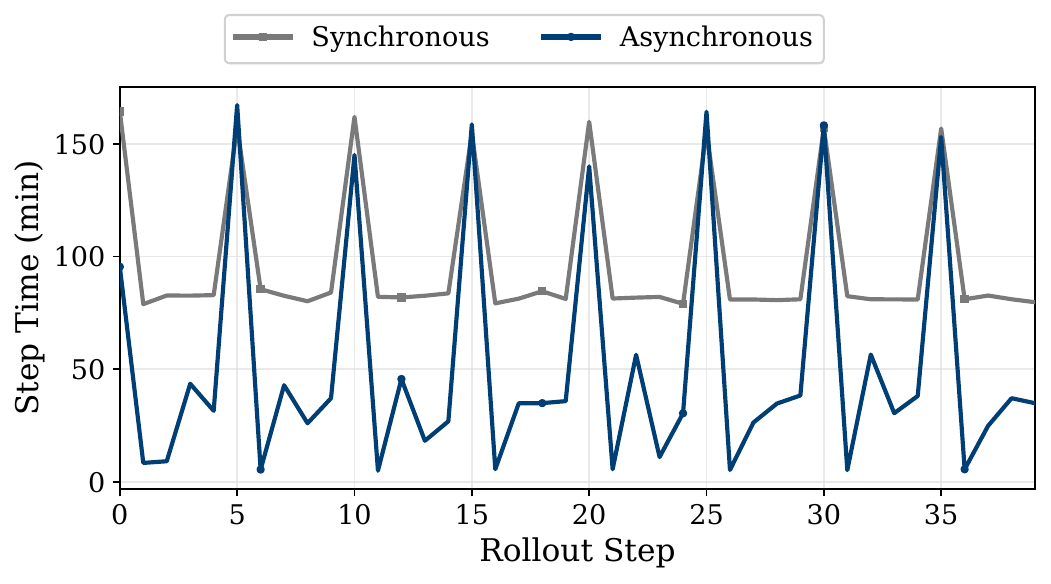}
  \caption{Wall-clock step-time comparison for synchronous and asynchronous rollout collection over 40 matched rollout steps. Mean step time is 97.0 minutes for the synchronous run and 50.8 minutes for the asynchronous run.}
  \label{fig:async-vs-sync-step-time}
\end{figure}

\textbf{Wall-clock benefit and task balance.}
Figure~\ref{fig:async-vs-sync-step-time} compares measured wall-clock step time for synchronous and asynchronous rollouts.
Across the 40 matched steps, mean step time is 97.0 minutes for the synchronous run and 50.8 minutes for the asynchronous run, corresponding to a $1.91\times$ ratio.
Although asynchronous collection can consume fast or immediately failing tasks more often under a fixed wall-clock budget, task exposure remains balanced in practice: in two representative asynchronous runs, per-task step counts stay within $\pm 2$ steps of the run median, with coefficients of variation of $1.56\%$ and $2.06\%$.

\subsection{Reward Normalization and Entropic Advantage}
\label{app:reward-advantage}

We distinguish three quantities: a score-derived \emph{base reward} used for logging and static comparisons, an adaptive-bound \emph{processed reward} used as the group reward when dynamic bounds are enabled, and an entropic processed advantage returned by the reward post-processing hook for RL updates.
For a raw task score $s$, define the signed score
\begin{equation}
\label{eq:signscore}
z =
\begin{cases}
s, & \text{if larger raw scores are better},\\
-s, & \text{otherwise}.
\end{cases}
\end{equation}
Static signed bounds are computed from task metadata after the same sign conversion. For larger-is-better metrics,
\begin{equation}
\label{eq:metric}
\begin{aligned}
B_{\mathrm{static}} &= \max(\texttt{theoretical\_max},\, \texttt{leaderboard\_max}), \\
W_{\mathrm{static}} &= \min(\texttt{theoretical\_min},\, \texttt{leaderboard\_min}),
\end{aligned}
\end{equation}
ignoring missing values; lower-is-better metrics use the corresponding signed versions of these quantities.

\textbf{Adaptive score range and processed reward.}
Fixed leaderboard or theoretical bounds can be much wider than the scores produced by the current policy, making two meaningfully different programs receive nearly the same reward. We instead rescale each task using a moving score range built from successful historical programs and the current rollout group. After converting every metric so that larger values are better, we sort the available scores as
\[
x_{(1)} \ge x_{(2)} \ge \cdots \ge x_{(K)} .
\]
The best observed score sets the upper end of the range. The 16th-best score sets the lower reference point; when fewer than 16 scores are available, we use the lowest available score:
\[
B_{\mathrm{dyn}} = x_{(1)},\qquad
W_{\mathrm{dyn}} = x_{(\min(16,K))}.
\]
We then extend the lower end downward by one quarter of the gap between these two reference scores. This keeps moderately successful programs from being clipped to zero when the observed scores are tightly clustered:
\[
W_{\mathrm{dyn}} \leftarrow W_{\mathrm{dyn}} - 0.25 \max(B_{\mathrm{dyn}}-W_{\mathrm{dyn}},0).
\]
When task metadata provides valid theoretical or leaderboard limits, they prevent the moving range from extending beyond the task's valid score range:
\[
B=\min(B_{\mathrm{dyn}},B_{\mathrm{static}}),\qquad
W=\max(W_{\mathrm{dyn}},W_{\mathrm{static}}),
\]
with fallback to static bounds if the resulting pair is invalid.
The resolved range maps scores to $[0,1]$ using the same transformation as Eq.~\ref{eq:base-reward}; scores below the lower end receive zero:
\[
r_{\mathrm{proc}}(z)=r_{\mathrm{base}}(z;B,W).
\]
Because the range is recomputed from recent and historical policy outputs, it follows the score frontier as training improves and preserves useful reward differences among current candidates. The implementation retains both static and adaptive reward views in metadata, while the adaptive view supplies the reward used by the main RL configuration.

\textbf{Entropic group advantages.}
This path replaces the usual GRPO-style normalization of group rewards into advantages.
When entropic post-processing is enabled, rewards are grouped by prompt group before advantage computation.
For group processed rewards $r_{\mathrm{proc},1},\ldots,r_{\mathrm{proc},K}$, the implementation returns zero advantages if $K<2$ or all rewards are equal.
Otherwise it centers rewards by $c_i=r_{\mathrm{proc},i}-\max_j r_{\mathrm{proc},j}$ and defines
\begin{equation}
\label{eq:entr}
q_i(\beta)=\frac{\exp(\beta c_i)}{\sum_{j=1}^{K}\exp(\beta c_j)}.
\end{equation}
The scalar $\beta$ is chosen by binary search so that
\begin{equation}
\label{eq:kl}
\mathrm{KL}\!\left(q_\beta\,\|\,\mathrm{Unif}(K)\right)
= \sum_{i=1}^{K} q_i(\beta)\left(\log q_i(\beta)+\log K\right)
\approx \log 2,
\end{equation}
with maximum search value $10^6$ and 60 bisection iterations.
The returned advantage for sample $i$ uses a leave-one-out denominator,
\[
e_i=\exp(\beta c_i),\qquad
Z_{-i}=\frac{1}{K-1}\sum_{j\ne i} e_j,\qquad
A_i=\frac{e_i}{Z_{-i}+10^{-12}}-1.
\]
The post-processing hook returns the input rewards as \texttt{raw\_rewards} and these $A_i$ values as \texttt{processed\_rewards}.

\subsection{Detection and prevention of reward hacking.}

During the training process, we observed that our models—particularly the smaller ones used in our early experiments—experience a specific issue during RL on difficult tasks. The reward quickly plateaus at a very low level.

Based on our case study, we discovered that the models are exhibiting significant reward hacking behavior. A representative example of this is when a model takes the sample submission, randomly shuffles it, and submits it as a solution.

To mitigate this, we implemented the following workflow:
1. We use o3-mini as an LLM judge during the RL process.
2. Before the code is executed in the sandbox, the judge performs a reward hack check.
3. If reward hacking is detected, the code bypasses sandbox execution and is assigned a reward of -0.5.

\section{\openmleevo{} Inference Details}

\subsection{Evolutionary Parent Fitness}
\label{app:evolutionary-fitness}

For a task $\tau$, let $\mathcal{P}_{\tau}$ be the set of programs stored in the task-local program database.
The implementation recomputes parent fitness at the task level after each insertion.
For each node $p\in\mathcal{P}_{\tau}$, define its linked children as
\[
\mathcal{C}(p)=\{c:\mathrm{parent}(c)=p\}\cup\{c:p\in \mathrm{crossover\_parents}(c)\}.
\]
The three raw components are:
\[
U_p = R^{\mathrm{base}}_p,
\]
\[
L_p =
\begin{cases}
\varnothing, & |\mathcal{C}(p)|=0,\\
0, & |\mathcal{C}(p)|=1,\\
\frac{1}{|\mathcal{C}(p)|}\sum_{c\in\mathcal{C}(p)}
\left(R^{\mathrm{base}}_c-\frac{1}{|\mathcal{C}(p)|}\sum_{c'\in\mathcal{C}(p)}R^{\mathrm{base}}_{c'}\right)^2,
& |\mathcal{C}(p)|\ge 2,
\end{cases}
\]
and
\[
V_p=|\mathcal{C}(p)|,\qquad
C^{\mathrm{raw}}_p =
\begin{cases}
1, & \max_{p'\in\mathcal{P}_{\tau}} V_{p'}=0,\\
\max\!\left(0,1-\frac{V_p}{\max_{p'\in\mathcal{P}_{\tau}}V_{p'}}\right), & \text{otherwise}.
\end{cases}
\]
For finite task-level values $\{a_p\}$, min--max normalization is
\[
N_{\eta}(a_p)=
\begin{cases}
\eta, & \max_{p'}a_{p'}-\min_{p'}a_{p'} < 10^{-12},\\
\frac{a_p-\min_{p'}a_{p'}}{\max_{p'}a_{p'}-\min_{p'}a_{p'}}, & \text{otherwise}.
\end{cases}
\]
The learning term uses optional normalization: missing $L_p=\varnothing$ receives the neutral value $\eta=0.5$, while finite learning values are normalized over only the finite entries.
Equivalently,
\[
N_{\eta}^{\mathrm{opt}}(L_p)=
\begin{cases}
\eta, & L_p=\varnothing,\\
N_{\eta}(L_p), & \text{otherwise, with } N_{\eta} \text{ computed over finite } \{L_{p'}\}.
\end{cases}
\]
The implemented fitness is
\[
F(p)=N_{0.5}(U_p)+N_{0.5}^{\mathrm{opt}}(L_p)+N_{0.5}(C^{\mathrm{raw}}_p).
\]
For \textsc{Improve} and \textsc{Crossover}, the candidate set is restricted to programs with positive stored reward; for \textsc{Debug}, it is restricted to programs with non-positive stored reward.
Within the candidate set $\mathcal{S}$, selection is roulette-wheel sampling without replacement:
\[
\Pr(p\mid \mathcal{S})=
\begin{cases}
\frac{\max(F(p),0)}{\sum_{p'\in\mathcal{S}}\max(F(p'),0)}, & \sum_{p'\in\mathcal{S}}\max(F(p'),0)>0,\\
\frac{1}{|\mathcal{S}|}, & \text{otherwise}.
\end{cases}
\]
Crossover draws two parents by applying this rule sequentially and removing the first sampled parent before the second draw.

\subsection{Operator Prompt Templates}
\label{app:openmleevo-operator-prompts}

\openmleevo{} uses four generation operators with distinct lineage context:
\textsc{Draft} proposes a new solution from the task specification,
\textsc{Improve} revises one selected parent, \textsc{Crossover} synthesizes two
selected parents, and \textsc{Debug} repairs a failed or non-positive-reward
parent. We reproduce below the complete system- and user-message schemas used
for MLE-Bench inference. Double-braced expressions denote task- or
attempt-specific runtime fields. To make the templates independent of a
particular search trajectory, optional retrieved memory and runtime data-preview
content are left empty; consequently, the empty-memory fallback text is shown.

\paragraph{\textsc{Draft}.}
\begin{Verbatim}
[SYSTEM MESSAGE]

You are a Kaggle Grandmaster attending a high-stakes competition. In order to win this competition, you need to come up with an excellent and creative plan for a solution and then implement this solution in Python.
{{public_system_prompt}}

[USER MESSAGE TEMPLATE]

We will now provide a description of the task.

Task Description:

{{task_description}}

Data Description:

{{data_description}}

Draft Guidance:

Please focus on proposing an advanced idea addressing this task.
Propose exactly one new solution idea that is different from the previously explored ideas.
Keep the evaluation method consistent across iterations.

AIRA Evo Search Context:

Previously explored ideas: none yet.

Execution Budget:

Be aware of the running time of the code, it should complete within {{execution_timeout}}.
Your code will be executed in a sandbox after generation.
A single sandbox run must finish within {{execution_timeout}}.
Always choose a validation/training strategy that leaves enough time to write ./submission.csv.

Submission Check:

Before finishing, verify that ./submission.csv exists, has the same row count and required columns/order as sample_submission.csv when available, preserves the sample id/order, and contains no NaN/inf values.
When building features, align train/test columns explicitly and handle missing values before model fitting or prediction.

{{public_user_prompt}}
\end{Verbatim}

\paragraph{\textsc{Improve}.}
\begin{Verbatim}
[SYSTEM MESSAGE]

You are a Kaggle Grandmaster attending a high-stakes competition. In order to win this competition, you need to come up with an excellent and creative plan for a solution and then implement this solution in Python, that improves upon an existing solution to the task.
{{public_system_prompt}}

[USER MESSAGE TEMPLATE]

We will now provide a description of the task.

Task Description:

{{task_description}}

Data Description:

{{data_description}}

A previous solution was attempted with the following results:

Code:

```python
{{previous_code}}
```

Feedback:

```
{{previous_terminal_output_with_official_sandbox_score_redacted}}
```

Improve Guidance:

Based on the previous attempt, its score, and execution output, please analyze what worked well and what didn't, then improve the solution.
Propose exactly one improvement idea that is different from the previously explored improvement ideas.
Keep the evaluation method consistent across iterations.

AIRA Evo Search Context:

Targeted memory for this improvement: none available.

Execution Budget:

Be aware of the running time of the code, it should complete within {{execution_timeout}}.
Your code will be executed in a sandbox after generation.
A single sandbox run must finish within {{execution_timeout}}.
Always choose a validation/training strategy that leaves enough time to write ./submission.csv.

Submission Check:

Before finishing, verify that ./submission.csv exists, has the same row count and required columns/order as sample_submission.csv when available, preserves the sample id/order, and contains no NaN/inf values.
When building features, align train/test columns explicitly and handle missing values before model fitting or prediction.

{{public_user_prompt}}
\end{Verbatim}

\paragraph{\textsc{Crossover}.}
\begin{Verbatim}
[SYSTEM MESSAGE]

You are a Kaggle Grandmaster attending a high-stakes competition. Your goal is to synthesize the strengths of two provided solutions, neutralize their respective weaknesses, and propose novel improvements that neither original solution possessed. A simple "merge" of existing code won't suffice to win. You must identify the hidden synergies between previous attempts and engineering a third-generation solution that transcends its predecessors.
{{public_system_prompt}}

[USER MESSAGE TEMPLATE]

We will now provide a description of the task.

Task Description:

{{task_description}}

Data Description:

{{data_description}}

Two previous solutions were attempted with the following results:

Code 1:

```python
{{previous_code_1}}
```

Feedback 1:

```
{{previous_terminal_output_1_with_official_sandbox_score_redacted}}
```

Code 2:

```python
{{previous_code_2}}
```

Feedback 2:

```
{{previous_terminal_output_2_with_official_sandbox_score_redacted}}
```

Crossover Guidance:

Based on the previous attempts, their scores, and execution outputs, please synthesize the winning components of both solutions into a superior hybrid while introducing novel improvements to address shared weaknesses.
Propose exactly one crossover plan and keep the evaluation method consistent across iterations.

AIRA Evo Search Context:

Targeted memory for this crossover: none available.

Execution Budget:

Be aware of the running time of the code, it should complete within {{execution_timeout}}.
Your code will be executed in a sandbox after generation.
A single sandbox run must finish within {{execution_timeout}}.
Always choose a validation/training strategy that leaves enough time to write ./submission.csv.

Submission Check:

Before finishing, verify that ./submission.csv exists, has the same row count and required columns/order as sample_submission.csv when available, preserves the sample id/order, and contains no NaN/inf values.
When building features, align train/test columns explicitly and handle missing values before model fitting or prediction.

{{public_user_prompt}}
\end{Verbatim}

\paragraph{\textsc{Debug}.}
\begin{Verbatim}
[SYSTEM MESSAGE]

You are a Kaggle Grandmaster attending a high-stakes competition. In order to win this competition, you need to come up with an excellent and creative plan for a solution and then implement this solution in Python, that debugs and fixes an existing solution to the task.
{{public_system_prompt}}

[USER MESSAGE TEMPLATE]

We will now provide a description of the task.

Task Description:

{{task_description}}

Data Description:

{{data_description}}

A previous solution was attempted with the following results:

Code:

```python
{{previous_buggy_code}}
```

Feedback:

```
{{execution_output_with_official_sandbox_score_redacted}}
```

Debug Guidance:

Based on the previous attempt, its score, and execution output, please identify the root cause of the failure and fix the code so it runs successfully and follows the output requirements. After it runs, improve the solution if possible.
Preserve the current core idea unless the feedback shows that it cannot be made valid.

AIRA Evo Search Context:

Targeted debug memory: none available.

Execution Budget:

Be aware of the running time of the code, it should complete within {{execution_timeout}}.
Your code will be executed in a sandbox after generation.
A single sandbox run must finish within {{execution_timeout}}.
Always choose a validation/training strategy that leaves enough time to write ./submission.csv.
If the previous attempt timed out, reduce compute first: fewer epochs/folds, smaller input size, smaller model, cached features, or simpler ensembling.

Submission Check:

Before finishing, verify that ./submission.csv exists, has the same row count and required columns/order as sample_submission.csv when available, preserves the sample id/order, and contains no NaN/inf values.
When building features, align train/test columns explicitly and handle missing values before model fitting or prediction.

{{public_user_prompt}}
\end{Verbatim}

\subsection{Structured Experience Records}
\label{app:openmleevo-experience-records}

\openmleevo{} represents inference-time search state with two deterministic structures. An \emph{experience card} is created after every sandbox evaluation and remains attached to that node. The task-level \emph{experience board} is recomputed from all accumulated cards whenever the controller needs global search state. Table~\ref{tab:openmleevo-experience-card-schema} gives the complete card content, and Table~\ref{tab:openmleevo-experience-board-schema} gives the board aggregation. We use \emph{experience board} consistently throughout the paper. The current evaluator persists the same object under the historical filename \nolinkurl{strategy_board.json} for backward compatibility; this filename does not denote a separate conceptual structure.

\begingroup
\small
\setlength{\tabcolsep}{5pt}
\renewcommand{\arraystretch}{1.12}
\setlength{\LTcapwidth}{\textwidth}
\begin{longtable}{@{}>{\raggedright\arraybackslash}p{0.17\textwidth}>{\raggedright\arraybackslash}p{0.39\textwidth}>{\raggedright\arraybackslash}p{0.38\textwidth}@{}}
\caption{Deterministic node-level experience-card schema in \openmleevo{}. The fields are populated from the program node, sandbox result, and usage record rather than inferred from test outcomes.}
\label{tab:openmleevo-experience-card-schema}\\
\toprule
\textbf{Card component} & \textbf{Stored fields} & \textbf{Role in search} \\
\midrule
\endfirsthead
\multicolumn{3}{c}{\tablename\ \thetable\ (continued)}\\
\toprule
\textbf{Card component} & \textbf{Stored fields} & \textbf{Role in search} \\
\midrule
\endhead
\midrule
\multicolumn{3}{r}{Continued on next page}\\
\endfoot
\bottomrule
\endlastfoot
Identity and lineage & \nolinkurl{schema_version}, \nolinkurl{node_id}, \nolinkurl{step_id}, \nolinkurl{operator}, \nolinkurl{parents}, \nolinkurl{parent_node_ids}, \nolinkurl{generation_id} & Identifies the evaluated program and reconstructs its incoming search edges. \\
Observed outcome & \nolinkurl{score}, \nolinkurl{fitness}, \nolinkurl{reward}, \nolinkurl{status}, \nolinkurl{status_code}, \nolinkurl{is_buggy}, \nolinkurl{error_signature} & Records the validation result and a normalized failure signature derived from execution status and logs. \\
Resource accounting & \nolinkurl{sandbox_time_used}, \nolinkurl{model_time_used}, \nolinkurl{model_plus_sandbox_time_used}, \nolinkurl{cost}, and prompt/completion/total token counts & Exposes the computational cost of the attempted method and supports budget-aware reasoning. \\
Method characterization & \nolinkurl{imports}, \nolinkurl{method_family_auto}, \nolinkurl{family_count_before} & Detects a coarse modeling family from the program and measures how often that direction has already been explored. \\
Derived search signals & \nolinkurl{delta_vs_parent}, \nolinkurl{novelty_score}, \nolinkurl{is_new_direction}, \nolinkurl{rank}, \nolinkurl{current_best}, \nolinkurl{selection_utility} & Supplies the progress and novelty factors used alongside validation quality, plus an auditable parent-selection trace. \\
Semantic evidence & \nolinkurl{plan}, \nolinkurl{analysis}, and optional \nolinkurl{rich_summary} with \nolinkurl{method_overview} and \nolinkurl{parent_comparison_experience} & Retains the model proposal and execution analysis; richer reusable summaries are generated lazily only when this node is retrieved. \\
\end{longtable}
\endgroup

\begingroup
\footnotesize
\setlength{\tabcolsep}{5pt}
\renewcommand{\arraystretch}{1.12}
\setlength{\LTcapwidth}{\textwidth}
\begin{longtable}{@{}>{\raggedright\arraybackslash}p{0.17\textwidth}>{\raggedright\arraybackslash}p{0.39\textwidth}>{\raggedright\arraybackslash}p{0.38\textwidth}@{}}
\caption{Task-global experience-board schema. The board is a deterministic aggregation of the cards available at the current search step.}
\label{tab:openmleevo-experience-board-schema}\\
\toprule
\textbf{Board component} & \textbf{Aggregated fields} & \textbf{Information exposed to the controller} \\
\midrule
\endfirsthead
\multicolumn{3}{c}{\tablename\ \thetable\ (continued)}\\
\toprule
\textbf{Board component} & \textbf{Aggregated fields} & \textbf{Information exposed to the controller} \\
\midrule
\endhead
\midrule
\multicolumn{3}{r}{Continued on next page}\\
\endfoot
\bottomrule
\endlastfoot
Global incumbent & \nolinkurl{num_nodes}, \nolinkurl{best_node}, \nolinkurl{best_score}, \nolinkurl{current_best_family} & The size and strongest valid member of the current population. \\
Method-family coverage & \nolinkurl{method_family_stats}, \nolinkurl{family_best_nodes}, \nolinkurl{underexplored_families} & Per-family counts, valid/failure counts, best score and node, failure rate, and low-coverage directions. \\
Progress and failures & \nolinkurl{score_history}, \nolinkurl{recent_delta_trend}, \nolinkurl{repeated_errors}, \nolinkurl{status_count}, \nolinkurl{operator_counts} & Recent search progress, recurring error signatures, and the empirical mix of operators and outcomes. \\
Topology and node state & \nolinkurl{parent_graph}, \nolinkurl{novelty_by_node}, \nolinkurl{rank_by_node}, \nolinkurl{current_best_by_node} & Reconstructs ancestry and provides per-node novelty, rank, and incumbent indicators for retrieval. \\
Resources and audit trail & \nolinkurl{runtime_stats}, \nolinkurl{parent_selection_weights} & Aggregates model/sandbox runtime and preserves the candidate factors, utilities, probabilities, and selected IDs from parent sampling. \\
\end{longtable}
\endgroup

\paragraph{Grounded record.}
For a successful \textsc{Improve} node on \texttt{leaf-classification}, the stored card records \nolinkurl{step_id=10}, validation log loss \nolinkurl{fitness=0.012080}, \nolinkurl{delta_vs_parent=0.753352}, \nolinkurl{method_family_auto=ensemble+xgboost+neural_net+cv}, \nolinkurl{novelty_score=0.57735}, \nolinkurl{rank=1}, \nolinkurl{current_best=true}, 178.58 seconds of sandbox time, and 24,564 total model tokens. Immediately before generating this node, the experience board identified a \nolinkurl{neural_net+cv} incumbent with score 0.031825, reported three previous timeouts, and exposed a recent mean parent-relative gain of 0.01649. Thus, the card states exactly what the new node achieved, while the board states where that result sits within the surrounding population.

\subsection{Experience-Guided Parent Selection}
\label{app:openmleevo-parent-selection}

This subsection specifies the inference-time parent-selection rule used by \openmleevo{} and is distinct from the training-time parent fitness in Appendix~\ref{app:evolutionary-fitness}. For a sampled island $\mathcal{I}$, each candidate $i\in\mathcal{I}$ is associated with a deterministic experience card. Let $s_i$ denote its validation score and let $s_i^{\mathrm{par}}$ be the score of its strongest parent under the task metric. The default score component applies direction-aware min--max normalization:
\[
\tilde{s}_i =
\begin{cases}
\dfrac{s_i-s_{\min}}{s_{\max}-s_{\min}}, & \text{if higher is better},\\[6pt]
\dfrac{s_{\max}-s_i}{s_{\max}-s_{\min}}, & \text{if lower is better}.
\end{cases}
\]
When all candidate scores are equal, we set $\tilde{s}_i=0.5$. The improvement component retains only progress over the best available parent:
\[
\Delta_i =
\begin{cases}
\max(0,s_i-s_i^{\mathrm{par}}), & \text{if higher is better},\\
\max(0,s_i^{\mathrm{par}}-s_i), & \text{if lower is better},
\end{cases}
\qquad
\widetilde{\Delta}_i =
\begin{cases}
\dfrac{\Delta_i}{\max_{j\in\mathcal{I}}\Delta_j}, & \max_{j\in\mathcal{I}}\Delta_j>0,\\[6pt]
0, & \text{otherwise}.
\end{cases}
\]
Nodes without a valid parent receive zero improvement. Finally, if $N_{f_i}$ is the number of other previously recorded cards in candidate $i$'s automatically detected method family $f_i$, the novelty component is
\[
\nu_i = \frac{1}{\sqrt{1+N_{f_i}}}.
\]
The utility and temperature-controlled softmax are given in Equation~\ref{eq:openmleevo-parent-selection}. The default implementation uses min--max normalization for score and improvement, while rank-based and hybrid normalizations are configurable. For \textsc{Crossover}, two parents are sampled sequentially without replacement from the same island; for \textsc{Improve}, one parent is sampled. Final submission selection is not sampled: it deterministically chooses the best executable candidate with a valid validation outcome.

\thispagestyle{fancy}

\subsection{Operation-Triggered Memory Synthesis}
\label{app:openmleevo-memory-synthesis}

The controller stores every deterministic experience card, but it does not eagerly ask a language model to summarize every node. Once an operation and its parent nodes have been selected, it first retrieves a bounded, operation-specific evidence set (Table~\ref{tab:openmleevo-operation-memory}). For any retrieved node without a cached rich summary, the memory model receives the task description; current and parent metadata; both plans, programs, and execution outputs; score, parent-relative delta, runtime, status, and error signature. It must return exactly two JSON fields: \nolinkurl{method_overview}, a concrete description of the model, features, validation, ensembling, runtime choices, and submission logic; and \nolinkurl{parent_comparison_experience}, an evidence-based account of what changed, whether score/status/runtime improved, and what should be reused or avoided. Each synthesis call summarizes one retrieved node relative to one direct parent; when an operation retrieves multiple nodes or two crossover branches, their independently cached summaries are composed only in the subsequent operator-specific context. The resulting summary is cached in the node and its card, so later operations reuse it without another synthesis call. 

\begingroup
\footnotesize
\setlength{\tabcolsep}{5pt}
\renewcommand{\arraystretch}{1.12}
\begin{longtable}{@{}>{\raggedright\arraybackslash}p{0.13\textwidth}>{\raggedright\arraybackslash}p{0.46\textwidth}>{\raggedright\arraybackslash}p{0.35\textwidth}@{}}
\caption{Operation-conditioned memory retrieval in \openmleevo{}. Ancestor, sibling, and related-node caps are configurable; the table reports the defaults used by the implementation.}
\label{tab:openmleevo-operation-memory}\\
\toprule
\textbf{Operator} & \textbf{Default retrieved evidence} & \textbf{Purpose of the rendered memory} \\
\midrule
\endfirsthead
\multicolumn{3}{c}{\tablename\ \thetable\ (continued)}\\
\toprule
\textbf{Operator} & \textbf{Default retrieved evidence} & \textbf{Purpose of the rendered memory} \\
\midrule
\endhead
\midrule
\multicolumn{3}{r}{Continued on next page}\\
\endfoot
\bottomrule
\endlastfoot
\textsc{Draft} & No inherited node memory. & Starts an independent branch from the task specification. \\
\textsc{Improve} & The selected parent, its three most recent ancestors, and its top three direct siblings. Siblings share at least one parent and are ranked by the same quality--progress--novelty utility used for parent selection. Relevant board fields are appended. & Preserves what works in the parent, identifies changes that helped or hurt along its lineage, and contrasts nearby alternatives without replaying the full tree. \\
\textsc{Crossover} & Two selected parents; for each parent, two recent ancestors and two top-ranked direct siblings; family-level statistics, repeated errors, and a method-family complementarity cue. & Identifies compatible strengths and conflicts between branches and discourages a mechanical concatenation of both programs. \\
\textsc{Debug} & The current buggy node followed first by prior nodes with the same error signature and then by recent attempts, up to a default total of three related nodes; repeated-error counts are included. & Reuses fixes for the same failure mode while retaining a recent-context fallback for previously unseen errors. \\
\end{longtable}
\endgroup

The rendered memory groups selected cards, cached summaries, and board fields into named sections. \textsc{Improve} receives the selected parent, vertical ancestors, horizontal siblings, and related board statistics; \textsc{Crossover} repeats the branch sections for both parents and adds the complementarity cue; and \textsc{Debug} receives the current signature, repeated-error counts, current buggy node, and related errors. This bounded branch-local context replaces an unbounded concatenation of prior analyses. The exact system and user-message templates are provided below.

\hypertarget{openmleevo-rich-memory-prompt}{\paragraph{Complete rich-memory synthesis prompt.}}
The production system prompt and parameterized user-message template used to extract reusable memories from past trajectories are shown below. Double-braced expressions denote runtime fields populated from the task, the current node, and its direct parent.

\begin{Verbatim}
[SYSTEM MESSAGE]

You are an expert machine learning experiment analyst.
Your job is to distill one completed experiment node into reusable memory.

The memory should explain:
1. What the current node's code actually does, with concrete modeling and implementation details.
2. What was learned by comparing the current node with its parent, especially whether the change improved or worsened the result.

Be specific and evidence-based.
Use only the provided code, execution outputs, scores, delta, runtime, and status.
Do not invent results or claim improvements that are not supported by the metadata.

[USER MESSAGE TEMPLATE]

# Task Description
{{task_description}}

# Current Node Metadata
node_id: {{node_id}}
operator_that_created_this_node: {{operator}}
method_family: {{method_family}}
status: {{status}}
score: {{score}}
delta_vs_parent: {{delta_vs_parent}}
runtime_seconds: {{runtime}}
error_signature: {{error_signature}}

# Parent Node Metadata
parent_node_id: {{parent_node_id}}
parent_method_family: {{parent_method_family}}
parent_status: {{parent_status}}
parent_score: {{parent_score}}
parent_runtime_seconds: {{parent_runtime}}

# Current Node Plan
{{current_plan}}

# Parent Node Plan
{{parent_plan}}

# Current Node Code
{{current_code}}

# Current Node Execution Output
{{current_execution_output}}

# Parent Node Code
{{parent_code}}

# Parent Node Execution Output
{{parent_execution_output}}

# Instructions
Write reusable memory for this experiment node.

Focus on two things:

1. method_overview
Summarize what the current code does.
Include concrete details such as:
- model family and main estimator
- feature engineering
- validation strategy
- ensembling or post-processing
- training/runtime choices
- submission generation logic

2. parent_comparison_experience
Compare the current node with its parent.
Explain:
- what changed from the parent
- whether the score/status/runtime improved, worsened, or stayed similar
- what the delta suggests
- what experience should be reused if it improved
- what should be avoided or fixed if it worsened or failed

If the node has no parent, say that it is an initial draft and summarize the initial strategy.
If the node failed, focus on the failure mode and what should be avoided or fixed.

Keep the answer concise but information-dense.

# Output Format
Return exactly one valid JSON object and nothing else.
Do not return markdown, code fences, headings, or prose outside the JSON object.

Use exactly these keys:

{
  "method_overview": string,
  "parent_comparison_experience": string
}

Field requirements:
- method_overview: 2-5 sentences describing the concrete method used by the current node.
- parent_comparison_experience: 2-5 sentences comparing current node vs parent and extracting success or failure experience.
\end{Verbatim}

\paragraph{Grounded rendered context.}
In the \texttt{leaf-classification} example, the selected ResNet50, handcrafted-geometry, and XGBoost parent had log loss 0.765432. Horizontal memories exposed a simpler ResNet18 plus tabular-feature incumbent at 0.031825 and a feature-heavy sibling that timed out after 7,200 seconds; the board reported \nolinkurl{current_best_family=neural_net+cv}, \nolinkurl{repeated_errors: timeout=3}, and the parent-family best and failure rate. The next \textsc{Improve} operation accordingly dropped the oversized ResNet50 and slow handcrafted pipeline, retained ResNet18 and tabular features, and added calibrated out-of-fold ensembling, yielding 0.012080. This grounds generated memory in both local alternatives and global search state.

\section{Supplementary Experiments}

\subsection{Repeated-Evaluation Statistics on MLE-Bench Lite}
\label{app:openmle-repeat-evaluation}

For model--harness configurations with available repeat-level records, we
aggregate three evaluation epochs and report the corresponding mean and
standard deviation in Table~\ref{tab:openmle-repeat-results}. Repeated
evaluation reduces sensitivity to favorable sampling trajectories and
transient sandbox behavior, while the standard deviation quantifies the
remaining run-to-run variability.

Owing to the substantial inference and sandbox cost, Codex, Claude Code,
and Gemini CLI references were evaluated only once and are
therefore retained as point estimates in the main results table. Entries
without $\pm$ in Table~\ref{tab:openmle-repeat-results} indicate metrics
for which the archived summary contains the aggregate estimate but not
the underlying repeat-level dispersion. All reported $\pm$ values denote
standard deviations rather than confidence intervals.

\begin{table*}[t]
\centering
\scriptsize
\setlength{\tabcolsep}{5.5pt}
\renewcommand{\arraystretch}{1.10}
\caption{
Repeated-evaluation results on MLE-Bench Lite under the OpenMLE
harnesses. Values are mean $\pm$ standard deviation across three
evaluation epochs. Valid Rate is the mean number of tasks producing
valid submissions out of 22; Medal Average and Human Rank are
higher-is-better. 
}
\label{tab:openmle-repeat-results}
\begin{tabular}{@{}llccc@{}}
\toprule
Model / system
& Framework
& Valid Rate
& Medal Average $\uparrow$
& Human Rank $\uparrow$ \\
\midrule

\shortstack[l]{Qwen3-30B-A3B-Thinking-2507}
& \openmleevo{}
& $17.33 \pm 0.47$/22
& $34.85\% \pm 2.14\%$
& $0.5573 \pm 0.0074$ \\

Frontis-MA1-30B
& \openmleevo{}
& $21.67 \pm 0.47$/22
& $53.03\% \pm 4.29\%$
& $0.7055 \pm 0.0505$ \\

Frontis-MA1-30B
& \openmleevomax{}
& $22.00 \pm 0.00$/22
& $66.67\% \pm 5.67\%$
& $0.8053 \pm 0.0236$ \\

\addlinespace[2pt]

Qwen3.6-35B-A3B
& \openmleevo{}
& $19.67 \pm 0.47$/22
& $39.39\% \pm 5.67\%$
& $0.5828 \pm 0.0278$ \\

Frontis-MA1-35B
& \openmleevo{}
& $21.67 \pm 0.47$/22
& $60.61\% \pm 7.73\%$
& $0.7647 \pm 0.0376$ \\

Frontis-MA1-35B
& \openmleevomax{}
& $22.00 \pm 0.00$/22
& $\mathbf{71.21\% \pm 8.57\%}$
& $0.8126 \pm 0.0388$ \\
\midrule

GLM-5.2
& \openmleevomax{}
& $22.00 \pm 0.00$/22
& $66.67\% \pm 8.57\%$
& $\mathbf{0.8164 \pm 0.0233}$ \\

MiniMax M3
& \openmleevo{}
& $22.00 \pm 0.00$/22
& $59.09\% \pm 0.00\%$
& $0.7994 \pm 0.0225$ \\

MiniMax M3
& \openmleevomax{}
& $22.00 \pm 0.00$/22
& $65.15\% \pm 2.14\%$
& $0.8007 \pm 0.0089$ \\

Kimi K2.6
& \openmleevo{}
& $21.67 \pm 0.47$/22
& $66.67\% \pm 5.67\%$
& $0.7859 \pm 0.0285$ \\

Grok-4.5
& \openmleevo{}
& $22.00 \pm 0.00$/22
& $65.15\% \pm 2.14\%$
& $0.8052 \pm 0.0170$ \\

LongCat-2.0
& \openmleevo{}
& $21.00 \pm 0.82$/22
& $56.06\% \pm 5.67\%$
& $0.7343 \pm 0.0150$ \\

Doubao Seed 2.1 Pro
& \openmleevo{}
& $20.33 \pm 0.47$/22
& $56.06\% \pm 2.14\%$
& $0.7170 \pm 0.0397$ \\

Qwen3.7 Plus
& \openmleevo{}
& $21.67 \pm 0.47$/22
& $54.55\% \pm 6.43\%$
& $0.7234 \pm 0.0408$ \\

DeepSeek-V4-Pro
& \openmleevo{}
& $21.67 \pm 0.47$/22
& $54.55\% \pm 3.71\%$
& $0.6849 \pm 0.0258$ \\

DeepSeek-V4-Flash
& \openmleevo{}
& $21.33 \pm 0.47$/22
& $51.52\% \pm 5.67\%$
& $0.6957 \pm 0.0200$ \\

GLM-4.7
& \openmleevo{}
& $21.33 \pm 0.47$/22
& $51.52\% \pm 7.73\%$
& $0.6521 \pm 0.0543$ \\

MiniMax M2.7
& \openmleevo{}
& $22.00 \pm 0.00$/22
& $50.00\% \pm 7.42\%$
& $0.7039 \pm 0.0298$ \\

MiMo-V2.5-Pro
& \openmleevo{}
& $17.00 \pm 1.63$/22
& $40.91\% \pm 3.71\%$
& $0.5213 \pm 0.0422$ \\

Step-3.7 Flash
& \openmleevo{}
& $19.00 \pm 0.00$/22
& $27.27\% \pm 6.43\%$
& $0.4953 \pm 0.0385$ \\

\bottomrule
\end{tabular}
\end{table*}

\subsection{NatureBench Lite Task Composition}
\label{app:naturebench-lite}

Our generalization study uses the fixed 10-task NatureBench Lite subset summarized in Table~\ref{tab:naturebench-lite-tasks}.
The subset favors moderately tractable tasks while preserving coverage across all six NatureBench scientific domains and diverse data structures, including biological sequences, omics matrices, molecular structures, temporal signals, images, and tabular features.
This breadth makes the subset useful for rapid model--harness comparisons, but its ten-task size means that each task changes All S or All M by ten percentage points.

\section{Simplified comparison of public release surfaces}

The public release landscape remains fragmented. Table~\ref{tab:mle-open-release-matrix} summarizes representative MLE-agent and MLE-resource work by the artifacts needed to reproduce a full post-training stack. Scores are not strictly comparable because the systems differ in backbone
model, compute and wall-clock budgets, hardware, external-resource access,
number of runs, and aggregation procedure. Checkmarks denote artifacts that
were publicly accessible and independently verifiable from the cited paper or
repository at the time of audit. High-scoring systems often release an inference framework, evaluation entry point, or leaderboard result; resource projects provide tasks and environments; training-oriented papers describe RL designs for MLE agents~\citep{qiang2025mlesmith,liu2025mlagent,li2025mlerl,yang2025rlmleagents,cai2026acegrpo}. Across these threads, the combination of executable training data, sandbox infrastructure, training code, RL method, evaluation framework, and model weights is still rare. This gap motivates treating open MLE capability as a full-stack problem spanning data, execution, optimization, and inference.

\begin{table}[htbp]
\centering
\caption{The 10 tasks in NatureBench Lite used for the generalization experiment.}
\label{tab:naturebench-lite-tasks}
\vspace{0.4em}
\scriptsize
\setlength{\tabcolsep}{3pt}
\renewcommand{\arraystretch}{1.1}
\begin{tabular}{@{}>{\raggedright\arraybackslash}p{0.31\linewidth}>{\raggedright\arraybackslash}p{0.17\linewidth}>{\raggedright\arraybackslash}p{0.24\linewidth}>{\raggedright\arraybackslash}p{0.20\linewidth}@{}}
\toprule
Task & Domain & Input modality & ML task type \\
\midrule
Spatial RNA Velocity Inference & Cellular Omics & Single-cell and spatial omics & Simulation / operator learning \\
Disease-Specific Variant Effect Prediction & Cellular Omics & Biological sequence & Prediction / regression \\
Metabolomic Profile Prediction from Microbial Composition & Cellular Omics & Tabular / feature matrix & Prediction / regression \\
Protein Variant Effect Prediction & Protein Biology & Biological sequence & Prediction / regression \\
Lasso Peptide Property Prediction & Protein Biology & Biological sequence & Prediction / regression \\
Anomalous Diffusion Out-of-Distribution Dynamics Detection & Physical Modeling & Temporal / signal / spectra & Classification \\
Zeolite--Molecule Binding Affinity Prediction & Physical Modeling & Molecular / materials structure & Prediction / regression \\
Spatial Clustering of Single-Molecule Localization Point Clouds & Biomedical Modeling & Image / volumetric & Clustering / integration \\
Molecular Property Prediction & Molecular Design & Molecular / materials structure & Prediction / regression \\
Categorical Counterfactual Outcome Estimation & Relational Reasoning & Tabular / feature matrix & Classification \\
\bottomrule
\end{tabular}
\end{table}

\begin{table}[H]
\centering
\caption{
Audited comparison of public release surfaces for representative MLE agents and resources.
Artifact availability was rechecked against the cited papers, official repositories, and the official MLE-Bench leaderboard submissions as of July 2026.
}
\label{tab:mle-open-release-matrix}
\vspace{0.3em}
\scriptsize
\setlength{\tabcolsep}{2.6pt}
\renewcommand{\arraystretch}{1.08}
\resizebox{\linewidth}{!}{%
\begin{tabular}{@{}lccccccccc@{}}
\toprule
Work
& Data
& Sandbox
& \shortstack{Train\\code}
& \shortstack{RL\\method}
& Eval
& Weights
& \shortstack{MLE-Bench Lite\\Medal Rate}
& \shortstack{Run\\setting}
& Best model \\
\midrule

\multicolumn{10}{@{}l}{\textit{No trained MLE model released}} \\

AIDE~\citep{jiang2025aide}
& \xmark & \cmark & \xmark & \xmark & \cmark & \xmark
& 35.91\%
& \mbox{24h $\cdot$ 1$\times$A10}
& o1-preview \\

AutoMLGen / InternAgent~\citep{du2025automlgen}
& \xmark & \cmark & \xmark & \xmark & \cmark & \xmark
& 62.12\%
& \mbox{12h $\cdot$ 1$\times$A800}
& DeepSeek-R1 \\

ML-Master 2.0~\citep{liu2025ml,zhu2026toward}
& \xmark & \cmark & \xmark & \xmark & \cmark & \xmark
& 75.76\%
& \mbox{24h $\cdot$ 2$\times$RTX~4090}
& DeepSeek-V3.2-Speciale \\

MLE-STAR-Pro-1.5~\citep{nam2025mlestar}
& \xmark & \cmark & \xmark & \xmark & \cmark & \xmark
& 68.18\%
& \mbox{24h $\cdot$ 2$\times$A100-40G}
& Gemini-2.5-Pro \\

MLZero~\citep{fang2025mlzero}
& \xmark & \cmark & \xmark & \xmark & \cmark & \xmark
& 36.36\%$^{\dagger}$
& \mbox{24h $\cdot$ 8$\times$A100-40G}
& Claude-Sonnet-3.7 \\

AIRA-dojo~\citep{toledo2025airesearchagents}
& \xmark & \cmark & \xmark & \xmark & \cmark & \xmark
& 55.00\%
& \mbox{24h $\cdot$ 1$\times$H200}
& o3 \\

Famou-Agent 2.0~\citep{li2025fmagent}
& \xmark & \xmark & \xmark & \xmark & \cmark & \xmark
& 80.30\%
& \mbox{24h $\cdot$ 1$\times$A800}
& Gemini-3-Pro-Preview \\

MLEvolve~\citep{du2026mlevolve}
& \xmark & \cmark & \xmark & \xmark & \cmark & \xmark
& 80.30\%
& \mbox{12h $\cdot$ 1$\times$H200}
& Gemini-3-Pro-Preview \\

AIBuildAI~\citep{zhang2026aibuildai}
& \cmark & \cmark & \xmark & \xmark & \cmark & \xmark
& 77.27\%
& \mbox{24h $\cdot$ 1$\times$A100}
& Claude-Opus-4.6 \\

MLAgentBench~\citep{huang2023mlagentbench}
& \cmark & \cmark & \xmark & \xmark & \cmark & \xmark
& --
& --
& -- \\

MLGym~\citep{nathani2025mlgym}
& \cmark & \cmark & \xmark & \xmark & \cmark & \xmark
& --
& --
& -- \\

MLE-Dojo~\citep{qiang2025mledojo}
& \cmark & \cmark & \xmark & \xmark & \cmark & \xmark
& --
& --
& -- \\

MLE-Smith~\citep{qiang2025mlesmith}
& \xmark & \xmark & \xmark & \xmark & \xmark & \xmark
& --
& --
& -- \\

R\&D-Agent~\citep{yang2025r}
& \xmark & \cmark & \xmark & \xmark & \cmark & \xmark
& 68.18\%
& \mbox{12h $\cdot$ 1$\times$V100}
& GPT-5 \\

\midrule
\multicolumn{10}{@{}l}{\textit{Post-trained MLE agents}} \\

RL-MLE~\citep{yang2025rlmleagents}
& \xmark & \xmark & \xmark & \cmark & \xmark & \xmark
& --
& --
& -- \\

ML-Agent~\citep{liu2025mlagent}
& \xmark & \xmark & \cmark$^{\ddagger}$ & \cmark & \xmark & \xmark
& --
& --
& -- \\

MLE-RL~\citep{li2025mlerl}
& \xmark & \xmark & \xmark & \cmark & \xmark & \xmark
& 33.30\%
& \mbox{12h $\cdot$ A10 (count NR)}
& MLE-RL-32B-S \\

AceGRPO~\citep{cai2026acegrpo}
& \xmark & \xmark & \xmark & \cmark & \xmark & \xmark
& 51.52\%
& \mbox{12h $\cdot$ GPU NR}
& Ace-30B \\

\textbf{OpenMLE (ours)}
& \cmark & \cmark & \cmark & \cmark & \cmark & \cmark
& \textbf{\frontisThirtyFiveEvoMaxResult{}}
& \mbox{12h $\cdot$ RTX~4090(12G VRAM)}
& \thirtyfivebmodel{} \\

\bottomrule
\end{tabular}%
}

\vspace{0.3em}
\begin{minipage}{\linewidth}
\tiny
\textit{Release criteria.}
For \emph{Data}, \emph{Sandbox}, \emph{Train code}, \emph{Eval}, and
\emph{Weights}, a tick requires an artifact that was publicly accessible
at the time of audit.
\emph{Data} requires downloadable task/training data or scripts for
reconstructing it; \emph{Sandbox} requires released code capable of
running model-generated programs or constructing their execution
environment; \emph{Train code} requires model-parameter training
entry points or configurations; \emph{Eval} accepts runnable evaluation
assets or official per-run grading reports; and \emph{Weights} requires
downloadable trained MLE-agent weights.
A tick under \emph{RL method} indicates that the work develops or applies
an MLE-specific RL post-training method; it does not by itself imply that
the RL implementation is public.
Run settings report the per-task wall-clock limit and sandbox GPU
allocation; remote LLM-serving compute and CPU/RAM are omitted.
NR means not reported, and ``--'' means that no corresponding
MLE-Bench Lite result is available.
GPU-h denotes total sandbox accelerator time rather than elapsed
wall-clock time.

$^{\dagger}$ MLZero evaluated 21 of the 22 MLE-Bench Lite tasks,
excluding one task because of preprocessing inconsistencies.
The displayed rate uses the standard 22-task denominator, counting
the excluded task as a non-medal result.

$^{\ddagger}$ ML-Agent currently releases its exploration-enriched SFT
training code only; its step-wise RL implementation, training dataset,
evaluation code, and model checkpoints remain unreleased.
\end{minipage}
\end{table}